\documentclass[pdflatex,sn-nature]{sn-jnl}% Style for submissions to Nature Portfolio journals
\usepackage{graphicx}%
\usepackage{multirow}%
\usepackage{amsmath,amssymb,amsfonts}%
\usepackage{thmtools,thm-restate}
\usepackage{amsthm}%
\usepackage{mathrsfs}%
\usepackage[title]{appendix}%
\usepackage{xcolor}%
\usepackage{textcomp}%
\usepackage{manyfoot}%
\usepackage{booktabs}%
\usepackage{algorithm}%
\usepackage{algorithmicx}%
\usepackage{algpseudocode}%
\usepackage{listings}%
\usepackage{threeparttable}
\usepackage{array}
\usepackage{titletoc}
\usepackage{xparse}

\usepackage[most]{tcolorbox}
\usepackage{xcolor}

\newtcolorbox{promptbox}{
    enhanced,
    breakable,
    colback=black!3,
    colframe=black!18,
    boxrule=0.4pt,
    arc=0.8mm,
    left=2.5mm,
    right=2.5mm,
    top=2mm,
    bottom=2mm,
    before skip=7pt,
    after skip=7pt,
    fontupper=\small,
}

\usepackage{geometry}
\theoremstyle{thmstyleone}%
\newtheorem{theorem}{Theorem}%  meant for continuous numbers
\theoremstyle{thmstyletwo}%
\newtheorem*{remark*}{Remark}

\theoremstyle{thmstylethree}%
\newtheorem{definition}[theorem]{Definition}%

\begin{document}

\title[Representational Atoms in Neural Networks]{Representational Atoms in Neural Networks}


%\author*[1,2]{\fnm{First} \sur{Author}}\email{iauthor@gmail.com}
%\affil*[1]{\orgdiv{Department}, \orgname{Organization}, \orgaddress{\street{Street}, \city{City}, \postcode{100190}, \state{State}, \country{Country}}}
%
%\affil[2]{\orgdiv{Department}, \orgname{Organization}, \orgaddress{\street{Street}, \city{City}, \postcode{10587}, \state{State}, \country{Country}}}
%
%\affil[3]{\orgdiv{Department}, \orgname{Organization}, \orgaddress{\street{Street}, \city{City}, \postcode{610101}, \state{State}, \country{Country}}}

\author[1,2]{\fnm{Chenhui} \sur{Hu}}
\author[1,2]{\fnm{Pengfei} \sur{Cao}}
\author[1,2]{\fnm{Kang} \sur{Liu}}
\author*[1,2]{\fnm{Jun} \sur{Zhao}}\email{jzhao@nlpr.ia.ac.cn}

\affil[1]{\centering\orgdiv{The Key Laboratory of Cognition and Decision Intelligence for Complex Systems},\\ \orgname{Institute of Automation, Chinese Academy of Sciences}, \orgaddress{\city{Beijing}, \country{China}}}

\affil[2]{\centering\orgdiv{School of Artificial Intelligence},\\ \orgname{University of Chinese Academy of Sciences}, \orgaddress{\city{Beijing}, \country{China}}}


\abstract{Understanding the internal representations of neural networks and characterizing their relationship with human conceptual structure are two critical questions for elucidating model mechanisms. The former reveals how learned knowledge is organized in internal representations, whereas the latter provides a basis for interpreting these representations through human conceptual structure. Here, we formally define fundamental representational units (FRUs) and establish two evaluation criteria: fidelity and stability. We further prove that threshold-activated sparse autoencoders (TSAEs) can identify FRUs with high fidelity and stability under appropriate conditions, and validate this result through large-scale experiments. Further analyses reveal that FRUs with higher stability exhibit greater monosemanticity and can be composed across layers to recover semantic organization corresponding to human conceptual structure. These results provide a unified framework for defining, evaluating and identifying FRUs in neural networks, and establish a foundation for characterizing the relationship between neural network representations and human conceptual structure.}

\maketitle

%\section{Introduction}\label{sec1}

Neural networks progressively form internal representations during forward propagation, encoding input-related information that supports subsequent prediction or generation \cite{lecun2015deep,bengio2013representation}. However, how these representations encode and organize the information learned by the model remains incompletely understood \cite{bau2020understanding,klindt2026unifying}. A fundamental question is whether this representational organization can be understood in terms of basic units that can be rigorously defined and reliably identified. If so, these units would provide a natural basis for characterizing the semantic organization of internal representations, raising a further question of whether and how this organization relates to human conceptual structure. Addressing these questions is essential for understanding neural representations and their relationship with human conceptual structure.

To answer the first question, we develop a theoretical framework for defining, evaluating and identifying fundamental representational units (FRUs) underlying internal representations. Drawing on compressed sensing \cite{donoho2006compressed,candes2006robust}, we formulate internal representations as sparse linear combinations of FRUs: linearity provides a simple and analytically tractable form for characterizing how these FRUs combine, while sparsity permits them to remain distinguishable under appropriate conditions. On this basis, we formally define the FRUs as representational atoms, or atoms for brevity. This definition operates at the level of internal representation spaces and does not depend on particular neural network architectures.

Within this framework, we next evaluate how closely candidate representational units conform to the definition of representational atoms. Specifically, we introduce two evaluation criteria, fidelity and stability. Fidelity measures how well candidate FRUs reconstruct the original representations, whereas stability measures whether distinct candidate FRUs remain distinguishable from one another. We further establish that the distinguishability captured by stability is a prerequisite for monosemanticity, that is, for each FRU to support a unique semantic interpretation. These criteria enable two widely studied representational units, neurons and learned features \cite{chen2025knowledge,huben2024sparse,bricken2023monosemanticity,templeton2024scaling}, to be quantitatively compared as candidate FRUs within a unified framework. Our experiments show that neurons generally exhibit high fidelity but low stability, whereas learned features exhibit higher stability but lower fidelity. This contrast is consistent with previous observations that learned features tend to be more monosemantic than individual neurons \cite{elhage2022toy,chen2025microscope,bricken2023monosemanticity,templeton2024scaling}.

The limitations of existing candidate FRUs raises a critical question: what kind of representation learning mechanism can reliably identify atoms with both high fidelity and stability? To address this question, we prove that threshold-activated sparse autoencoders (TSAEs) can effectively identify representational atoms under appropriate conditions. In practice, we instantiate TSAEs as BiJumpReLU SAEs and evaluate them on internal representations from diverse pretrained neural networks spanning different architectures and representation modalities. Across these settings, BiJumpReLU SAEs consistently identify atoms with high fidelity and stability, outperforming representative SAE variants spanning different activation and sparsity mechanisms \cite{huben2024sparse,gao2025scaling,zhu2026abstopk,rajamanoharan2024jumping}. These large-scale experiments support our theoretical results and establish a practical basis for identifying and analyzing FRUs in neural networks. We further examine whether the theoretical connection between stability and monosemanticity is reflected empirically and find a significant positive correlation between the two, with more stable atoms exhibiting clearer and more consistent semantics.

We finally examine whether and how representational atoms relate to human conceptual structure using HowNet \cite{Dong2003HowNetA,qi2019openhownet}, an expert-annotated knowledge base that represents concepts in terms of fine-grained semantic primitives known as sememes. We find that their relationship is not one-to-one: although a small subset of atoms directly aligns with complete sememes, individual sememes are more commonly represented in a distributed manner by multiple atoms from different layers. These findings indicate that representational atoms function as finer-grained semantic components that can be composed into higher-order structures corresponding to human conceptual organization. Notably, their compositions recover not only sememe structures with substantially greater alignment than single atoms but also additional semantic structures that, upon manual verification, are consistent with the corresponding sememes yet not captured by their HowNet annotations. Overall, our work integrates the definition, evaluation and identification of FRUs with the analysis of their semantic and compositional organization, providing a systematic characterization of neural representational structure and its relationship with human conceptual structure.

\section*{Results}\label{Section:Results}

In the following, we first formally define FRUs and introduce fidelity and stability as two evaluation criteria. We then quantitatively compare neurons and learned features as candidate FRUs within this unified framework. Next, we prove that TSAEs can effectively identify representational atoms under appropriate conditions and validate their practical effectiveness through large-scale experiments. We further examine the relationship between stability and monosemanticity, and finally use HowNet to characterize the relationship between neural representations and human conceptual structure.

\subsection*{Representational atoms}\label{Section:Representational atoms}

Neurons and learned features are two widely studied candidates for the basic units of internal representations \cite{chen2025knowledge,huben2024sparse,bricken2023monosemanticity,templeton2024scaling}, yet neither provides a principled basis for determining what properties representational components must possess to qualify as FRUs. We therefore begin by formalizing this problem and developing a principled definition of FRUs.

Given a set of high-dimensional internal representations \(M=\{\boldsymbol m_i\mid \boldsymbol m_i\in\mathbb R^H\}_{i=1}^{|M|}\), we seek a set of FRUs \(D=\{\boldsymbol d_j\mid \boldsymbol d_j\in\mathbb R^H\}_{j=1}^{|D|}\). A natural starting point is the standard orthonormal basis of the representation space, treating the coordinate directions (i.e. neurons) as candidate FRUs. However, this choice has two limitations. First, individual coordinate directions need not support stable and well-defined semantics, consistent with the widespread polysemanticity of individual neurons \cite{elhage2022toy,bricken2023monosemanticity,templeton2024scaling}. Second, orthogonality restricts the number of candidate FRUs to the dimensionality of the representation space, limiting their ability to capture richer representational structure.

\begin{figure}
    \centering
    \includegraphics[width=0.98\textwidth]{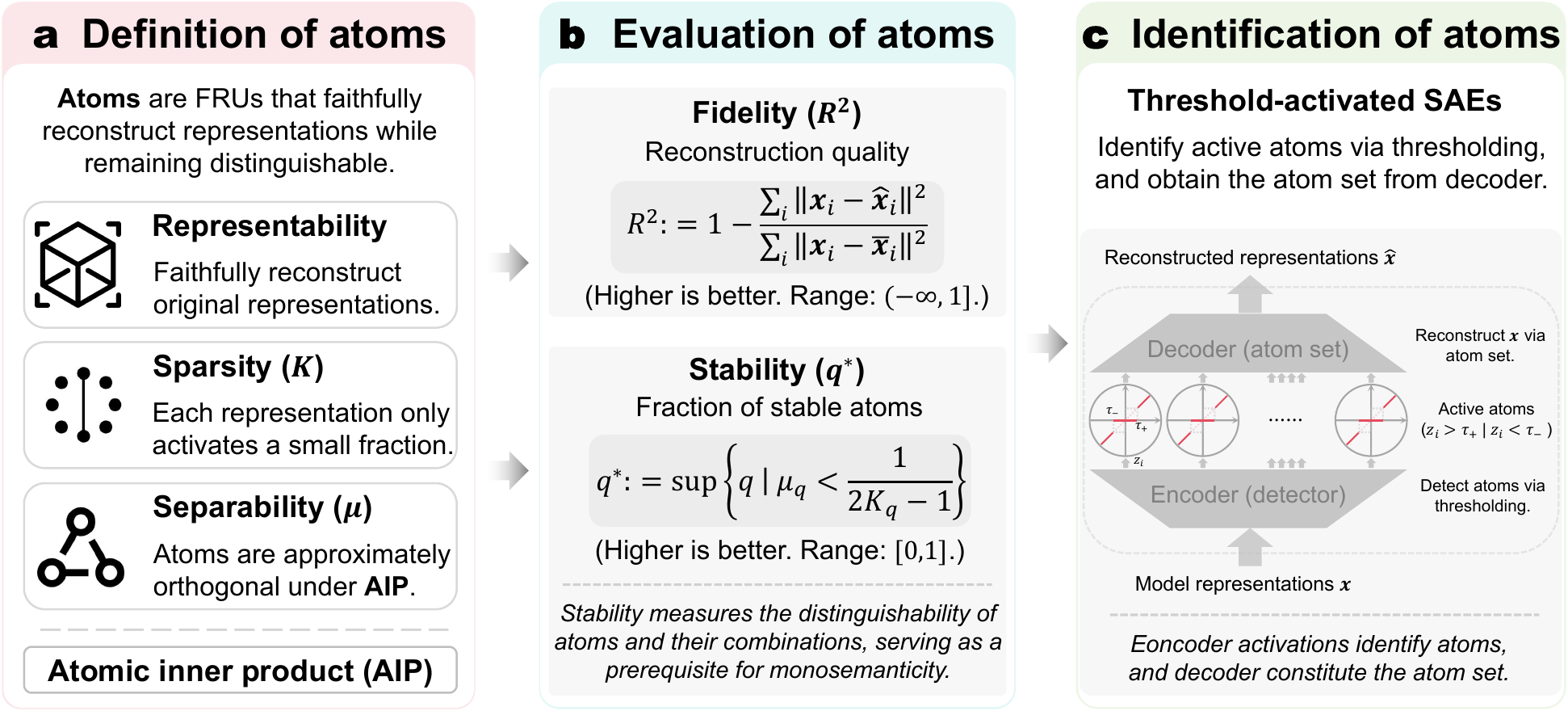}
    \caption{\textbf{Overview of the framework for defining, evaluating and identifying representational atoms.} \textbf{a}, Representational atoms are defined as fundamental representational units (FRUs) that faithfully reconstruct model representations while remaining distinguishable. This definition is characterized by three properties: representability, sparsity and separability under the atomic inner product (AIP). \textbf{b}, Candidate FRUs are evaluated along two quantitative dimensions. Fidelity, quantified by the coefficient of determination \(R^2\), measures the reconstruction of the original representations, whereas stability, quantified by \(q^*\), measures the extent to which atoms remain distinguishable under the condition \(\mu_q < 1/(2K_q-1)\). Higher values indicate better performance for both metrics, with \(R^2\in(-\infty,1]\) and \(q^*\in[0,1]\). Stability captures the distinguishability of atoms and their compositions and constitutes a prerequisite for monosemanticity. \textbf{c}, Representational atoms are identified using threshold-activated sparse autoencoders (TSAEs). The encoder acts as a detector that activates atoms when its response exceeds the positive or negative threshold, while the corresponding decoder vectors constitute the atom set and reconstruct the original representations.}
  \label{fig:Fig_1_Overview}
\end{figure}

The key property of the standard orthonormal basis is that orthogonality makes distinct coordinate directions and their combinations unambiguously distinguishable, which ensures a unique decomposition of each representation. However, a key insight here is that orthogonality is not necessary for distinguishability. In particular, compressed sensing shows that sparsity can preserve the distinguishability among non-orthogonal components under appropriate conditions \cite{donoho2006compressed,candes2006robust}, which we specify below. We therefore model each internal representation as a sparse linear combination
\begin{equation}
    \forall i\in [|M|],\quad \boldsymbol m_i=\boldsymbol D\boldsymbol\delta_i, \qquad \|\boldsymbol\delta_i\|_0\le K,
\end{equation}
where \([|M|]=\{1,2,\ldots,|M|\}\), \(\boldsymbol D=[\boldsymbol d_1\ \boldsymbol d_2\ \cdots\ \boldsymbol d_{|D|}]\in\mathbb R^{H\times |D|}\) denotes the dictionary of candidate FRUs, \(\boldsymbol\delta_i\in\mathbb R^{|D|}\) is the sparse coefficient vector associated with \(\boldsymbol m_i\), \(\|\cdot\|_0\) denotes the number of non-zero entries, and \(K\in\mathbb N^+\) is an upper bound on the sparsity level.

Candidate FRUs that satisfy the distinguishability condition derived below are termed representational atoms, or simply atoms (Fig.~\ref{fig:Fig_1_Overview}a). Characterizing this condition in the non-orthogonal setting requires an inner product that captures the relative geometry among candidate FRUs. The standard Euclidean inner product, \(\langle\boldsymbol d_i,\boldsymbol d_j\rangle=\boldsymbol d_i^\top\boldsymbol d_j\), characterizes geometry in the ambient representation space, but does not fully capture the geometry induced by internal representations (see `Representational atoms theoretical framework' in Methods). We therefore introduce the atomic inner product (AIP), \(\langle\boldsymbol d_i,\boldsymbol d_j\rangle_{\tilde S}=\boldsymbol d_i^\top\tilde S\boldsymbol d_j\), where \(\tilde S\) denotes the representation-induced metric, to characterize their relative geometry (see `Atomic inner product' in Methods). Empirical analyses further show that the AIP better captures the geometry relevant to representational distinguishability (see `Empirical validation of the atomic inner product' section in Supplementary Information). 

Using the AIP, we derive a condition under which candidate FRUs remain distinguishable in the non-orthogonal setting. Specifically, extending the classical coherence-based condition
\cite{donoho2003optimally,tropp2004greed}, we prove that, for a \(K\)-sparse representation, distinguishability is preserved when \(\mu<\frac{1}{2K-1}\), where \(\mu=\max_{i\neq j}\left|\langle\boldsymbol d_i,\boldsymbol d_j\rangle_{\tilde S}\right|\) denotes the maximum coherence among candidate FRUs (see `Evaluation of atoms' in Methods). Within this regime, distinct units and their compositions remain distinguishable from alternative units or compositions, a property we term monorepresentationality. Monorepresentationality constitutes a necessary structural condition for monosemanticity, as each representational unit can support a unique semantic interpretation only when it remains distinguishable from alternative units or compositions. See `Evaluation of atoms' in Methods for the formal definition and discussion of monorepresentationality.

To quantitatively assess how closely candidate FRUs conform to the properties of representational atoms, we introduce two complementary metrics, fidelity and stability (Fig.~\ref{fig:Fig_1_Overview}b). Fidelity measures how well candidate FRUs reconstruct the original internal representations and is quantified by the coefficient of determination
\begin{equation}
    R^2:=1-\frac{\sum_i\|\boldsymbol m_i-\hat{\boldsymbol m}_i\|_2^2}{\sum_i\|\boldsymbol m_i-\bar{\boldsymbol m}\|_2^2},
\end{equation}
where \(\hat{\boldsymbol m}_i\) denotes the reconstructed representation and \(\bar{\boldsymbol m}\) denotes the mean of the internal representations (see `Fidelity metrics and computation' in Methods). Stability measures the extent to which candidate FRUs satisfy the distinguishability condition and is quantified by the maximum quantile
\begin{equation}
    q^*:=\sup\left\{q\mid \mu_q<\frac{1}{2K_q-1}\right\},
\end{equation}
where \(\mu_q\) and \(K_q\) denote the coherence and sparsity level at quantile \(q\), respectively (see `Stability metrics and computation' in Methods). Ideal representational atoms achieve \(R^2=1\) and \(q^*=1\), providing a quantitative reference for evaluating candidate FRUs in the following analyses.

\subsection*{Limitations of existing representational units}\label{Section:Limitations of existing representational units}

\begin{figure}
    \centering
    \includegraphics[width=0.98\textwidth]{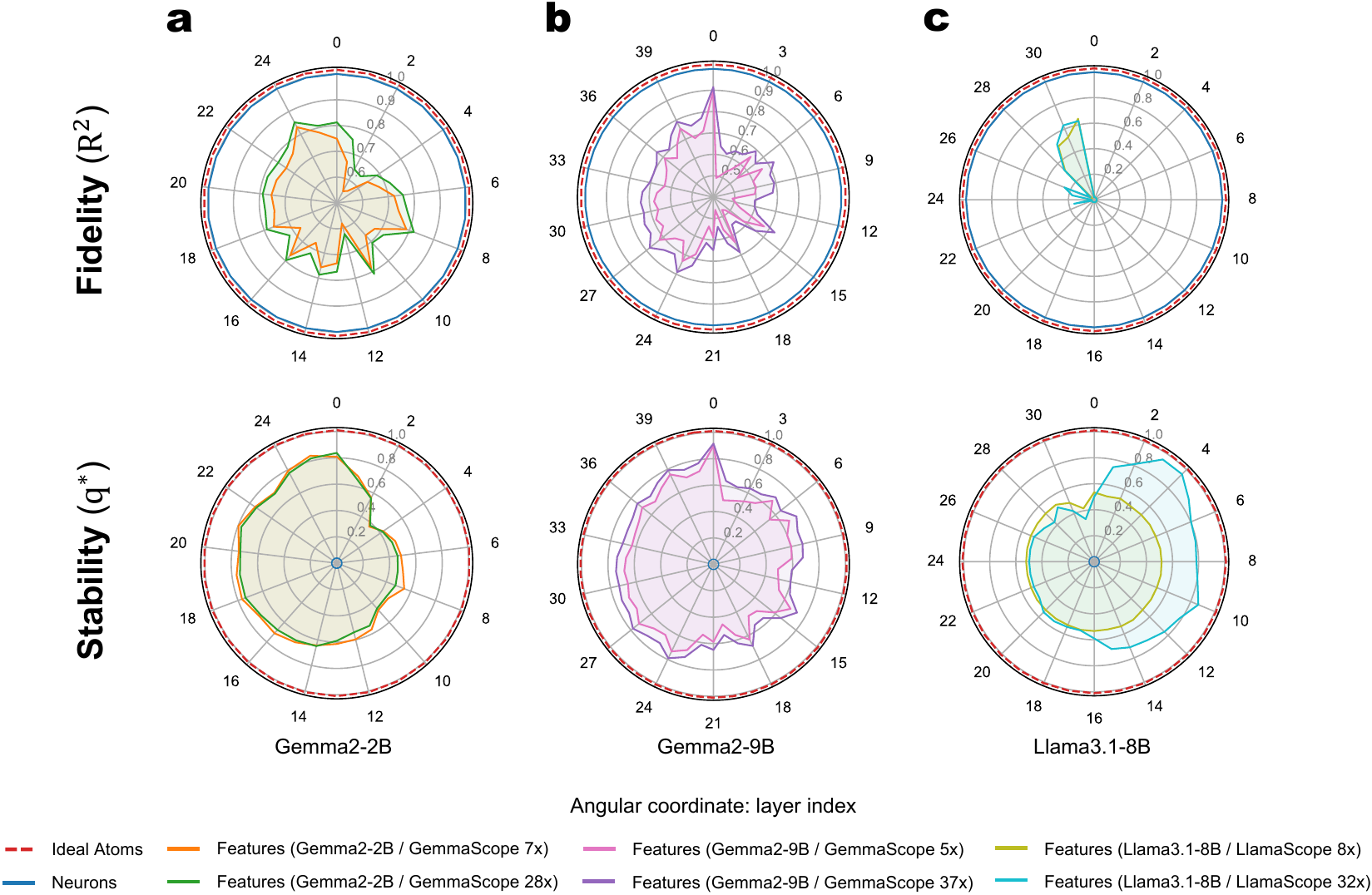}
    \caption{\textbf{Existing representational units exhibit distinct limitations in fidelity and stability.} \textbf{a–c}, Fidelity (\(R^2\), top) and stability (\(q^*\), bottom) of neurons and learned features across layers of Gemma2-2B (\textbf{a}), Gemma2-9B (\textbf{b}) and Llama3.1-8B (\textbf{c}). Angular coordinates denote layer indices, and radial coordinates denote the corresponding metric values. The dashed red circle marks the ideal-atom reference, for which both fidelity and stability attain their maximum value of 1. Neurons consistently exhibit high fidelity but substantially lower stability, whereas learned features from Gemma Scope and Llama Scope show improved stability to varying degrees but generally sacrifice fidelity, with neither class simultaneously approaching the ideal reference on both criteria. Across model scales and feature widths, these results reveal a systematic gap between existing candidate FRUs and the criteria for ideal representational atoms.}
  \label{fig:Fig_2_Radar_Chart}
\end{figure}

Having defined fidelity and stability, we next use these criteria to evaluate two widely studied classes of candidate FRUs: neurons and learned features \cite{chen2025knowledge,huben2024sparse,bricken2023monosemanticity,templeton2024scaling}. Neurons correspond directly to the native coordinates of internal representations, whereas learned features are obtained by decomposing these representations into latent components. For learned features, we use Gemma Scope \cite{lieberum-etal-2024-gemma} and Llama Scope \cite{he2024llama}, both trained on large-scale general-domain corpora (see `Evaluation of existing representational units' in Methods). The results are shown in Fig.~\ref{fig:Fig_2_Radar_Chart}.

In terms of fidelity, neurons and learned features exhibit a clear contrast. Because internal representations are expressed directly in the original neuronal coordinate system, treating neurons as candidate FRUs preserves the representational information without introducing an additional approximation, making fidelity rarely a primary limitation. Learned features, by contrast, are obtained through learned decompositions of internal representations, and their reconstruction quality depends on model architecture and scale, training data and optimization settings. Pretrained learned-feature models such as Gemma Scope and Llama Scope may therefore not be fully optimized for the target representation distribution, potentially leading to information loss and lower fidelity than neurons, even when yielding more interpretable components \cite{huben2024sparse,bricken2023monosemanticity,templeton2024scaling}. 

In terms of stability, neither neurons nor learned features fully satisfy the distinguishability required of ideal representational atoms. Neurons correspond directly to the coordinate directions of the internal representation space, but these directions need not be sufficiently distinguishable from one another. Consequently, neurons generally exhibit low stability, consistent with the widespread polysemanticity of individual neurons \cite{olah2020zoom}. Learned features reduce representational entanglement through learned decomposition and exhibit higher stability than neurons, consistent with their greater monosemanticity \cite{bricken2023monosemanticity,templeton2024scaling}. However, their stability remains substantially below the ideal regime.

Overall, neurons and learned features exhibit distinct limitations: neurons preserve high fidelity but exhibit low stability, whereas learned features improve stability but exhibit reduced fidelity and remain below the ideal stability regime. Accordingly, neither fully satisfies the requirements of ideal representational atoms.

\begin{table*}[t]
\centering
\caption{\textbf{Evaluation of sparse autoencoder representations across datasets and models.}}
\label{tab:sae_evaluation_ci95}

\scriptsize
\setlength{\tabcolsep}{2.5pt}
\renewcommand{\arraystretch}{1.08}

\resizebox{\textwidth}{!}{
\begin{tabular}{@{}lllccccccc@{}}
\toprule
\multirow{2}{*}{Dataset} & \multirow{2}{*}{Model} & \multirow{2}{*}{Metric} & \multicolumn{2}{c}{GemmaScope} & \multicolumn{5}{c}{Trained SAEs} \\
\cmidrule(lr){4-5}
\cmidrule(lr){6-10}
& & & Small & Large & ReLU & TopK & AbsTopK & JumpReLU & BiJumpReLU \\
\midrule

% =========================================================
% Wikipedia
% =========================================================

\multirow{6}{*}{Wikipedia}
& \multirow{2}{*}{Gemma2-2B}
& Fidelity $\uparrow$ & $72.17\,{\scriptstyle(\pm 13.21)}$ & $75.52\,{\scriptstyle(\pm 13.12)}$ & $75.46\,{\scriptstyle(\pm 16.17)}$ & $88.68\,{\scriptstyle(\pm 7.05)}$ & $88.11\,{\scriptstyle(\pm 6.93)}$ & $94.91\,{\scriptstyle(\pm 2.38)}$ & $\mathbf{97.25\,{\scriptstyle(\pm 1.44)}}$ \\
&
& Stability $\uparrow$ & $67.87\,{\scriptstyle(\pm 16.69)}$ & $67.43\,{\scriptstyle(\pm 18.43)}$ & $85.58\,{\scriptstyle(\pm 7.40)}$ & $39.34\,{\scriptstyle(\pm 1.01)}$ & $39.07\,{\scriptstyle(\pm 1.02)}$ & $98.41\,{\scriptstyle(\pm 1.57)}$ & $\mathbf{99.78\,{\scriptstyle(\pm 0.13)}}$ \\
\addlinespace[2pt]
&
\multirow{2}{*}{Gemma2-9B}
& Fidelity $\uparrow$ & $65.56\,{\scriptstyle(\pm 17.21)}$ & $70.78\,{\scriptstyle(\pm 15.96)}$ & $74.25\,{\scriptstyle(\pm 22.06)}$ & $87.42\,{\scriptstyle(\pm 7.56)}$ & $87.65\,{\scriptstyle(\pm 7.05)}$ & $93.38\,{\scriptstyle(\pm 3.24)}$ & $\mathbf{96.75\,{\scriptstyle(\pm 1.86)}}$ \\
&
& Stability $\uparrow$ & $91.93\,{\scriptstyle(\pm 5.17)}$ & $89.00\,{\scriptstyle(\pm 8.03)}$ & $89.46\,{\scriptstyle(\pm 4.57)}$ & $49.25\,{\scriptstyle(\pm 1.81)}$ & $48.12\,{\scriptstyle(\pm 0.70)}$ & $97.92\,{\scriptstyle(\pm 1.43)}$ & $\mathbf{99.81\,{\scriptstyle(\pm 0.08)}}$ \\
\addlinespace[2pt]
&
\multirow{2}{*}{Qwen3.5-9B}
& Fidelity $\uparrow$ & -- & -- & $77.37\,{\scriptstyle(\pm 7.32)}$ & $87.13\,{\scriptstyle(\pm 6.34)}$ & $86.43\,{\scriptstyle(\pm 6.33)}$ & $94.47\,{\scriptstyle(\pm 4.52)}$ & $\mathbf{95.66\,{\scriptstyle(\pm 2.14)}}$ \\
&
& Stability $\uparrow$ & -- & -- & $82.84\,{\scriptstyle(\pm 6.25)}$ & $51.11\,{\scriptstyle(\pm 1.20)}$ & $50.73\,{\scriptstyle(\pm 1.03)}$ & $82.74\,{\scriptstyle(\pm 14.20)}$ & $\mathbf{99.35\,{\scriptstyle(\pm 0.57)}}$ \\
\midrule

% =========================================================
% Wikidata
% =========================================================

\multirow{6}{*}{Wikidata}
& \multirow{2}{*}{Gemma2-2B}
& Fidelity $\uparrow$ & $72.44\,{\scriptstyle(\pm 6.51)}$ & $76.50\,{\scriptstyle(\pm 5.75)}$ & $87.42\,{\scriptstyle(\pm 5.58)}$ & $92.15\,{\scriptstyle(\pm 4.33)}$ & $92.03\,{\scriptstyle(\pm 3.86)}$ & $93.78\,{\scriptstyle(\pm 2.97)}$ & $\mathbf{95.21\,{\scriptstyle(\pm 3.96)}}$ \\
&
& Stability $\uparrow$ & $70.15\,{\scriptstyle(\pm 16.62)}$ & $69.21\,{\scriptstyle(\pm 17.35)}$ & $84.77\,{\scriptstyle(\pm 6.95)}$ & $42.06\,{\scriptstyle(\pm 3.97)}$ & $40.95\,{\scriptstyle(\pm 2.38)}$ & $96.97\,{\scriptstyle(\pm 2.21)}$ & $\mathbf{98.32\,{\scriptstyle(\pm 2.00)}}$ \\
\addlinespace[2pt]
&
\multirow{2}{*}{Gemma2-9B}
& Fidelity $\uparrow$ & $67.81\,{\scriptstyle(\pm 14.09)}$ & $73.76\,{\scriptstyle(\pm 12.01)}$ & $86.72\,{\scriptstyle(\pm 9.09)}$ & $90.57\,{\scriptstyle(\pm 6.68)}$ & $91.19\,{\scriptstyle(\pm 6.35)}$ & $92.52\,{\scriptstyle(\pm 4.49)}$ & $\mathbf{96.85\,{\scriptstyle(\pm 1.39)}}$ \\
&
& Stability $\uparrow$ & $92.47\,{\scriptstyle(\pm 9.33)}$ & $89.53\,{\scriptstyle(\pm 13.16)}$ & $86.18\,{\scriptstyle(\pm 9.03)}$ & $53.69\,{\scriptstyle(\pm 4.78)}$ & $49.90\,{\scriptstyle(\pm 1.41)}$ & $93.70\,{\scriptstyle(\pm 6.48)}$ & $\mathbf{99.11\,{\scriptstyle(\pm 1.02)}}$ \\
\addlinespace[2pt]
&
\multirow{2}{*}{Olmo-Hybrid-7B}
& Fidelity $\uparrow$ & -- & -- & $80.04\,{\scriptstyle(\pm 12.38)}$ & $86.18\,{\scriptstyle(\pm 9.35)}$ & $86.60\,{\scriptstyle(\pm 8.52)}$ & $90.29\,{\scriptstyle(\pm 6.87)}$ & $\mathbf{96.61\,{\scriptstyle(\pm 1.70)}}$ \\
&
& Stability $\uparrow$ & -- & -- & $79.04\,{\scriptstyle(\pm 12.59)}$ & $46.80\,{\scriptstyle(\pm 8.05)}$ & $49.31\,{\scriptstyle(\pm 1.24)}$ & $90.04\,{\scriptstyle(\pm 7.57)}$ & $\mathbf{98.56\,{\scriptstyle(\pm 1.62)}}$ \\
\midrule

% =========================================================
% LAION-COCO
% =========================================================

\multirow{4}{*}{LAION-COCO}
& \multirow{2}{*}{Stable Diffusion v1.5}
& Fidelity $\uparrow$ & -- & -- & $63.77\,{\scriptstyle(\pm 8.90)}$ & $81.04\,{\scriptstyle(\pm 18.14)}$ & $80.06\,{\scriptstyle(\pm 16.34)}$ & $75.22\,{\scriptstyle(\pm 22.51)}$ & $\mathbf{85.50\,{\scriptstyle(\pm 21.18)}}$ \\
&
& Stability $\uparrow$ & -- & -- & $72.94\,{\scriptstyle(\pm 14.41)}$ & $29.71\,{\scriptstyle(\pm 0.49)}$ & $29.44\,{\scriptstyle(\pm 0.40)}$ & $82.86\,{\scriptstyle(\pm 28.67)}$ & $\mathbf{98.52\,{\scriptstyle(\pm 4.14)}}$ \\
\addlinespace[2pt]
&
\multirow{2}{*}{FLUX.2-klein-9B}
& Fidelity $\uparrow$ & -- & -- & $80.61\,{\scriptstyle(\pm 11.97)}$ & $88.33\,{\scriptstyle(\pm 8.43)}$ & $90.05\,{\scriptstyle(\pm 9.11)}$ & $87.43\,{\scriptstyle(\pm 7.18)}$ & $\mathbf{96.03\,{\scriptstyle(\pm 2.37)}}$ \\
&
& Stability $\uparrow$ & -- & -- & $81.83\,{\scriptstyle(\pm 18.80)}$ & $41.56\,{\scriptstyle(\pm 14.95)}$ & $53.56\,{\scriptstyle(\pm 12.21)}$ & $64.12\,{\scriptstyle(\pm 10.20)}$ & $\mathbf{98.47\,{\scriptstyle(\pm 2.48)}}$ \\
\bottomrule
\end{tabular}
}
\vspace{2pt}

\begin{minipage}{\textwidth}
\footnotesize
Values are reported as mean followed by the half-width of the two-sided 95\% \(t\)-confidence interval in parentheses, and all values are scaled by 100. Higher values indicate better performance. \textbf{Bold} indicates the highest mean value in each row. For Wikipedia and Wikidata, confidence intervals are computed across six evaluated model layers. For LAION-COCO, confidence intervals are computed across three diffusion timesteps for Stable Diffusion v1.5 and across six layer-step combinations for FLUX.2-klein-9B. GemmaScope Small denotes width 16k; GemmaScope Large denotes width 65k for Gemma2-2B and width 131k for Gemma2-9B. TopK and AbsTopK use \(K=50\), and additional TopK results with \(K=20\) for Gemma2-2B and Gemma2-9B are reported in Extended Data Table~\ref{tab:topk_k_ablation}. Dashes indicate that the corresponding GemmaScope model is unavailable.
\end{minipage}

\end{table*}

\begin{figure}
    \centering
    \includegraphics[width=0.98\textwidth]{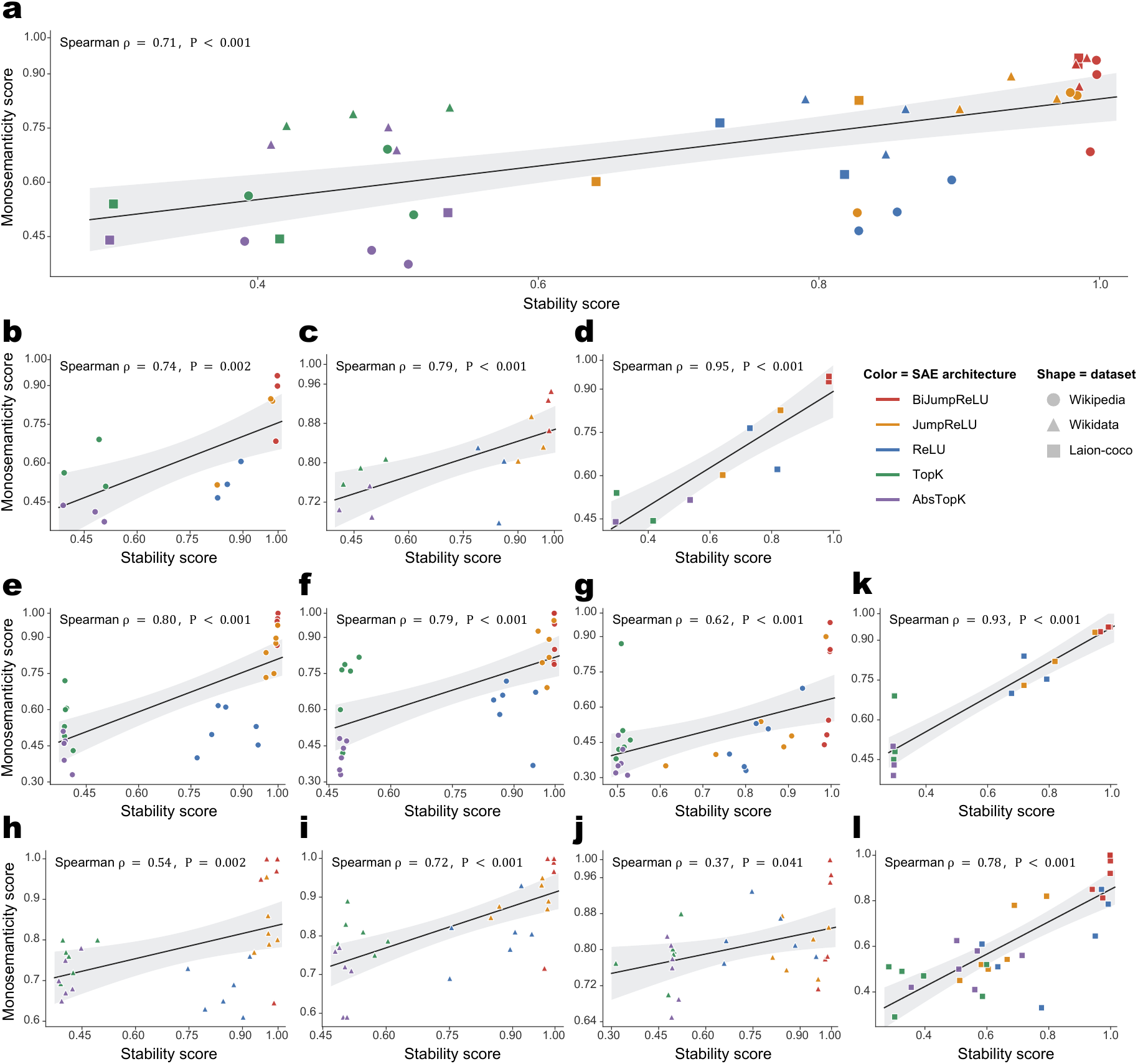}
    \caption{\textbf{Monosemanticity is positively correlated with stability across datasets and models.} \textbf{a}, Overall relationship between stability and monosemanticity across all datasets, models and SAE architectures. Each point denotes one dataset-model-SAE combination (\(N=40\)). For language models, stability and monosemanticity are averaged across the six evaluated layers; for Stable Diffusion v1.5, across three diffusion timesteps; and for FLUX.2-klein-9B, across six layer-step combinations. The positive correlation indicates a consistent association between stability and monosemanticity across heterogeneous representation settings. \textbf{b-d}, Relationships within Wikipedia (\textbf{b}; \(N=15\)), Wikidata (\textbf{c}; \(N=15\)) and LAION-COCO (\textbf{d}; \(N=10\)), using the same model-SAE-level averages as in \textbf{a}. Positive correlations within all three datasets show that the overall association is preserved within individual datasets and modalities. \textbf{e-l}, Relationships within individual models: Gemma2-2B (\textbf{e}), Gemma2-9B (\textbf{f}) and Qwen3.5-9B (\textbf{g}) on Wikipedia; Gemma2-2B (\textbf{h}), Gemma2-9B (\textbf{i}) and Olmo-Hybrid-7B (\textbf{j}) on Wikidata; and Stable Diffusion v1.5 (\textbf{k}) and FLUX.2-klein-9B (\textbf{l}) on LAION-COCO. For language models, each point denotes one SAE architecture at one layer; for Stable Diffusion v1.5, one SAE architecture at one diffusion timestep; and for FLUX.2-klein-9B, one SAE architecture at one layer-step combination. Monosemanticity is averaged across the manually evaluated features at each corresponding layer, timestep or layer-step combination. These model-level results show that the positive correlation persists at finer representational granularity, although its strength varies across models. Colors denote SAE architectures (BiJumpReLU, JumpReLU, ReLU, TopK with \(K=50\), and AbsTopK with \(K=50\)), and marker shapes denote datasets. Black lines show linear regression fits with 95\% confidence bands. Spearman's rank correlation coefficients (\(\rho\)) and two-sided \(P\) values are shown in each panel.}
  \label{fig:Fig_3_Stability_Monosemanticity_Correlation}
\end{figure}

\subsection*{Threshold-activated SAEs identify representational atoms}\label{Section:Threshold-activated SAEs identify representational atoms}

The limitations of existing candidate FRUs raise a central question: what kind of representation learning mechanism can reliably identify representational atoms with both high fidelity and stability?

Our theoretical analysis shows that, under appropriate conditions, threshold-activated sparse autoencoders (TSAEs) can identify FRUs with both high fidelity and stability (see `Identification of atoms' in Methods; Fig.~\ref{fig:Fig_1_Overview}c). We instantiate TSAEs using BiJumpReLU SAEs, which utilize learnable activation thresholds to identify representational atoms (see `SAE architectures' in Methods). We evaluate their effectiveness across multiple model families, network architectures, and representation modalities, spanning diverse data distributions. We compare BiJumpReLU SAEs with several established SAE architectures \cite{huben2024sparse,gao2025scaling,rajamanoharan2024jumping,zhu2026abstopk} under the same evaluation framework. Full experimental settings are described in `Training details' in Methods. Table~\ref{tab:sae_evaluation_ci95} reports the main results for fidelity and stability, the two criteria used to evaluate representational atoms, with additional metrics provided in Extended Data Table~\ref{tab:sae_evaluation_ci95_full}.

Among the evaluated SAE architectures, BiJumpReLU SAEs consistently maintain high fidelity and stability across all settings, achieving the highest mean value for both metrics in all eight model--dataset combinations. TopK and AbsTopK SAEs retain relatively high fidelity but exhibit substantially lower stability, whereas ReLU SAEs achieve higher stability but generally lower fidelity. JumpReLU SAEs provide the strongest competing results in many settings, but their stability varies markedly across models, from 64.12 to 98.41. In contrast, BiJumpReLU SAEs maintain stability above 98 across all settings while further improving fidelity.

Together, these theoretical and empirical results show that TSAEs provide an effective approach for identifying representational atoms. BiJumpReLU SAEs serve as a concrete instantiation of this approach and perform consistently across diverse models, architectures, data distributions and representation modalities, providing a practical basis for identifying atoms in neural representations.

\subsection*{Stable atoms are more monosemantic}\label{Section:Stable atoms are more monosemantic}

Monosemanticity refers to a representational unit has a consistent semantic meaning, such that its activation reliably corresponds to a specific semantic pattern \cite{huben2024sparse,bricken2023monosemanticity,templeton2024scaling}. Such consistency is necessary for a well-defined interpretation of the unit, as its semantic attribution otherwise remains ambiguous. Monosemanticity therefore provides a basis for analyzing the semantic organization of internal representations.

Our theoretical analysis shows that stability characterizes the distinguishability required for monorepresentationality, which constitutes a necessary structural condition for monosemanticity (see `Evaluation of atoms' in Methods). However, monorepresentationality is not sufficient, because a structurally well-defined unit may still encode distinctions that do not align with human semantic categories. The relationship between stability and monosemanticity therefore requires empirical validation.

To test this relationship, we evaluate the monosemanticity of the identified FRUs and examine its association with stability (see `Monosemanticity evaluation details' in Methods). Across all 40 dataset--model--SAE combinations, stability is positively correlated with monosemanticity (Spearman \(\rho=0.71\), \(P<0.001\); Fig.~\ref{fig:Fig_3_Stability_Monosemanticity_Correlation}a). This association is preserved within each dataset, with correlations of \(\rho=0.74\) for Wikipedia, \(\rho=0.79\) for Wikidata and \(\rho=0.95\) for LAION-COCO (Fig.~\ref{fig:Fig_3_Stability_Monosemanticity_Correlation}b--d). Across individual models and representation settings, all eight correlations remain positive, ranging from \(\rho=0.37\) to \(0.93\) (Fig.~\ref{fig:Fig_3_Stability_Monosemanticity_Correlation}e--l). Architecture-stratified correlations and monosemanticity distributions across SAE architectures are reported in Extended Data Figs.~\ref{fig:Extended_Fig_1} and \ref{fig:Extended_Fig_2}. Together, these results show that greater stability is consistently associated with greater monosemanticity across model families, representation modalities and SAE architectures.

\begin{figure}
    \centering
    \includegraphics[width=0.98\textwidth]{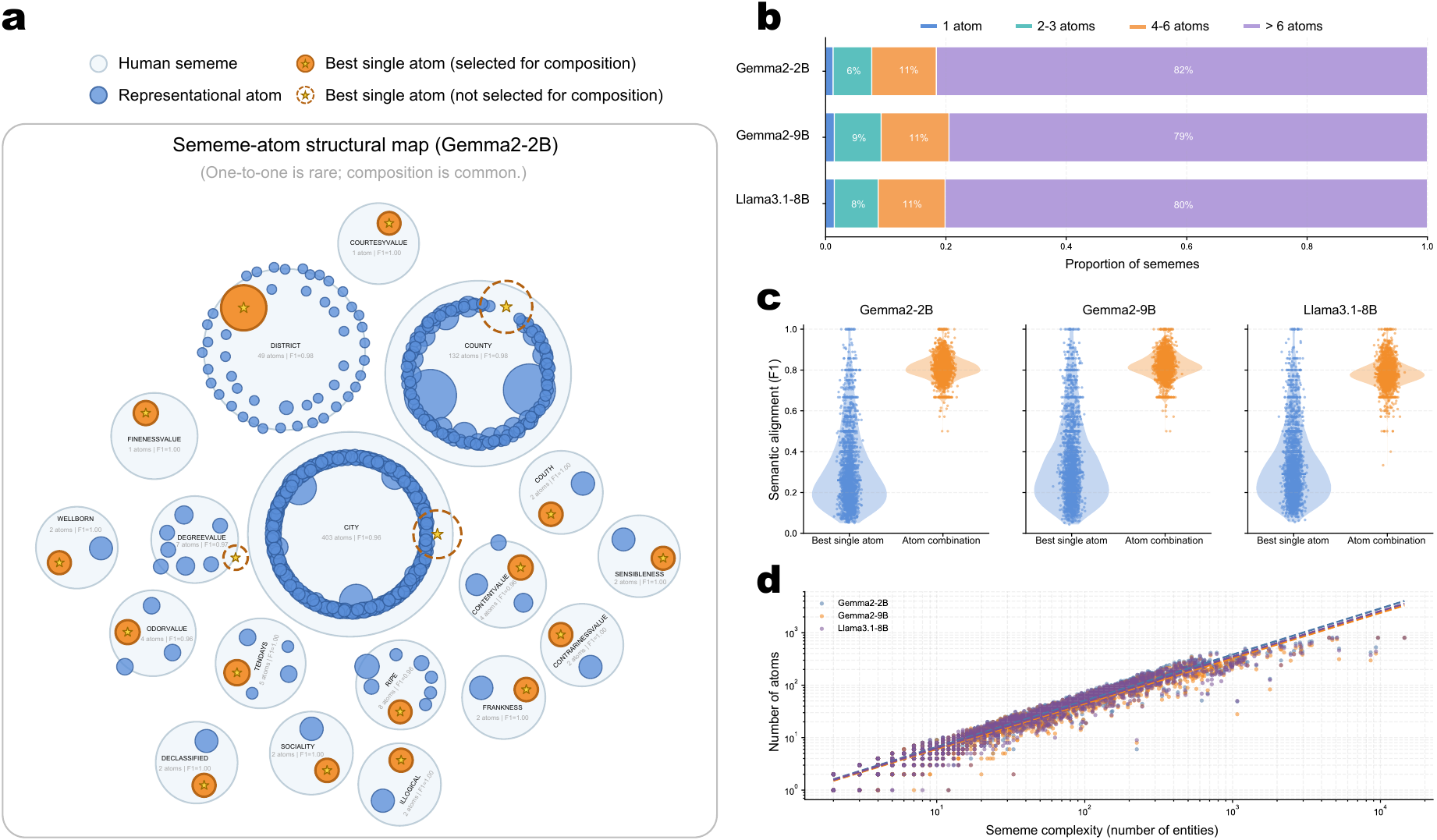}
    \caption{\textbf{Human sememes are compositionally represented by multiple representational atoms.} \textbf{a}, Sememe--atom structural map for Gemma2-2B. Large light-blue circles denote human sememes and smaller blue circles denote representational atoms selected for their compositional recovery. Solid orange circles indicate the best single atoms that are retained in the final composition, whereas dashed orange circles indicate the best single atoms that are not selected. Circle sizes reflect the sizes of the corresponding semantic sets, and the geometric overlap between an atom and its sememe reflects their single-atom precision. Representative examples show that near one-to-one sememe--atom correspondence is rare, whereas compositional recovery by multiple atoms is common. More complete sememe--atom structural maps for Gemma2-2B, Gemma2-9B, and Llama3.1-8B are shown in Extended Data Figs.~\ref{fig:Extended_Fig_3}--\ref{fig:Extended_Fig_5}, respectively. \textbf{b}, Distribution of the number of atoms selected for each sememe across Gemma2-2B, Gemma2-9B and Llama3.1-8B, grouped into 1, 2--3, 4--6 and \(>6\) atoms. Most sememes are recovered using more than six atoms. \textbf{c}, Semantic recovery of the best single atom and the corresponding atom combination across the three models. Each point denotes an individual sememe, and F1 measures the agreement between the recovered representational set and the corresponding HowNet sememe set. Atom combinations consistently achieve substantially higher semantic alignment than the best single atoms. \textbf{d}, Relationship between sememe complexity and the number of atoms selected for its compositional recovery across the three models. Sememe complexity is measured by the number of entities associated with each HowNet sememe. Each point denotes an individual sememe with merged F1 \(\geq 0.5\), and dashed lines show model-specific linear regression fits in log--log space. More complex sememes tend to require more atoms for recovery.}
  \label{fig:Fig_4_Sememes_Atoms_Alignment}
\end{figure}

\subsection*{Atoms compositionally encode sememe structures}\label{Section:Atoms compositionally encode sememe structures}

Having shown that representational atoms identified by BiJumpReLU SAEs exhibit higher stability and greater monosemanticity, we next examine how these atoms relate to human conceptual structure. Monosemanticity captures the consistency of an FRU's semantic interpretation, but does not determine whether that FRU alone captures a complete human semantic concept. We therefore investigate whether higher-order semantic structures are encoded by single atoms or by their compositions.

We use HowNet \cite{qi2019openhownet, Dong2003HowNetA}, an expert-annotated knowledge base in which concepts are described by fine-grained semantic units called sememes. We extract internal representations of HowNet entities across layers from multiple models and identify corresponding representational atoms using BiJumpReLU SAEs (see `Representation extraction for cross-layer semantic alignment' in Methods). These SAEs maintain high fidelity and stability across the evaluated models and layers (Extended Data Table~\ref{tab:hownet_sae_evaluation_mean}). For each sememe, we identify the single atom with the highest set-matching F1 score across layers, where F1 summarizes the precision and recall of the atom--sememe match, and then search for the cross-layer atom combination that maximizes the same score (see `Semantic alignment evaluation details' in Methods).

As shown in Fig.~\ref{fig:Fig_4_Sememes_Atoms_Alignment}, near one-to-one correspondences between sememes and atoms are rare. Although some sememes can be recovered almost completely by a single atom, with F1 reaching 1 (Fig.~\ref{fig:Fig_4_Sememes_Atoms_Alignment}a), such cases account for only approximately 1\% of all sememes (Fig.~\ref{fig:Fig_4_Sememes_Atoms_Alignment}b). Most sememes require multiple atoms, with the majority involving more than six atoms across all three models (Fig.~\ref{fig:Fig_4_Sememes_Atoms_Alignment}b). These atom combinations achieve substantially higher semantic alignment (F1) than the corresponding best single atoms (Fig.~\ref{fig:Fig_4_Sememes_Atoms_Alignment}c), primarily through higher recall without sacrificing precision (Extended Data Table~\ref{tab:sememe_alignment_prf_ci95}). The number of selected atoms also increases systematically with sememe complexity, measured by the number of associated entities (Fig.~\ref{fig:Fig_4_Sememes_Atoms_Alignment}d). These results suggest that representational atoms generally function as finer-grained semantic components whose compositions encode higher-order human semantic structures.

\begin{figure}
    \centering
    \includegraphics[width=0.98\textwidth]{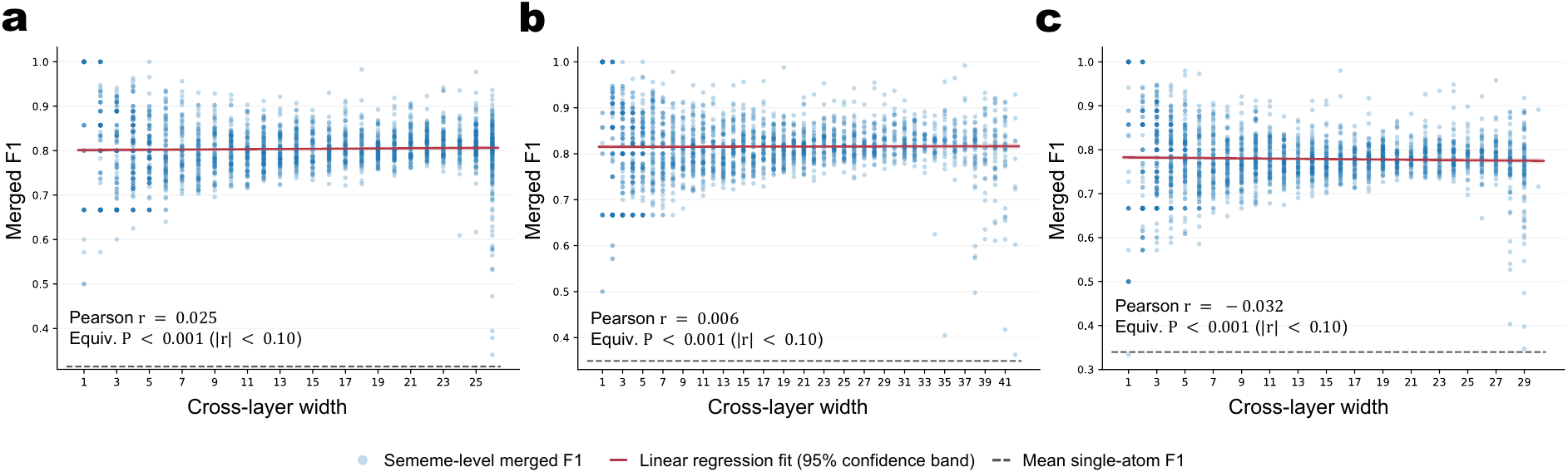}
    \caption{\textbf{Sememe-level merged F1 is largely insensitive to cross-layer width.} \textbf{a–c}, Relationship between cross-layer width and sememe-level merged F1 for Gemma2-2B (\textbf{a}), Gemma2-9B (\textbf{b}) and Llama3.1-8B (\textbf{c}). Each blue point denotes the merged F1 obtained for an individual sememe at corresponding cross-layer width. Red lines show linear regression fits with 95\% confidence bands, and dashed grey lines indicate the mean single-atom F1 baseline. Across all three models, the fitted slopes are negligible and Pearson correlations remain close to zero (\(r=0.025\), \(0.006\) and \(-0.032\), respectively). Equivalence tests further show that the absolute correlations fall within the prespecified bound of \(|r|<0.10\) (\(P<0.001\) in all cases), indicating that sememe-level performance is not systematically associated with cross-layer width.}
  \label{fig:Fig_5_Merged_F1}
\end{figure}

To determine whether this compositional advantage could arise from the cross-layer search procedure, we examine the association between cross-layer width and final F1. The Pearson correlation between them is close to zero, and the equivalence test \cite{lakens2017equivalence} confirms that it falls within the prespecified bound of \(|r|<0.10\) (\(P<0.001\); Fig.~\ref{fig:Fig_5_Merged_F1}), indicating no practically meaningful association. Final F1 therefore does not increase systematically with cross-layer width, supporting the interpretation that the advantage of atom combinations reflects compositional structure in the internal representations rather than the search procedure.

Case-study analysis further reveals discrepancies between the entity sets recovered by atom combinations and the corresponding HowNet annotations (Extended Data Table~\ref{tab:sememe_case_study}; see `Sememe evaluation details' in Methods). Manual inspection identifies unannotated entities that are consistent with the target sememes, as well as annotated entities that appear inconsistent with them. These cases suggest that some apparent mismatches may reflect incomplete or inaccurate annotations. Together, these findings show that single atoms capture finer-grained semantic components, while their compositions recover higher-order semantic organization that closely corresponds to human conceptual structure.

\section*{Discussion}\label{Section:Discussion}

This study focuses on a fundamental question concerning the basic units that constitute internal representations of neural networks and how their organization relates to human conceptual structure. We formulate this problem in terms of the definition, evaluation, and identification of fundamental representational units (FRUs). Drawing on compressed sensing \cite{donoho2006compressed,candes2006robust}, we define FRUs as representational atoms and introduce fidelity and stability to characterize information preservation and structural distinguishability. We further show that threshold-activated sparse autoencoders (TSAEs) can identify such atoms under appropriate conditions, instantiate TSAEs as BiJumpReLU SAEs in practice, and validate their effectiveness across diverse models and representation settings. Further analysis indicates that more stable atoms also exhibit greater empirical monosemanticity, and their cross-layer compositions recover higher-order semantic organization corresponding to human conceptual structure. Together, these findings provide a unified theoretical and practical framework for characterizing the basic organization of neural representations and its relationship to human conceptual structure.

This framework provides a principled view of commonly studied representational units. Neurons and learned features can be evaluated within the same theoretical setting through fidelity and stability, rather than being treated solely as empirical units of interpretation. More broadly, the framework connects two influential perspectives on neural representation: the linear representation hypothesis \cite{park2023linear}, in which semantic information is encoded along specific directions, and the superposition hypothesis \cite{elhage2022toy}, which explains how finite-dimensional spaces can support substantially more linear components than their dimensionality. Their combination within a sparse linear framework provides a unified account of how representational components can remain both expressive and distinguishable.

The compositional organization of representational atoms also has implications for interventions on internal representations, including knowledge editing \cite{meng2022locating,meng2023mass,hu2024wilke} and representation steering \cite{rimsky2024steering,rodriguez2025controlling}. Such methods often rely on mapping human-defined target concepts to localizable representational structures. Our results suggest that these correspondences are often mediated by finer-grained representational atoms and their combinations, which jointly form higher-order semantic structures. Intervening on a single direction or localized representation may therefore affect only part of a target concept or simultaneously perturb multiple overlapping semantic components. This perspective may help explain incomplete edits, collateral effects and instability, and suggests that future intervention methods should explicitly consider the compositional and superpositional organization of FRUs.

This representational perspective is also relevant to continual learning \cite{kirkpatrick2017overcoming,van2022three}. Continual learning modifies internal representations through parameter updates, and superposition implies that changes to representations may perturb multiple overlapping components and thereby affect existing knowledge. This provides a possible structural account of catastrophic forgetting and suggests that robust continual learning may therefore benefit from explicitly considering the superpositional organization of FRUs. More broadly, this framework provides a basis for studying how models integrate new information while preserving previously encoded knowledge.

A natural question is whether this framework generalizes across diverse neural networks and internal representations. Its generality ultimately depends on whether stable and semantically meaningful representational atoms emerge across different models and representation spaces. Importantly, fidelity and stability provide quantitative criteria for assessing this applicability without assuming it a priori. Across the models and representation settings examined here, we observe consistent evidence for representational atoms, supporting their use as a testable and extensible basis for studying neural representations and their semantic organization.

\section*{Methods}\label{Section:Methods}

\subsection*{Representational atoms theoretical framework}\label{Section:Representational atoms theoretical framework}

In neural networks, all information is embedded in high-dimensional representations. Our objective is to identify the fundamental representational units (FRUs), which we term representational atoms, or atoms for brevity. Formally, we consider a collection of representations \(M = \{\boldsymbol{m}_i\}_{i=1}^{|M|}\), where \(\boldsymbol{m}_i \in \mathbb{R}^H\). Each representation can be expressed as \(\boldsymbol{m}_i=\sum_{j}\delta(i,j)\boldsymbol d_j\), where \(\delta(i,j)\) denotes the presence and magnitude of the \(j\)-th representational unit \(\boldsymbol d_j\in\mathbb{R}^H\) in the \(i\)-th representation \(\boldsymbol{m}_i\). However, in representation space, the family of representational units admitting such a decomposition is, in principle, infinite.

The central question is how to define FRUs. A natural criterion is distinguishability: each atom should be detectable or manipulable without interfering with others. In high-dimensional spaces, this criterion translates into orthogonality: atoms occupy mutually orthogonal directions, making their identities distinguishable via inner products. This motivates introducing an inner product to analyze the geometry of FRUs.

Thus, the choice of inner product is critical. Although the Euclidean inner product is commonly used, it is not necessarily appropriate for characterizing the intrinsic geometry of neural representation spaces \cite{park2023linear, hu2025knowledge}. To illustrate this point, let \(\boldsymbol{h}\in\mathbb{R}^{H}\) denote the internal representation at a given layer and \(f_{\mathrm{down}}\) the subsequent network mapping, so that the downstream computation is \(\boldsymbol{y}=f_{\mathrm{down}}(\boldsymbol{h})\). Because our analysis concerns the linear structure and its geometry in the representation space, we consider reparameterizations that preserve this linear structure. In a finite-dimensional real vector space, such reparameterizations are exactly invertible linear transformations, so that any bijection that preserves linearity can be uniquely written as \(\boldsymbol{h}'=\boldsymbol{A}\boldsymbol{h}\), where \(\boldsymbol{A}\in\mathbb{R}^{H\times H}\) is invertible. Accordingly, for any invertible linear transformation \(\boldsymbol{A}\), we can introduce the following reparameterization:
\begin{equation}\label{Equation:Reparameterization}
    f'_{\mathrm{down}}(\boldsymbol{z})\leftarrow f_{\mathrm{down}}(\boldsymbol{A}^{-1}\boldsymbol{z}),\quad \boldsymbol{h}'\leftarrow\boldsymbol{A}\boldsymbol{h}.
\end{equation}
Then it follows that \(f'_{\mathrm{down}}(\boldsymbol{h}')=f_{\mathrm{down}}(\boldsymbol{h})\), leaving the input–output function unchanged. The results of Park et al. \cite{park2023linear} and Hu et al. \cite{hu2025knowledge} can be viewed as specific instances of this reparameterization principle under particular model parameterizations.

Since the above reparameterization leaves the network input–output function unchanged, \(\boldsymbol{h}\) and \(\boldsymbol{h}'=\boldsymbol{A}\boldsymbol{h}\) are indistinguishable from the perspective of observable network behavior. Thus, even for a fully trained network with a fixed input–output function, internal representations are identifiable only up to the equivalence class induced by invertible linear transformations \(\boldsymbol{A}\), rather than by unique coordinate representations determined by the network itself. By linearity, this non-uniqueness extends to the corresponding representational units \(\boldsymbol{d}_j\). Under transformation \(\boldsymbol{d}'_j=\boldsymbol{A}\boldsymbol{d}_j\), the transformed units describe an equivalent representational structure associated with the same network function.

This has direct consequences for the geometry of the representation space. In particular, the Euclidean inner product is not invariant under the reparameterization in equation \eqref{Equation:Reparameterization}: \(\langle \boldsymbol{d}_i,\boldsymbol{d}_j\rangle\neq\langle \boldsymbol{A}\boldsymbol{d}_i,\boldsymbol{A}\boldsymbol{d}_j\rangle\). Therefore, geometric relations induced by the Euclidean inner product, such as angles and orthogonality, between \(\boldsymbol d_i\) and \(\boldsymbol d_j\) depend on the chosen parameterization. Thus, the Euclidean inner product does not provide a canonical geometry of neural representations that is independent of the chosen parameterization.

\bmhead{Atomic inner product}\label{Section:Atomic inner product}

To better understand neural representations and thereby define atoms within them, we require additional principles to specify an appropriate inner product. We therefore introduce an inner product with the desired properties.

\begin{definition}[Atomic inner product; AIP]
    Let \(\mathcal D\!=\!\mathrm{span}(D)\), where \(D\!=\!\{\boldsymbol d_j\}_{j=1}^{|D|}\) denotes the atom set. The atomic inner product \(\langle \cdot, \cdot \rangle_S\) is an inner product on \(\mathcal D\) such that \(\langle \boldsymbol{d}_i, \boldsymbol{d}_j \rangle_S\! =\! 0\) for all \(\boldsymbol{d}_i, \boldsymbol{d}_j \in D\) with \(i \neq j\).
\end{definition}

Atoms are indexed arbitrarily, and any permutation of indices leaves their geometry invariant. Hence, there is no principled basis for assigning different norms to different atoms. By this symmetry, we assume a common norm under the chosen inner product, i.e. \(\|\boldsymbol d_i\|_S \!=\! c>0\), \(\forall i \in [|D|]\). The constant \(c\) cancels naturally in subsequent analysis.

By abuse of notation, we also use \(\boldsymbol D \in \mathbb{R}^{H \times |D|}\) to denote the matrix with columns \(\boldsymbol d_j\). We next characterize the atomic inner product in an explicit form.

\begin{restatable}[Explicit form of the atomic inner product]{theorem}{ExplicitForm}\label{theorem:Explicit_Form}
    Let \(\langle \boldsymbol{d}_i, \boldsymbol{d}_j \rangle_S = \boldsymbol{d}_i^\top \boldsymbol S \boldsymbol{d}_j\) be an atomic inner product with \(\boldsymbol S\in \mathbb{R}^{H\times H}\) symmetric and positive definite. If the columns of \(\boldsymbol D = [\boldsymbol{d}_1, \boldsymbol{d}_2, \cdots, \boldsymbol{d}_{|D|}]\) form the atom set such that \(\forall i,\ \|\boldsymbol{d}_i\|_S = c>0\), and \(\mathcal{D} \simeq \mathbb{R}^H\), then \(\boldsymbol S = c^2 (\boldsymbol D\boldsymbol D^\top)^{-1}\).
\end{restatable}

All proofs are provided in the `Proofs' section of the Supplementary Information. By analogy to cosine similarity, we introduce the normalized atomic inner product to remove the dependence on \(c\).

\begin{restatable}[Normalized atomic inner product; NAIP]{corollary}{NAIP}\label{corollary:NAIP}
    Let the atomic inner product be defined by \(\langle \boldsymbol{d}_i, \boldsymbol{d}_j \rangle_S = \boldsymbol{d}_i^\top \boldsymbol S \boldsymbol{d}_j\), where \(\boldsymbol S\) is symmetric and positive definite. Suppose the columns of \(\boldsymbol D = [\boldsymbol{d}_1, \cdots, \boldsymbol{d}_{|D|}]\) form the atom set satisfying \(\forall i ,\ \|\boldsymbol{d}_i\|_S = c>0\). Then, for any \(i, j\),
    \begin{equation}
        \langle\boldsymbol{d}_i, \boldsymbol{d}_j\rangle_{\tilde S} := \frac{\langle \boldsymbol{d}_i, \boldsymbol{d}_j \rangle_S}{\|\boldsymbol{d}_i\|_S \|\boldsymbol{d}_j\|_S} = \boldsymbol{d}_i^\top \tilde{\boldsymbol S} \boldsymbol{d}_j,\quad \tilde{\boldsymbol S} = (\boldsymbol D\boldsymbol D^\top)^{-1}.
    \end{equation}
    Consequently, the bilinear form \(\langle \boldsymbol{d}_i, \boldsymbol{d}_j \rangle_{\tilde{S}} = \boldsymbol{d}_i^\top \tilde{\boldsymbol S} \boldsymbol{d}_j\) defines the normalized atomic inner product.
\end{restatable}

Unlike the causal inner product \citep{park2023linear}, which is defined in the static and input-independent unembedding space over the output tokens, our AIP (or NAIP) operates in the dynamic and input-dependent representation space, characterizing the geometry of representations and their constituent units. This enables direct analysis of internal representations and thereby allows the definition of atoms as fundamental representational units under the AIP.

\begin{remark*}
    Define \(\tilde{\boldsymbol{d}_i}\! = \!\tilde{\boldsymbol S}^{\frac{1}{2}} \boldsymbol{d}_i\) and \(\tilde{\boldsymbol{d}_j} \!=\! \tilde{\boldsymbol S}^{\frac{1}{2}} \boldsymbol{d}_j\). Under this transformation, \(\langle \boldsymbol{d}_i, \boldsymbol{d}_j \rangle_{\tilde{S}}=\langle \tilde{\boldsymbol{d}_i}, \tilde{\boldsymbol{d}_j} \rangle\), where the right-hand side denotes the Euclidean inner product. Hence, properties of the Euclidean inner product are transferred directly to the NAIP. Consequently, \(\tilde{\boldsymbol{d}_i}\) and \(\tilde{\boldsymbol{d}_j}\) are termed normalized atoms.
\end{remark*}

\bmhead{Formal definition of atoms}\label{Section:Formal definition of atoms}

Having established a principled perspective for understanding neural representations, we proceed to formally define atoms. However, the preceding analysis assumes an idealized setting in which atoms are strictly orthogonal under the chosen inner product. Although this assumption yields a clean notion, it constrains the number of atoms to at most the representation dimension (\(|D| = H\)), rendering the formulation impractical.

Elhage et al. \cite{elhage2022toy} observed that sparsity induces the emergence of approximately orthogonal representations in neural networks to cope with limited representational dimensionality, a phenomenon termed superposition. Hu et al. \cite{hu2025knowledge} later identified pervasive superposition in neural networks. Motivated by these findings, we introduce sparsity, which naturally leads to approximate orthogonality and enables a formally grounded, practical definition of atoms.

\begin{definition}[Sparsity level]
    Let \(M \!=\! \{\boldsymbol{m}_i\}_{i=1}^{|M|} \!\subset\! \mathbb{R}^H\) be a collection of representations. Suppose there exist \(\boldsymbol D \!=\! [\boldsymbol{d}_1,\! \cdots\!, \boldsymbol{d}_{|D|}]\! \in\! \mathbb{R}^{H \times |D|}\) and \(\boldsymbol{\delta}_{i} \!\in\! \mathbb{R}^{|D|}_{\geq 0}\) such that \(\boldsymbol{m}_i\! =\! \boldsymbol D \boldsymbol{\delta}_i\), \(\forall i\!\in\![|M|]\). The sparsity level is \(K\!:=\!\max_{ i}\|\boldsymbol \delta_i\|_0\), where \(\|\cdot\|_0\) denotes the \(\ell_0\) norm.
\end{definition}

Sparsity enables the number of atoms to substantially exceed the ambient dimension, yielding an overcomplete structure with \(|D|\gg H\) that captures rich world knowledge, while simultaneously minimizing mutual interference.

\begin{remark*}
    In the basic setting described in the `Atomic inner product' subsection, with \(|D| = H\), one can verify that \(\tilde{\boldsymbol S} = (\boldsymbol D\boldsymbol D^\top)^{-1}\) satisfies \(\boldsymbol D^\top \tilde{\boldsymbol S} \boldsymbol D = \boldsymbol I_{|D|}\), so the atoms form an orthonormal basis under \(\langle \cdot, \cdot \rangle_S\). When \(|D|\!\gg\! H\), \(\tilde{\boldsymbol S} \!= (\boldsymbol D\boldsymbol D^\top)^{-1}\) is well defined provided \(\mathrm{rank}(\boldsymbol D) = H\), but \(\boldsymbol G \!:= \boldsymbol D^\top \tilde{\boldsymbol S} \boldsymbol D\) becomes a rank-\(H\) projection rather than \(\boldsymbol I_{|D|}\), thus the atoms cannot all be mutually orthogonal under \(\langle \cdot, \cdot \rangle_S\). Therefore, exact orthogonality is unattainable, motivating the introduction of approximate orthogonality.
\end{remark*}

This consideration motivates the following definition.

\begin{definition}[\(\epsilon\)-Approximately Orthogonal Atoms]\label{definition:Approximately_Orthogonal_Atoms}
    The atom set \(\{\boldsymbol d_i\}_{i=1}^{|D|}\) is \(\epsilon\)-approximately orthogonal if \(\bigl|\langle \boldsymbol d_i, \boldsymbol d_j \rangle_{\tilde S}\bigr| \le \epsilon,\ \forall i \neq j,\) where \(\langle \boldsymbol x, \boldsymbol y \rangle_{\tilde S} := \boldsymbol x^\top \tilde{\boldsymbol S} \boldsymbol y\) denotes the normalized atomic inner product and \(\tilde{\boldsymbol S} := (\boldsymbol D\boldsymbol D^\top)^{-1}\).
\end{definition}

\begin{remark*}
    In the ideal setting of exact orthogonality, the inner products \(\langle \boldsymbol d_i, \boldsymbol d_j \rangle_{\tilde S}\) (equivalently, \(\langle \tilde{\boldsymbol{d}_i}, \tilde{\boldsymbol{d}_j} \rangle\)) for \(i \neq j\) follow a Dirac measure concentrated at the origin. Under \(\epsilon\)-approximate orthogonality, they are instead expected to follow a Gaussian distribution \(\mathcal N(0, s^2)\) with small variance, which converges to the Dirac measure as \(s \to 0\).
\end{remark*}

We now introduce a formal definition of atoms.

\begin{definition}[Atoms]\label{definition:Atoms}
    Let \(M = \{ \boldsymbol{m}_i \}_{i=1}^{|M|}\) be a collection of representations. Suppose there exists \(\boldsymbol D \!= \![\boldsymbol{d}_1,\! \cdots, \!\boldsymbol{d}_{|D|}] \!\in \mathbb{R}^{H \times |D|}\) and \(\boldsymbol \Delta\!=\![\boldsymbol\delta_1,\!\cdots\!,\boldsymbol\delta_{|M|}]\!\in\mathbb R_{\ge0}^{|D|\times |M|}\) such that, for a given sparsity level \(K \in \mathbb{N}^+\),
    \begin{equation}\label{Equation:Atoms}
         \forall i\in[|M|],\quad \boldsymbol{m}_i = \boldsymbol D \boldsymbol{\delta}_i\ \text{ with }\ \|\boldsymbol{\delta}_i\|_0 \leq K.
    \end{equation}
    Furthermore, \(\bigl|\langle \boldsymbol d_i, \boldsymbol d_j \rangle_{\tilde S}\bigr|\! \le \!\epsilon,\ \forall i\! \neq\! j\), where \(\tilde{\boldsymbol S} \!:=\! (\boldsymbol D\boldsymbol D^\top)^{-1}\). Under these conditions, \(\{\boldsymbol d_i\}_{i=1}^{|D|}\) is called the atom set of \(M\), and each \(\boldsymbol d_i\) is referred to as an atom.
\end{definition}

Intuitively, atoms are characterized by three properties: representability, where each representation can be faithfully expressed by atoms, i.e., \(\boldsymbol m_i = \boldsymbol D \boldsymbol \delta_i\); sparsity, where each representation involves only a few atoms, i.e., \(\|\boldsymbol \delta_i\|_0 \le K\); and separability, where atoms are approximately orthogonal, i.e., \(|\langle \tilde{\boldsymbol d_i}, \tilde{\boldsymbol d_j} \rangle| \le \epsilon\). Representability is a natural requirement, while sparsity and separability jointly enable efficient encoding of separable information under approximate orthogonality. We further quantitatively characterize the relationship between sparsity and separability in the following `Evaluation of atoms' subsection.

\begin{remark*}
    \(K\) quantifies sparsity without enforcing a specific sparsity regime, ensuring broad applicability of the definition. Moreover, pre-multiplying both sides of equation~\eqref{Equation:Atoms} by \(\tilde{\boldsymbol S}^{\frac{1}{2}}\) yields \(\tilde{\boldsymbol{m}}_i = \tilde{\boldsymbol D} \boldsymbol{\delta}_i\), with \(\tilde{\boldsymbol{m}}_i := \tilde{\boldsymbol S}^{\frac{1}{2}} \boldsymbol{m}_i\) and \(\tilde{\boldsymbol D} := \tilde{\boldsymbol S}^{\frac{1}{2}} \boldsymbol D = [\tilde{\boldsymbol{d}}_1, \cdots, \tilde{\boldsymbol{d}}_{|D|}]\), which simplifies subsequent derivations.
\end{remark*}

\bmhead{Evaluation of atoms}\label{Section:Evaluation of atoms}

Having formalized atoms via representability, sparsity, and separability, we now evaluate whether candidate representational units satisfy the criteria of ideal atoms. Representability is measured by the coefficient of determination \(R^2\! := \!1\! -\! \frac{\sum_i \|\boldsymbol x_i - \hat{\boldsymbol x}_i\|^2}{\sum_i \|\boldsymbol x_i - \bar{\boldsymbol x}\|^2}\), which quantifies the proportion of variance in the original representations explained by atoms and thus reflects fidelity. While sparsity and separability can be quantified by \(K\) and \(\epsilon\), they do not directly indicate unit quality. In this section, we introduce a unified metric that integrates sparsity and separability to characterize the stability of representational units.

We note that \(\boldsymbol{\delta}_i \!\in \mathbb{R}^{|D|}\) can be viewed as a sparse representation of high-dimensional semantics, which is compressed by \(\boldsymbol D\! \in\! \mathbb{R}^{H \times |D|}\) to produce \(\boldsymbol{m}_i\!\in\! \mathbb{R}^{H}\). This perspective reveals a close connection to compressed sensing \cite{donoho2006compressed,candes2006robust}, whose core idea is that a signal sparse in some basis can be recovered from far fewer linear measurements than its ambient dimension, with the Restricted Isometry Property (RIP) \cite{goessmann2020restricted,chen2021conditioning} providing the essential guarantee.

\begin{definition}[Restricted Isometry Property; RIP]
    A matrix \(\tilde{\boldsymbol D} \in \mathbb{R}^{H \times |D|}\) is said to satisfy the \(K\)-RIP if there exists a constant \(\delta_K \in [0,1)\) such that, for any \(K\)-sparse vector \(\boldsymbol{\delta} \in \mathbb{R}^{|D|}\) (i.e., \(\|\boldsymbol\delta\|_0\le K\)),
    \begin{equation}
        (1-\delta_K)\|\boldsymbol{\delta}\|_2^2 \le \|\tilde{\boldsymbol  D} \boldsymbol{\delta}\|_2^2 \le (1+\delta_K)\|\boldsymbol{\delta}\|_2^2.
    \end{equation}
    Here, \(\delta_K\) is called the \(K\)-RIP constant of \(\tilde{\boldsymbol D}\).
\end{definition}

Intuitively, projecting a sparse vector via \(\tilde{\boldsymbol D}\) to lower dimension preserves its geometric structure, ensuring the possibility of recovery. Direct verification of the RIP is NP-hard, but coherence \cite{foucart2013invitation,murdock2020dataless} provides a computable upper bound on \(\delta_k\).

\begin{restatable}[Coherence–RIP Upper Bound]{theorem}{RIPUpperBound}\label{theorem:Coherence–RIP_Upper_Bound}
    Let \(\tilde{\boldsymbol D}\!\in \mathbb{R}^{H\times |D|}\) and define the coherence \(\mu \!:= \max_{i\neq j} |\langle \tilde{\boldsymbol d_i}, \tilde{\boldsymbol d_j}\rangle| \le \varepsilon\). For any \(K\)-sparse vector \(\boldsymbol{\delta} \in \mathbb{R}^{|D|}\) with \(\|\boldsymbol{\delta}\|_0 \le K\),
    \begin{equation}
        \big(1\!-\!(K\!-\!1)\mu\big)\|\boldsymbol{\delta}\|_2^2 \le\!\|\tilde{\boldsymbol D}\boldsymbol{\delta}\|_2^2\!\le\!\big(1\!+\!(K\!-\!1)\mu\big)\|\boldsymbol{\delta}\|_2^2.
    \end{equation}
    Hence \(\delta_K(\tilde{\boldsymbol D}) \le (K-1)\mu\). In particular, \(\tilde{\boldsymbol D}\) satisfies the \(K\)-RIP whenever \((K-1)\mu < 1\).
\end{restatable}

In other words, coherence provides a computable criterion for verifying the RIP, ensuring that all \(K\)-sparse vectors projected through \(\tilde{\boldsymbol D}\) preserve geometric structure, an essential prerequisite in compressed sensing. Nevertheless, the RIP alone does not preclude non-uniqueness \cite{donoho2001uncertainty,candes2005decoding}: even if \((K-1)\mu < 1\) holds, the sparse coefficients associated with representations need not be unique.

\begin{restatable}[Uniqueness and Exact \(\ell_1\!\) Recoverability]{theorem}{UniquenessRecovery}\label{theorem:Uniqueness_of_Sparse_Representation_and_Exact_Recovery}
    Let \(\tilde{\boldsymbol D}\in \mathbb{R}^{H\times |D|}\) and define the coherence \(\mu := \max_{i\neq j} |\langle \tilde{\boldsymbol d_i}, \tilde{\boldsymbol d_j}\rangle| \le \varepsilon\). If \(\mu < \frac{1}{2K-1}\), then for every \(\boldsymbol{\delta}\in\mathbb R^{|D|}\) with \(\|\boldsymbol{\delta}\|_0 \le K\), the \(K\)-sparse representation determined by \(\tilde{\boldsymbol m} = \tilde{\boldsymbol D}\boldsymbol{\delta}\) is unique; that is, no other \(K\)-sparse vector yields the same \(\tilde{\boldsymbol m}\). Moreover, \(\boldsymbol{\delta}\) is the unique minimizer of the convex program 
    \begin{equation}
        \min_{\boldsymbol{x} \in \mathbb{R}^{|D|}} \|\boldsymbol{x}\|_1\quad \text{subject to} \quad\tilde{\boldsymbol D}\,\boldsymbol{x} = \tilde{\boldsymbol m}.
    \end{equation}
\end{restatable}

Crucially, this intrinsically characterizes the monorepresentationality of \(\tilde{\boldsymbol D}\): under the condition \(\mu < \tfrac{1}{2K-1}\), any representation formed as a \(K\)-sparse linear combination of representational units in \(\tilde{\boldsymbol D}\) has a unique sparse decomposition, with no other combination yielding the same representation.

\begin{restatable}[Monorepresentationality / Injectivity]{corollary}{Monorepresentationality}\label{corollary:Monorepresentationality}
    \ Under the condition \(\mu < \tfrac{1}{2K-1}\) of Theorem~\ref{theorem:Uniqueness_of_Sparse_Representation_and_Exact_Recovery}, define \(\Sigma_K := \{\boldsymbol{\delta} \in \mathbb{R}^{|D|} : \|\boldsymbol{\delta}\|_0 \le K\}\) and \(\Phi : \Sigma_K \to \mathbb{R}^H\) by \(\Phi(\boldsymbol{\delta}) := \tilde{\boldsymbol D}\boldsymbol{\delta}\). Then \(\Phi\) is injective on \(\Sigma_K\). That is, for any \(\boldsymbol{x}, \boldsymbol{y} \in \Sigma_K\), if \(\tilde{\boldsymbol D}\boldsymbol{x} = \tilde{\boldsymbol D}\boldsymbol{y}\), it follows that \(\boldsymbol{x} = \boldsymbol{y}\).
\end{restatable}

Within this regime, representational units and their combinations are unambiguous in representation space, i.e., monorepresentationality. By contrast, monosemanticity \cite{bricken2023monosemanticity, templeton2024scaling} concerns the alignment of a unit with a specific meaning, concept, or function. The former is formally provable, whereas the latter is statistical and interpretive. Monorepresentationality is a prerequisite for monosemanticity, as it provides the structural stability; otherwise, units would be non-unique and interpretation unstable. Without additional semantic anchoring or inductive assumptions, theoretical guarantees primarily concern structural stability induced by monorepresentationality, while monosemanticity must be determined empirically via further semantic alignment experiments.

Sparsity (\(K\)) and separability (\(\mu\)) are thus unified by the condition \(\mu < \frac{1}{2K-1}\), under which stability holds. More practical metrics are discussed in the following `Stability metrics and computation' subsection.

\begin{remark*}
    For intuition, the above result can be viewed as a generalization of the strictly orthogonal case (\(|D| = H\)). In this setting, \(\tilde{ \boldsymbol D} \in \mathbb{R}^{H \times H}\) has orthogonal columns (\(\mu = 0\)), i.e., \(\tilde{ \boldsymbol D}^\top \tilde{\boldsymbol  D} = \boldsymbol I\). Any representation \(\tilde{\boldsymbol m} \in \mathbb{R}^H\) then admits a unique decomposition \(\boldsymbol\delta = \tilde{\boldsymbol  D}^\top \tilde{\boldsymbol m}\), satisfying \(\tilde{\boldsymbol m} = \tilde{ \boldsymbol D} \boldsymbol\delta\). Here the sparsity level \(K = H = |D|\) satisfies \(\mu < \tfrac{1}{2K-1}\). Thus, under strict orthogonality, representations and atom coefficients are in one-to-one correspondence, yielding a direct and unique identification of atoms.
\end{remark*}

\bmhead{Identification of atoms}\label{Section:Identification of atoms}

We have defined atoms and established criteria for ideal atoms. A central question remains whether such atoms can be identified in practice. Since sparse autoencoders (SAEs) are a standard approach for learning disentangled representations \cite{huben2024sparse}, we next demonstrate that, under appropriate conditions, SAEs can indeed identify these atoms, rendering the theory practically applicable.

\begin{restatable}[Identifiability of Threshold-activated SAEs; TSAEs]{theorem}{IdentifiabilitySAEs}\label{theorem:Identifiability_of_SAEs}
    Let \(M=\{\boldsymbol m_i \}_{i=1}^{|M|}\subset \mathbb R^H\) with \(\boldsymbol m_i = \boldsymbol D \boldsymbol \delta_i\), where \(\boldsymbol D = [\boldsymbol d_1,\cdots,\boldsymbol d_{|D|}] \in \mathbb{R}^{H\times |D|}\) satisfies \(|\langle \tilde{\boldsymbol{ d}_i}, \tilde{\boldsymbol{ d}_j} \rangle| \le \epsilon\), \(\forall i\ne j\), and each \(\boldsymbol \delta_i \in \mathbb{R}^{|D|}\) is \(K\)-sparse, i.e. \(\|\boldsymbol \delta_i\|_0 \le K\). Define the positive and negative supports as \(\mathcal S_i^{+}:=\{j:\delta_{ij}>0\},\ \mathcal S_i^{-}:=\{j:\delta_{ij}<0\}\), with \(\mathcal S_i=\mathcal S_i^{+}\cup\mathcal S_i^{-}=\operatorname{supp}(\boldsymbol\delta_i)\). Suppose there exist constants \(0<\delta_{\min}^{+}\le\delta_{\max}^{+},\ 0<\delta_{\min}^{-}\le\delta_{\max}^{-}\), such that \(\delta_{\min}^{+}\le\delta_{ij}\le\delta_{\max}^{+},\ \forall j\in\mathcal S_i^{+}\), and \(-\delta_{\max}^{-}\le\delta_{ij}\le-\delta_{\min}^{-},\ \forall j\in\mathcal S_i^{-}\). Consider the bidirectional threshold activation function
    \begin{equation}
        \sigma_{\tau_-,\tau_+}(x)=\begin{cases} 0, & \tau_-\le x\le\tau_+,\\ x, & x<\tau_-\ \text{or}\ x>\tau_+, \end{cases}
    \end{equation}
    with \(\tau_-<0<\tau_+\). Define \(B_0:=\epsilon\left( |\mathcal S_i^{+}|\delta_{\max}^{+} + |\mathcal S_i^{-}|\delta_{\max}^{-} \right)\), \(B_{+}:=\epsilon\left( (|\mathcal S_i^{+}|-1)\delta_{\max}^{+} + |\mathcal S_i^{-}|\delta_{\max}^{-} \right)\), and \(B_{-}:=\epsilon\left( |\mathcal S_i^{+}|\delta_{\max}^{+} + (|\mathcal S_i^{-}|-1)\delta_{\max}^{-} \right)\). If the bidirectional thresholds satisfy \(-\delta_{\min}^{-}+B_{-}<\tau_{-}\le -B_0,\ B_0\le\tau_{+}<\delta_{\min}^{+}-B_{+}\), with these intervals being non-empty whenever \(\delta_{\min}^{-}>B_0+B_{-},\ \delta_{\min}^{+}>B_0+B_{+}\), then setting the TSAE with \(\boldsymbol W_{dec}=\boldsymbol D\) and \(\boldsymbol W_{\mathrm{enc}}=\boldsymbol D^\top \tilde{ \boldsymbol S}\) yields
    \begin{equation}
        \forall i,\quad \boldsymbol W_{\mathrm{dec}}\,\sigma_{\tau_-,\tau_+}\left(\boldsymbol W_{\mathrm{enc}}\boldsymbol m_i\right) = \boldsymbol m_i.
    \end{equation}
    This parameterization exactly recover the atom set \(D\).
\end{restatable}

Therefore, TSAEs can identify atoms in principle. By contrast, ReLU \cite{huben2024sparse}, lacking a threshold term, fails to satisfy the support-separation condition and is thus theoretically insufficient. This responds to O'Neill et al. \cite{o2024compute}: the limitation of SAEs arises not from their linear–nonlinear mechanism, but from the absence of threshold activation, which prevents effective atom identification.

The threshold activation considered here denotes a class of activation mechanisms rather than a specific instantiation. AbsTop\(K\) \cite{zhu2026abstopk} is formally related to this class, but relies on a prespecified \(K\), which limits adaptivity and practical applicability. JumpReLU \cite{rajamanoharan2024jumping} and Top\(K\) \cite{gao2025scaling} can also be viewed as special cases of threshold-based activation, but their one-sided activation implicitly fixes a positive orientation for each representational direction. In a general representation space, such an orientation is arbitrary, and one-sided activation may therefore introduce unnecessary dependence on the sign convention of the representational direction, whereas bidirectional threshold activation avoids this limitation. 

%Empirically, we also observe that the advantage of BiJumpReLU over one-sided activation becomes more pronounced as the training scale increases.

The motivations underlying these methods are also fundamentally different. We introduce threshold activation to enable effective atom identification under approximate orthogonality. By contrast, JumpReLU \cite{rajamanoharan2024jumping} was originally proposed primarily to alleviate feature shrinkage \cite{rajamanoharan2024improving}, whereas Top\(K\) was designed to control sparsity directly by retaining a fixed number of activations \cite{gao2025scaling}. Thus, although these methods are formally related, their original objectives differ from the atom identifiability considered here.

\begin{remark*} 
    Although the theorem is stated for uniform scalar thresholds \(\tau_-\) and \(\tau_+\), it extends directly to coordinate-wise threshold vectors \(\boldsymbol{\tau}_-\) and \(\boldsymbol{\tau}_+\) without affecting the squeeze condition or the proof. This generalization enlarges the feasible intervals when activation magnitudes vary across coordinates, thereby improving support separation and the robustness of atom identification. 
\end{remark*}

\subsection*{Training details}\label{Section:Training details}

To assess the applicability of the theoretical framework across model architectures and representation types, we conduct experiments on neural networks spanning multiple parameter scales, architectures and generative modalities. In the following section, we describe the extraction of internal representations and the implementation and optimization settings used to train the corresponding SAE variants.

\bmhead{Models and representation extraction}\label{Section:Models and representation extraction}

To cover diverse model architectures and generative paradigms, we consider Transformer-based language models (Gemma-2-2B, Gemma-2-9B \cite{team2024gemma}, Llama-3.1-8B \cite{grattafiori2024llama} and Qwen-3.5-9B \cite{team2026qwen3}), the hybrid-architecture language model Olmo-Hybrid-7B \cite{merrill2026olmo}, and diffusion models (Stable Diffusion v1.5 \cite{rombach2022high} and FLUX.2-klein-9B \cite{flux-2-2025}). We further cover multiple input and representation types using Wikipedia \cite{volkel2006semantic} for continuous-text representations, Wikidata \cite{vrandevcic2014wikidata} for discrete-entity representations, LAION-COCO \cite{schuhmann2022laioncoco} for representations arising during image generation, and OpenHowNet \cite{qi2019openhownet} entity–sememe annotations for analyzing alignment between neural representational structure and human conceptual structure.

\bmhead{Representation extraction for SAE benchmarking}\label{Section:Representation extraction for SAE benchmarking}

To compare BiJumpReLU SAEs with alternative SAE architectures, we select representative layers or inference stages at approximately uniform intervals and train each SAE independently on the corresponding internal representations. For language models, we follow common settings in prior representation-analysis studies \cite{lieberum-etal-2024-gemma, he2024llama} and sample layers uniformly over model depth. Specifically, for Gemma-2-2B (26 layers), we use \(l\in\{0,5,10,15,20,25\}\), and for Gemma-2-9B (42 layers), \(l\in\{0,8,16,24,32,40\}\), extracting the output at the location `post\_feedforward\_layernorm`. For Qwen-3.5-9B and Olmo-Hybrid-7B (both 32 layers), we use \(l\in\{0,6,12,18,24,30\}\) and extract the output at `down\_proj`. For diffusion models, we follow common settings from prior representation-analysis work \cite{tinaz2025emergence}. For Stable Diffusion v1.5, we extract the output of `unet.mid\_block.attentions` at normalized diffusion times \(t\in\{0.0,0.5,1.0\}\). For FLUX.2-klein-9B, we extract the raw image-token block output from the global `transformer\_blocks` at layers \(l\in\{5,15,30\}\) and inference steps \(s\in\{0,3\}\). All internal representations are extracted directly from the original forward passes of the models without modifying the underlying computation and are used as input for subsequent SAE training.

Following prior empirical guidance on matching SAE capacity to representation scale \cite{hu2026towards}, we choose dataset sizes to maintain this balance while remaining compatible with the largest SAE configurations that can be trained reliably on a single A100 80-GB GPU. For continuous-text representations, we use Wikipedia and extract 300\text{K} internal representations at each of six representative layers for Gemma-2-2B, Gemma-2-9B, and Qwen3.5-9B, yielding \(6\times300\text{K}=1.8\text{M}\) representations per model. For discrete-entity representations, we use Wikidata and extract \(1.8\text{M}\) representations per model for Gemma-2-2B, Gemma-2-9B and Olmo-Hybrid-7B under the same setting. For image-generation representations, we use LAION-COCO. Stable Diffusion v1.5 yields \(3\times(8\times8)\times10\text{K}=1.92\text{M}\) representations in three diffusion stages and \(8\times8\) spatial positions. FLUX.2-klein-9B yields \(2\times3\times(8\times8)\times10\text{K}=3.84\text{M}\) representations across two inference stages, three representative layers, and \(8\times8\) spatial positions.

\bmhead{Representation extraction for cross-layer semantic alignment}\label{Section:Representation extraction for cross-layer semantic alignment}

To characterize the alignment between the neural representational structure and human conceptual structure, we extend representation extraction to all model layers and obtain internal representations for HowNet entities at each layer, enabling subsequent cross-layer atom identification and compositional analysis. We consider Gemma-2-2B, Gemma-2-9B, and Llama-3.1-8B. For Gemma-2-2B (26 layers) and Gemma-2-9B (42 layers), internal representations are extracted from the output at `post\_feedforward\_layernorm` in every layer \(l\in[0,26)\) and \(l\in[0,42)\), respectively. For Llama-3.1-8B (32 layers), internal representations are extracted from the output at `down\_proj` in every layer \(l\in[0,32)\). BiJumpReLU SAEs are then trained and applied at each layer to obtain representational atoms for cross-layer composition analysis. HowNet provides approximately \(123\text{K}\) unique English entity entries for semantic alignment. Extracting representations across all layers yields approximately \(26\times123\text{K}\approx3.20\text{M}\) internal representations for Gemma-2-2B, \(42\times123\text{K}\approx5.17\text{M}\) for Gemma-2-9B and \(32\times123\text{K}\approx3.94\text{M}\) for Llama-3.1-8B.

\bmhead{SAE architectures}\label{Section:SAE architectures}

An SAE can generally be written as
\begin{equation}
    f:\boldsymbol x\mapsto \hat{\boldsymbol x}=\boldsymbol W_{\mathrm{dec}}\,
\sigma(\boldsymbol W_{\mathrm{enc}}\boldsymbol x),
\end{equation}
where \(\boldsymbol x\in\mathbb R^H\), \(\boldsymbol W_{\mathrm{enc}}^\top,\boldsymbol W_{\mathrm{dec}}\in\mathbb R^{H\times D}\), \(D\) denotes the SAE dictionary width and \(\sigma\) is the activation function applied to the latent representation. Different SAE architectures are primarily distinguished by the form of \(\sigma\). Cunningham et al. \cite{rajamanoharan2024improving} introduced the ReLU SAEs, for which
\begin{equation}
    \mathrm{ReLU}:\quad\sigma_i(\boldsymbol z)=\max(0,z_i).
\end{equation}
Gao et al. \cite{gao2025scaling} proposed Top\(K\) SAEs, which retain only the \(K\) largest components of \(\boldsymbol z\), i.e.
\begin{equation}
    \mathrm{TopK}:\quad \sigma_i(\boldsymbol z) = z_i\,\mathbf 1\!\left[ i\in\operatorname{TopK}(\boldsymbol z,K) \right],
\end{equation}
where \(\operatorname{TopK}(\boldsymbol z,K)\) denotes the index set of the \(K\) largest components of \(\boldsymbol z\). Rajamanoharan et al. \cite{rajamanoharan2024jumping} proposed the JumpReLU SAE, which introduces learnable thresholds \(\theta_i\) and retains only positive activations above the corresponding threshold 
\begin{equation}
    \mathrm{JumpReLU}:\quad \sigma_i(\boldsymbol z) = z_i\,\mathbf 1[z_i>\theta_i].
\end{equation}
Zhu et al. \cite{zhu2026abstopk} proposed the AbsTop\(K\) SAE, which selects the \(K\) components with the largest activation magnitudes while preserving their original signs, i.e.
\begin{equation}
    \mathrm{AbsTopK}:\quad \sigma_i(\boldsymbol z) = z_i\,\mathbf 1\!\left[ i\in\operatorname{TopK}(|\boldsymbol z|,K) \right].
\end{equation}
For the BiJumpReLU SAE introduced here, we construct a bidirectional threshold activation according to the theoretical conditions for representational atom identification:
\begin{equation}
    \mathrm{BiJumpReLU}:\quad \sigma_i(\boldsymbol z) = z_i\,\mathbf 1[z_i<\tau_i^-\vee z_i>\tau_i^+].
\end{equation}
This form follows from both the sign symmetry of representational directions and the theoretical requirements for atom identifiability. A representational direction and its negation have equivalent geometric status in the internal representation space and therefore should not be assigned an a priori preferred orientation. BiJumpReLU accordingly applies learnable thresholds to both positive and negative activations, allowing latent components to be selected without introducing a sign-dependent bias and aligning the architecture with the theoretical framework for representational atom identification.

\bmhead{Loss function}\label{Section:Loss function}

SAEs are trained by minimizing a joint reconstruction–sparsity objective (see Theorem~\ref{theorem:Uniqueness_of_Sparse_Representation_and_Exact_Recovery} in `Evaluation of atoms' subsection for the theoretical motivation):
\begin{equation}
    \mathcal L(\boldsymbol x) = \underbrace{\|\boldsymbol x-\hat{\boldsymbol x}\|_2^2}_{\mathcal L_{\mathrm{reconstruct}}} + \lambda \underbrace{\|\sigma(\boldsymbol z)\|_1}_{\mathcal L_{\mathrm{sparsity}}},
\end{equation}
where \(\boldsymbol z=\boldsymbol W_{\mathrm{enc}}\boldsymbol x\), \(\sigma\) denotes the activation function of the corresponding SAEs, and \(\lambda\) controls the trade-off between reconstruction and sparsity. ReLU SAEs use this objective directly. Because JumpReLU and BiJumpReLU SAEs contain learnable thresholds, we additionally use a straight-through estimator to approximate gradients through the thresholding operation during backpropagation \cite{rajamanoharan2024improving}. By contrast, Top\(K\) and AbsTop\(K\) SAEs explicitly control activation sparsity through a prespecified \(K\) and are therefore trained using only the reconstruction term, \(\mathcal L(\boldsymbol x)=\|\boldsymbol x-\hat{\boldsymbol x}\|_2^2\).

\bmhead{Data preprocessing and hyperparameter selection}\label{Section:Data preprocessing and hyperparameter selection}

Internal representations can differ substantially in numerical scale across models, layers, and extraction sites, making SAE training highly sensitive to hyperparameter choices and complicating the use of a unified training configuration. We first observed on Gemma-2-2B that SAEs trained directly on raw activations were stable in shallow layers but became increasingly unstable in deeper layers, largely due to pronounced variation in activation scale across depth. Scaling each internal representation by the global standard deviation restored stable training across layers, with consistent behavior in Gemma-2-9B. For representations extracted from sites such as `down\_proj`, which are not normalized by `post\_feedforward\_layernorm`, standard-deviation scaling alone did not yield sufficiently consistent numerical distributions. We therefore applied RMS normalization without learnable affine parameters, scaling each internal representation by its root mean square. This produced stable training for Llama-3.1-8B, Qwen3.5-9B, Olmo-Hybrid-7B, Stable Diffusion v1.5 and FLUX.2-klein-9B. Accordingly, representations extracted from `post\_feedforward\_layernorm` in Gemma-2-2B and Gemma-2-9B use standard-deviation scaling, whereas representations from `down\_proj` and other unnormalized sites use RMS normalization. Both operations apply only a scalar rescaling to each representation vector, preserving its direction and therefore the representational geometry considered here, and the original numerical scale can be recovered directly using the corresponding scaling factor.

The unified preprocessing yields comparable numerical scales across models, layers, and modalities, allowing all SAEs to use a common set of core hyperparameters without model- or layer-specific grid searches. BiJumpReLU SAEs use \(\lambda=5.0\) in all settings, with the negative and positive thresholds initialized to \(-0.5\) and \(0.5\), respectively. ReLU SAEs likewise use \(\lambda=5.0\), whereas JumpReLU SAEs use \(\lambda=5.0\) with the threshold initialized to \(0.5\). For Top\(K\) SAEs, which control sparsity by fixing the number of active components, we follow the standard setting of Gao et al. \cite{gao2025scaling} with \(K=50\) and additionally evaluate \(K=20\) as a lower-sparsity baseline. AbsTop\(K\) SAEs use \(K=50\), following the standard setting of Zhu et al.\cite{zhu2026abstopk}

All remaining training settings are held fixed to minimize confounding from hyperparameter differences across SAE architectures. We use the learning rate of \(2\times10^{-4}\), the Adam optimizer with \(\beta_1=0.9\) and \(\beta_2=0.999\), and the batch size of 4096 for all models. Increasing the batch size further did not yield measurable performance gains.

\bmhead{SAE Capacity}\label{Section:SAE Capacity}

For SAE benchmarking, we compare the ability of different SAE architectures to identify representational atoms under otherwise matched training conditions, while keeping dictionary sizes comparable across base models. We therefore choose model-specific expansion factors such that dictionary widths remain within approximately \(2.1\text{-}2.2\times10^5\): \(96\times2304\approx221\text{K}\) for Gemma-2-2B, \(60\times3584\approx215\text{K}\) for Gemma-2-9B, \(54\times4096\approx221\text{K}\) for Qwen3.5-9B, and \(58\times3840\approx223\text{K}\) for Olmo-Hybrid-7B. For cross-layer semantic alignment, the objective is instead to obtain representational atoms with high fidelity and high stability, defined as \(R^2>0.99\) and \(q^*>0.99\). Following the empirical guidance of Hu et al. \cite{hu2026towards}, we select the SAE dictionary as compact as possible that satisfies these criteria and keep its width fixed across all layers of the same base model. Specifically, we use \(64\times2304\approx147\text{K}\) for Gemma-2-2B, \(40\times3584\approx143\text{K}\) for Gemma-2-9B, and \(35\times4096\approx143\text{K}\) for Llama-3.1-8B.

\bmhead{Training dynamics and early stopping}\label{Section:Training dynamics and early stopping}

SAE training is commonly viewed as a trade-off between reconstruction quality and sparsity \cite{gao2025scaling}, but we observe a consistent training dynamic across multiple experimental settings. For BiJumpReLU SAEs, reconstruction and sparsity losses typically decrease together during early and intermediate training. As training progresses, the reconstruction loss gradually plateaus and may subsequently increase slightly but persistently, while sparsity continues to improve. Both objectives eventually reach an approximately stable regime. This pattern indicates that late-stage optimization primarily trades a limited change in reconstruction quality for increased sparsity, rather than continuing to improve both objectives simultaneously. Based on this dynamic, we adopt a trajectory-based early stopping criterion. For BiJumpReLU and JumpReLU SAEs, we smooth the training-loss curves and stop training near the transition at which reconstruction loss shifts from sustained decrease to a stable increase while sparsity continues to improve. This stage typically already corresponds to high fidelity and stability, allowing us to avoid unnecessary late-stage trade-offs while preserving atom-identification quality. For ReLU SAEs, where this transition is generally less pronounced, training is stopped when neither reconstruction nor sparsity loss shows further substantial improvement. For Top\(K\) and AbsTop\(K\) SAEs, whose sparsity is explicitly fixed by \(K\), training is stopped when the reconstruction performance no longer improves appreciably.

\subsection*{Evaluation details}\label{Section:Evaluation details}

To systematically evaluate the quality and semantic organization of identified representational atoms, we describe the implementation of fidelity and stability metrics, monosemanticity evaluation, and alignment analysis between neural representational structure and human conceptual structure in the following subsection.

\bmhead{Fidelity metrics and computation}\label{Section:Fidelity metrics and computation}

Fidelity measures how well candidate FRUs preserve or reconstruct the original internal representations and is quantified by the coefficient of determination
\begin{equation}
    R^2 := 1- \frac{\sum_i\|\boldsymbol m_i-\hat{\boldsymbol m}_i\|_2^2} {\sum_i\|\boldsymbol m_i-\bar{\boldsymbol m}\|_2^2},
\end{equation}
where \(\hat{\boldsymbol m}_i\) denotes the internal representation reconstructed from the candidate FRUs and \(\bar{\boldsymbol m}\) is the sample mean. Values of \(R^2\) closer to 1 indicate more faithful reconstruction of the original representations.

\bmhead{Stability metrics and computation}\label{Section:Stability metrics and computation}

To transform the distinguishability condition \(\mu<1/(2K-1)\) from Theorems~\ref{theorem:Uniqueness_of_Sparse_Representation_and_Exact_Recovery} and Corollary~\ref{corollary:Monorepresentationality} into a computable stability metric, we introduce quantile statistics that characterize the empirical distributions of sparsity and coherence. We first define the quantile sparsity \(K_q\) as
\begin{equation}
    K_q = \inf\left\{ k\in\mathbb N^+ \mid \mathbb P_{\boldsymbol\delta\sim\mathcal P_\Delta} \bigl(\|\boldsymbol\delta\|_0\le k\bigr) \ge q \right\},
\end{equation}
where \(\boldsymbol\delta\) is sampled from the coefficient distribution \(\mathcal P_\Delta\). Thus, \(K_q\) is the smallest integer such that at least a fraction \(q\) of samples have sparsity no greater than \(K_q\).

Similarly, we define the quantile coherence \(\mu_q\) as
\begin{equation}
    \mu_q = \inf\left\{ \mu\ge0 \mid \mathbb P_{(\mathcal I,\mathcal J)} \left( \left| \left\langle \tilde{\boldsymbol d}_{\mathcal I}, \tilde{\boldsymbol d}_{\mathcal J} \right\rangle \right| \le\mu \right) \ge q \right\},
\end{equation}
where \((\mathcal I,\mathcal J)\) is sampled uniformly from the unordered pairs of distinct representational-unit indices, and \(\left| \left\langle \tilde{\boldsymbol d}_{\mathcal I}, \tilde{\boldsymbol d}_{\mathcal J} \right\rangle \right|\) denotes their coherence. Consequently, \(\mu_q\) gives an upper bound on coherence satisfied by at least a fraction \(q\) of unit pairs.

Using the distinguishability criterion above, we evaluate \(\mu_q<\frac{1}{2K_q-1}\) at each quantile and define stability as the largest quantile that satisfies the condition
\begin{equation}
    q^* := \sup\left\{ q\mid \mu_q<\frac{1}{2K_q-1} \right\}.
\end{equation}
Thus, \(q^*\) denotes the largest quantile at which empirical sparsity and coherence satisfy the theoretical distinguishability condition, with the ideal value \(q^*=1\).

In practice, we compute \(q^*\) using binary search. Specifically, on \([0,0.999999]\), we iteratively evaluate
\begin{equation}
    \mu_\alpha:=\operatorname{Quantile}(\mu,\alpha), \quad K_\alpha:=\operatorname{Quantile}(K,\alpha),
\end{equation}
and test whether \(\mu_\alpha<\frac{1}{2K_\alpha-1}\).If the condition is satisfied, the lower search bound increases; otherwise, the upper bound decreases until convergence. The largest feasible quantile is reported as \(q^*\).

\bmhead{Evaluation of existing representational units}\label{Section:Evaluation of existing representational units}

To evaluate existing representational units within a unified framework, we treat neurons and learned features by publicly available SAEs as candidate FRUs and assess them using the fidelity and stability metrics defined above. All evaluations use approximately \(20\text{K}\) entities appearing in CounterFact \cite{meng2022locating}, which can be regarded as a subset of Wikidata entities. Each entity is converted into a corresponding prompt and fed into the model, and internal representations are extracted at the final token position of the target site. Neurons and learned features are evaluated directly using the original models and publicly released SAE checkpoints, without additional training or parameter tuning.

For neurons, we extract the input to the MLP `down\_proj` at each layer. This vector corresponds to the activations of the intermediate MLP neurons, with each component indicating the activation strength of the corresponding neuron. The columns of the `down\_proj` weight matrix are therefore treated as the representational directions associated with individual neurons, and the corresponding input activations as their coefficients, allowing the MLP residual update to be expressed as a linear combination of neuron directions. Fidelity and stability are then computed for neurons as candidate FRUs according to the definitions above.

For learned features, we use the pretrained SAEs provided by Gemma Scope \cite{lieberum-etal-2024-gemma} and Llama Scope \cite{he2024llama}, following their official settings. For Gemma-2-2B and Gemma-2-9B, we extract representations at the output of `post\_feedforward\_layernorm` from each layer and apply the corresponding Gemma Scope SAE. For Llama-3.1-8B, we extract representations at the output of `down\_proj` from each layer and apply the corresponding Llama Scope SAE. The decoder directions of each SAE are treated as candidate FRUs, and the latent activations as their coefficients. The resulting SAE reconstructions are used to compute fidelity, whereas feature coherence and activation sparsity are used to compute stability under the same evaluation procedure.

\bmhead{Monosemanticity evaluation details}\label{Section:Monosemanticity evaluation details}

To evaluate the monosemanticity of the candidate FRUs, we use a common scoring criterion across language models and diffusion models, with modality-specific evaluation procedures. For language models, including Gemma-2-2B, Gemma-2-9B, Qwen3.5-9B and Olmo-Hybrid-7B, we use an LLM-as-a-judge protocol supplemented by manual verification. For diffusion models, including Stable Diffusion v1.5 and FLUX.2-klein-9B, we use manual evaluation. Both procedures follow the same definition of monosemanticity and a closely matched scoring template.

For each model and representation extraction site, we randomly sample 10 candidate FRUs identified by the corresponding SAE and collect the input instances that activate each unit. If more than 10 instances are activated, we randomly sample 10 for evaluation. The evaluated instances are contextualized tokens for Wikipedia, entities for Wikidata and local image patches for LAION-COCO. For each candidate FRU, we identify the largest subset \(S_{\max}\) of evaluated instances that can be coherently explained by a common concept, attribute or semantic pattern, and define its monosemanticity score as
\begin{equation}
    \mathrm{Monosemanticity} = \frac{|S_{\max}|}{|S|},
\end{equation}
where \(S\) denotes the complete set of evaluated instances for that unit. Scores closer to 1 indicate that the unit activates more consistently on a coherent semantic pattern. Finally, we average the monosemanticity scores of the sampled units within each SAE setting to obtain its overall monosemanticity score.

\bmhead{Semantic alignment evaluation details}\label{Section:Semantic alignment evaluation details}

HowNet \cite{qi2019openhownet} contains 2,540 sememes, each associated with a set of entities belonging to the corresponding semantic category. To ensure that subsequent set-based comparisons are meaningful, we retain only sememes containing at least two unique entities, yielding 2,135 sememes, which we use as an operational representation of human conceptual semantic structure. On the model side, we extract internal representations corresponding to HowNet entities at every layer and train BiJumpReLU SAEs at each layer to identify atoms. For each unit, we record the set of entities that activate it and likewise retain only units associated with at least two unique entities. Each retained sememe and representational unit can therefore be represented as an entity set, providing a common basis for comparing neural representational structure with human conceptual structure.

To evaluate semantic alignment between single atoms and HowNet sememes, we compute precision, recall and F1 based on the overlap between their corresponding entity sets. For each sememe, we search across all atoms identified at all layers and define the unit with the highest F1 score as its best single-unit match. This metric jointly reflects the semantic purity of a atom and its coverage of the target sememe, thereby quantifying the extent to which an single atoms corresponds to human conceptual semantic structure. Detailed set construction, metric definitions and cross-layer search procedures are provided in `Single-atom semantic alignment evaluation' in the Supplementary Information.

Building on the single-atom analysis, we further assess whether combinations of atoms can recover sememe structure more completely. We aggregate candidate FRUs identified across all layers, prefilter them according to their correspondence with the target sememe entity set, and then perform greedy compositional search by iteratively adding the atom that yields the largest improvement in the current F1 score. Search terminates when no further positive gain is obtained or when a prespecified maximum combination size is reached. The candidate-filtering threshold and maximum combination size are selected by grid search (Supplementary Fig. 1). This analysis compares the coverage achieved by single atoms and atom combinations for the same sememe and tests whether higher-level human conceptual semantic structures can be formed compositionally from representational atoms distributed across layers. Detailed procedures for candidate construction, compositional search and hyperparameter selection are provided in `Semantic alignment evaluation' in the Supplementary Information.

\subsection*{Use of large language models}\label{Section:Use of large language models}

During the preparation of this manuscript, we used ChatGPT (OpenAI; GPT-5.6 Sol) as a writing-support tool. Its use was limited to language refinement, checks of logical consistency and coherence, and comparison of alternative phrasings. All AI-assisted suggestions were reviewed and evaluated by the authors, and any suggestions incorporated into the manuscript were revised as appropriate. The theoretical development, methodological design, experimental implementation and data analysis were conducted by the authors. The authors independently determined the scientific conclusions and take full responsibility for the accuracy and final content of the manuscript.


\section*{Data availability}

The Wikipedia data used in this study were obtained from the `20231101.en' configuration at \url{https://huggingface.co/datasets/wikimedia/wikipedia}. The Wikidata data were obtained from the publicly available Wikidata resources \cite{vrandevcic2014wikidata} at \url{https://www.wikidata.org/wiki/Wikidata:Data_access} and were further preprocessed to obtain a more balanced distribution across entity categories. The LAION-COCO data were obtained from the public LAION-COCO release \cite{schuhmann2022laioncoco} at \url{https://huggingface.co/datasets/guangyil/laion-coco-aesthetic}. HowNet data were obtained through OpenHowNet \cite{Dong2003HowNetA,qi2019openhownet} at \url{https://github.com/thunlp/OpenHowNet}.

\section*{Code availability}

The source code and configuration files used for representation extraction, sparse autoencoder training, fidelity and stability evaluation, monosemanticity analysis, and semantic alignment are publicly available at \url{https://github.com/ChenhuiHu/representational_atoms}. The repository also provides documentation and example configurations for reproducing the experimental pipeline described in this study.

\bibliography{sn-bibliography}% common bib file

\setcounter{table}{0}
\renewcommand{\tablename}{Extended Data Table}

\setcounter{figure}{0}
\renewcommand{\figurename}{Extended Data Fig.}

\makeatletter
\newlength{\oldfptop}
\newlength{\oldfpbot}
\newlength{\olddblfptop}
\newlength{\olddblfpbot}

\setlength{\oldfptop}{\@fptop}
\setlength{\oldfpbot}{\@fpbot}
\setlength{\olddblfptop}{\@dblfptop}
\setlength{\olddblfpbot}{\@dblfpbot}

\setlength{\@fptop}{0pt}
\setlength{\@fpbot}{0pt plus 1fil}
\setlength{\@dblfptop}{0pt}
\setlength{\@dblfpbot}{0pt plus 1fil}
\makeatother

\clearpage
\begin{figure}
    \centering
    \includegraphics[width=0.98\textwidth]{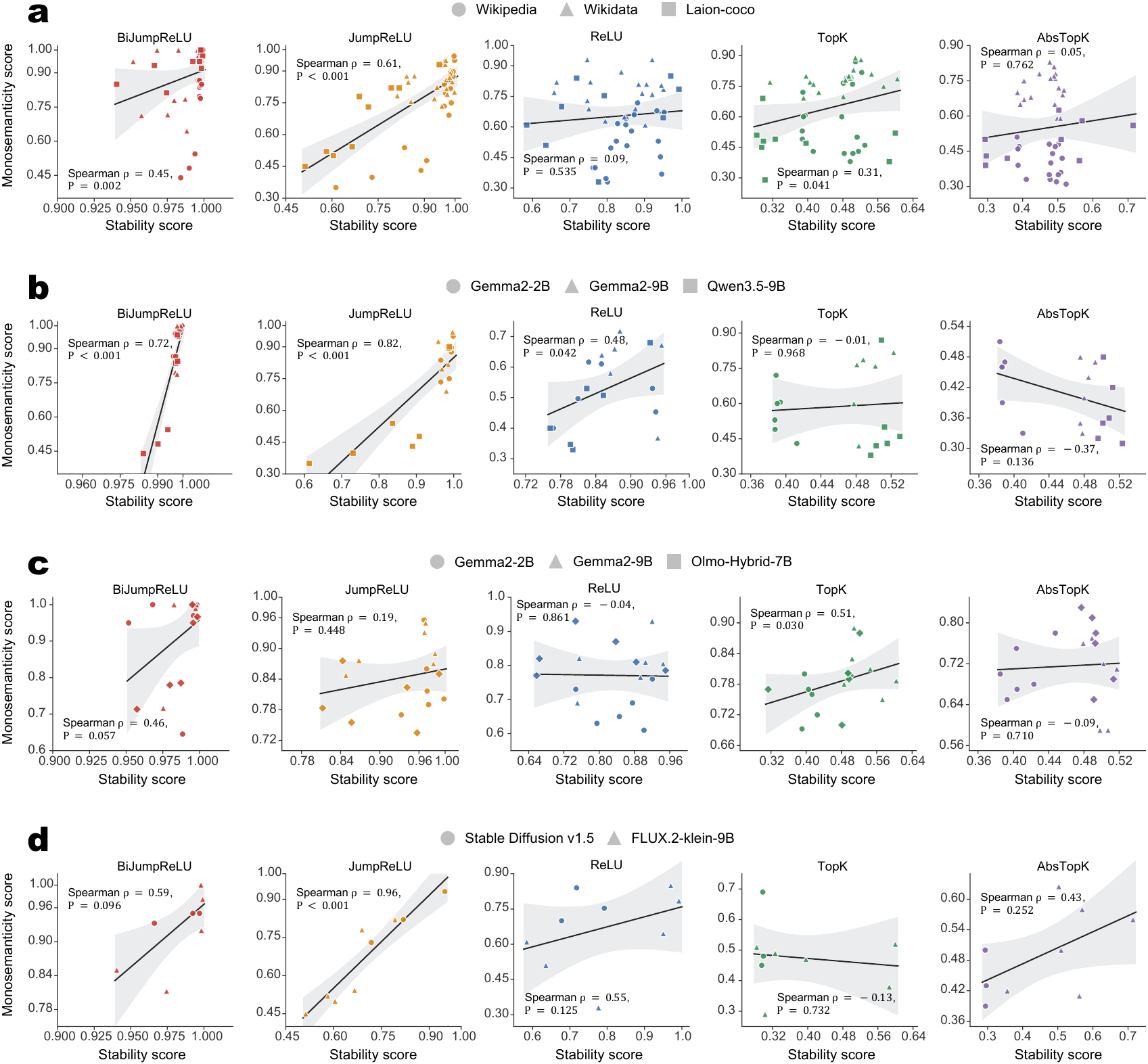}
    \caption{\textbf{Architecture-stratified associations between stability and monosemanticity.} \textbf{a}, Stability scores are plotted against mean monosemanticity scores separately for BiJumpReLU, JumpReLU, ReLU, TopK (\(K=50\)) and AbsTopK (\(K=50\)) SAEs, pooling all evaluated models and datasets. Within each SAE panel, each point represents one evaluated model--condition pair: one layer for language models, one diffusion timestep for Stable Diffusion v1.5, or one layer--step combination for FLUX.2-klein-9B (\(N=45\) per SAE architecture). Marker shapes denote datasets. \textbf{b}, The same analysis restricted to Wikipedia, pooling Gemma2-2B, Gemma2-9B and Qwen3.5-9B (\(N=18\) per SAE architecture). \textbf{c}, The same analysis restricted to Wikidata, pooling Gemma2-2B, Gemma2-9B and Olmo-Hybrid-7B (\(N=18\) per SAE architecture). \textbf{d}, The same analysis restricted to LAION-COCO, pooling Stable Diffusion v1.5 and FLUX.2-klein-9B (\(N=9\) per SAE architecture). In \textbf{b--d}, marker shapes denote individual models. Black lines show linear regression fits with 95\% confidence bands. Spearman's rank correlation coefficients (\(\rho\)) and two-sided \(P\) values are shown in each panel.}
  \label{fig:Extended_Fig_1}
\end{figure}

\clearpage
\begin{figure}
    \centering
    \includegraphics[width=0.98\textwidth]{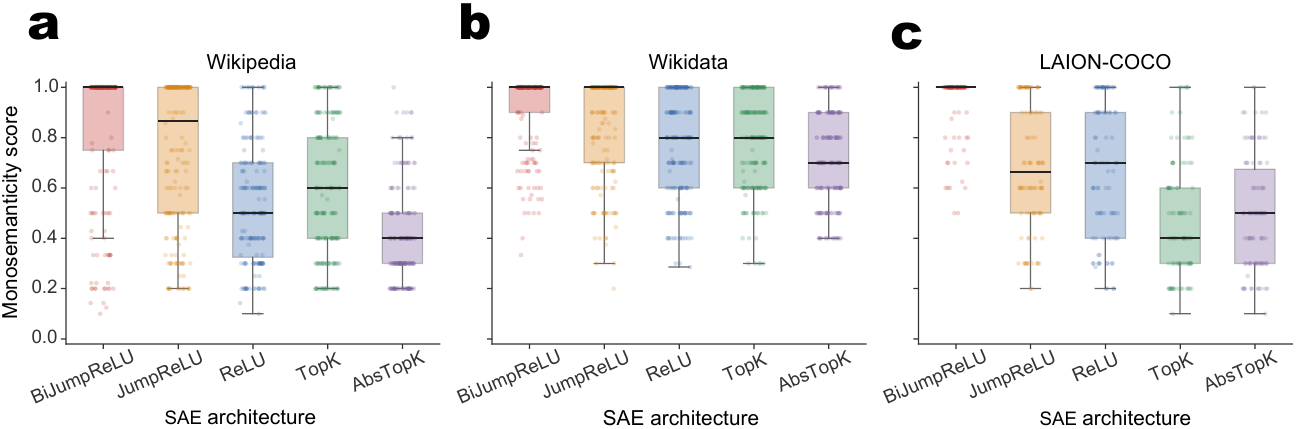}
    \caption{\textbf{Distributions of monosemanticity scores across SAE architectures.} \textbf{a--c}, Distributions of monosemanticity scores for BiJumpReLU, JumpReLU, ReLU, TopK (\(K=50\)) and AbsTopK (\(K=50\)) SAEs on Wikipedia (\textbf{a}), Wikidata (\textbf{b}) and LAION-COCO (\textbf{c}). For each SAE trained on representations from the corresponding evaluated layer, diffusion timestep or layer--step combination, ten FRUs were assessed individually using their activation contexts, yielding one monosemanticity score per FRU. The ten scores were averaged to obtain the monosemanticity score for that SAE at the corresponding representation condition.}
  \label{fig:Extended_Fig_2}
\end{figure}

\clearpage
\begin{figure}
    \centering
    \includegraphics[width=0.98\textwidth]{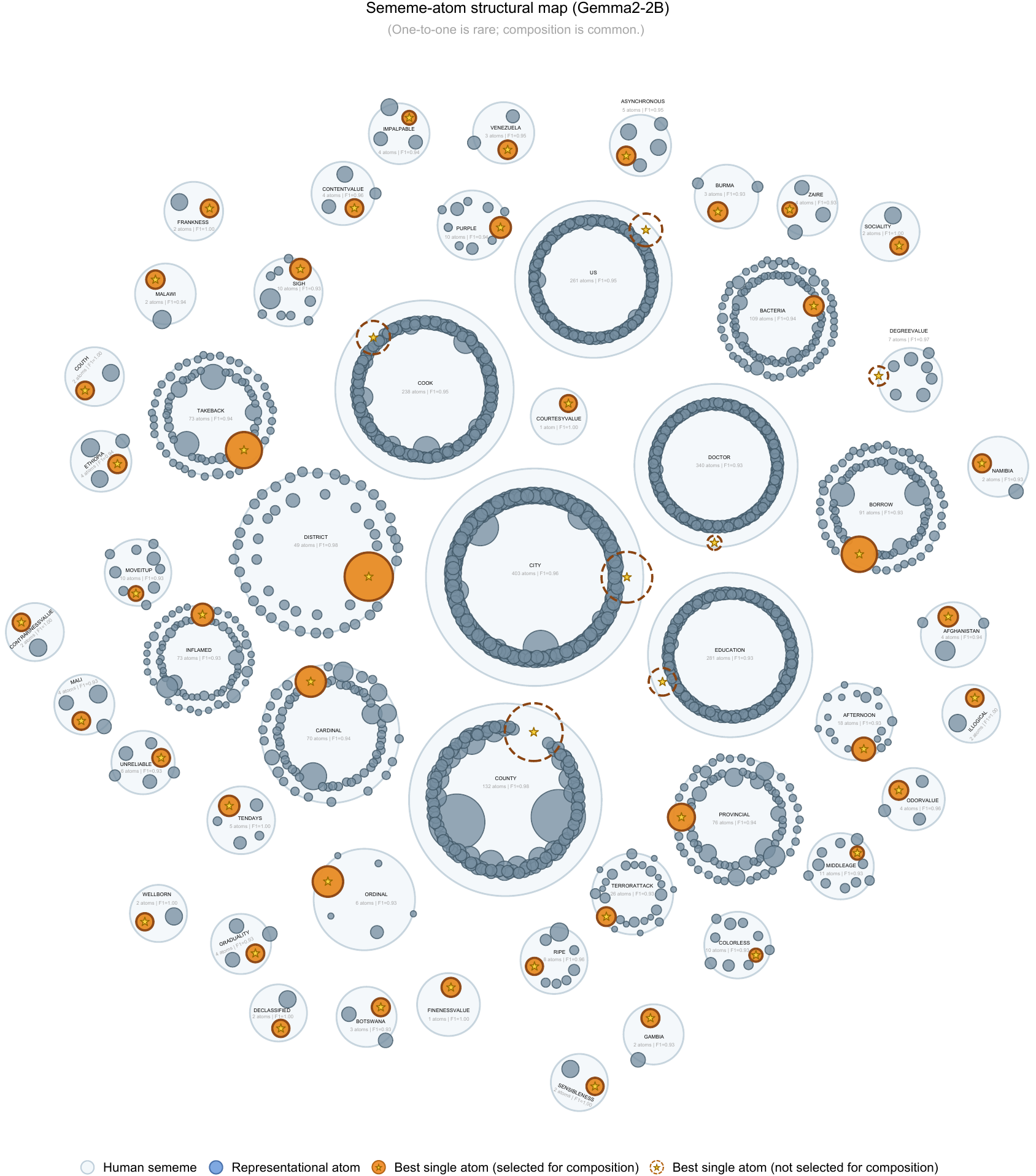}
    \caption{\textbf{Sememe--atom structural map for Gemma2-2B.} Visualization of the alignment between HowNet sememes and representational atoms identified by BiJumpReLU SAEs across layers of Gemma2-2B. Large light-blue circles denote human sememes and smaller blue circles denote representational atoms selected for their compositional recovery. Solid orange circles indicate the best single atoms that are retained in the composition, whereas dashed orange circles indicate the best single atoms that are not retained. Circle sizes reflect the sizes of the corresponding semantic sets, and the geometric overlap between an atom and its sememe reflects their single-atom precision. Geometric overlap between atom circles is not intended to represent overlap between their associated entity sets.}
  \label{fig:Extended_Fig_3}
\end{figure}

\clearpage
\begin{figure}
    \centering
    \includegraphics[width=0.98\textwidth]{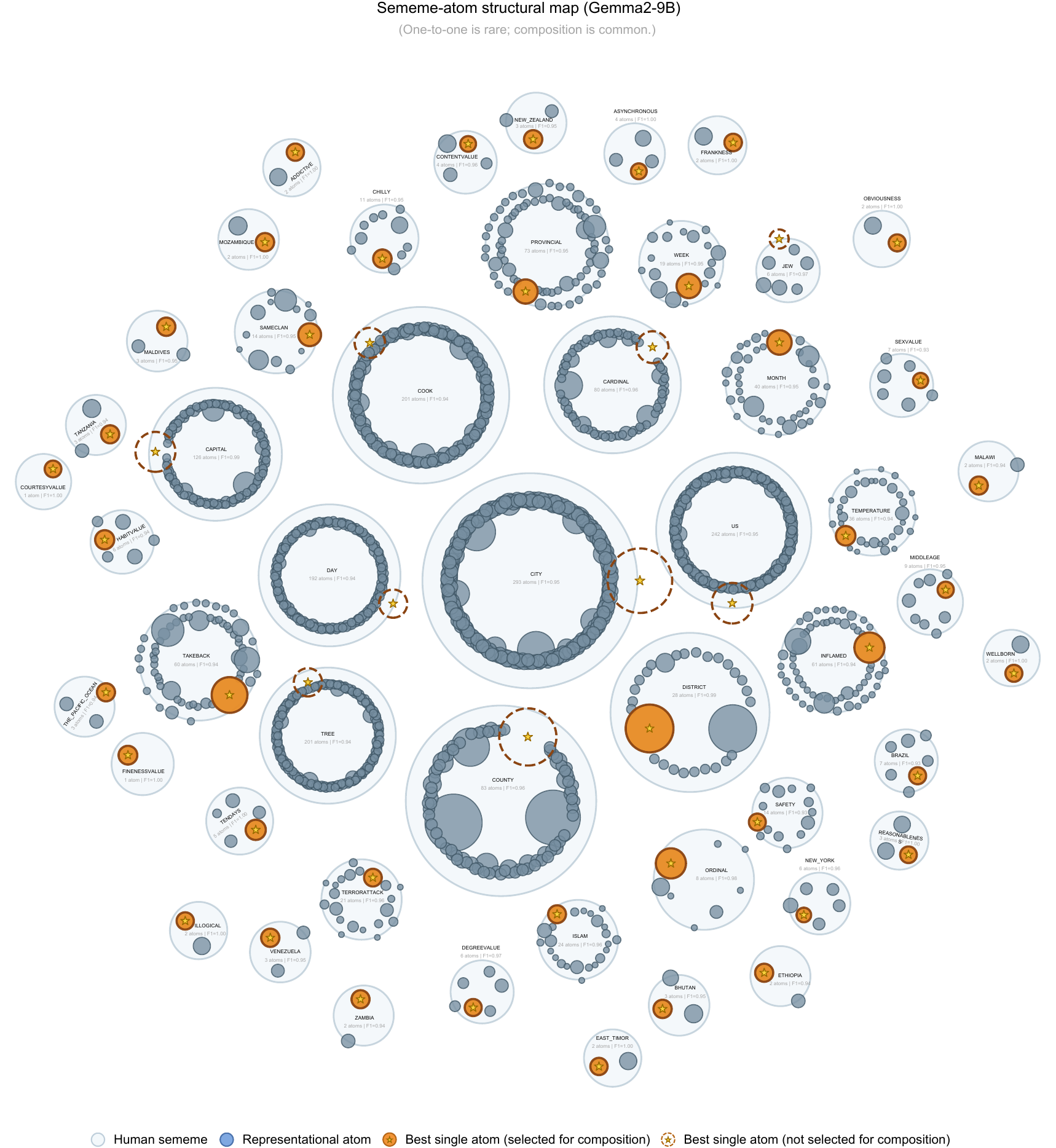}
    \caption{\textbf{Sememe--atom structural map for Gemma2-9B.} Visualization of the alignment between HowNet sememes and representational atoms identified by BiJumpReLU SAEs across layers of Gemma2-9B. Large light-blue circles denote human sememes and smaller blue circles denote representational atoms selected for their compositional recovery. Solid orange circles indicate the best single atoms that are retained in the composition, whereas dashed orange circles indicate the best single atoms that are not retained. Circle sizes reflect the sizes of the corresponding semantic sets, and the geometric overlap between an atom and its sememe reflects their single-atom precision. Geometric overlap between atom circles is not intended to represent overlap between their associated entity sets.}
  \label{fig:Extended_Fig_4}
\end{figure}

\clearpage
\begin{figure}
    \centering
    \includegraphics[width=0.98\textwidth]{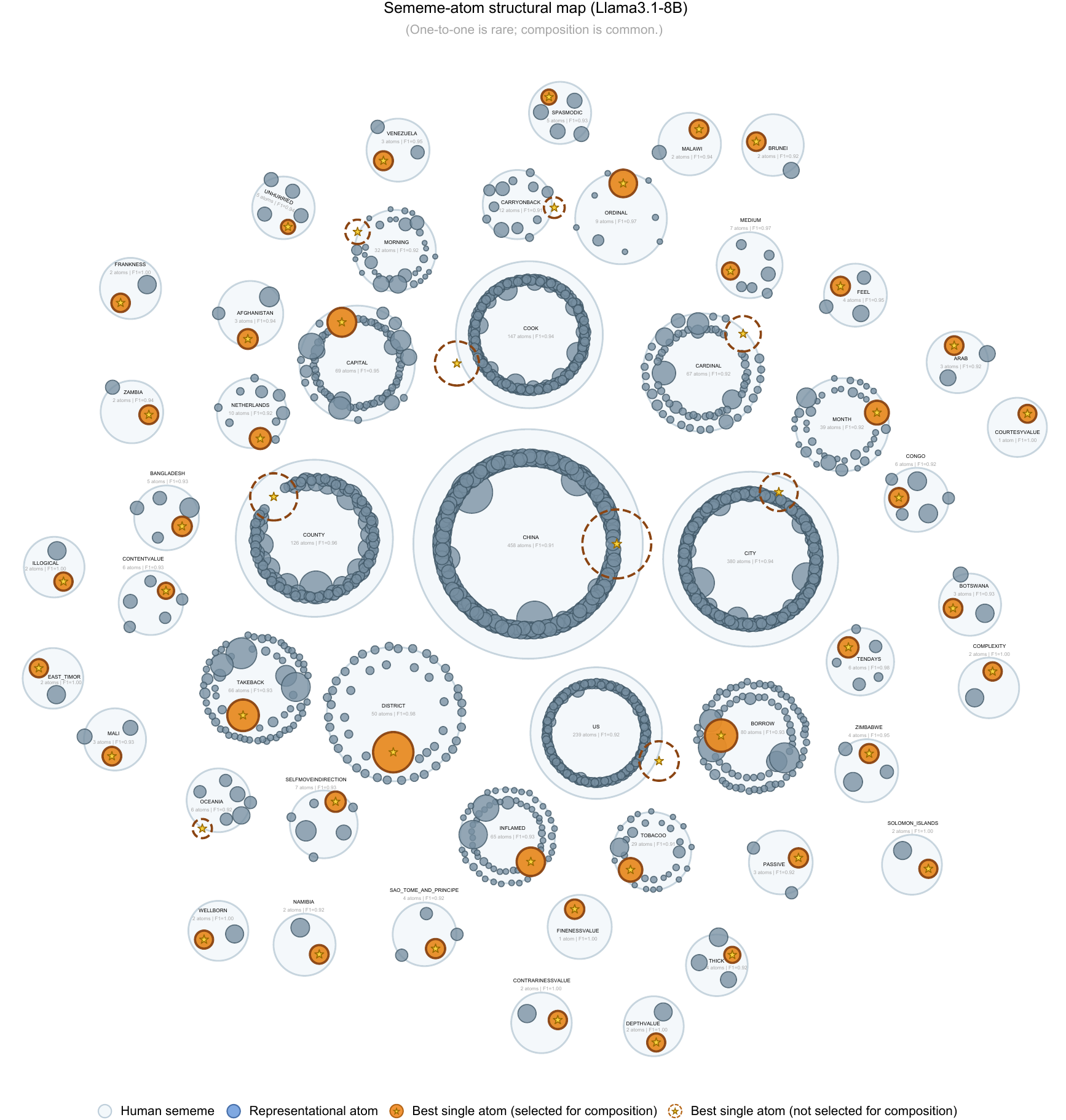}
    \caption{\textbf{Sememe--atom structural map for Llama3.1-8B.} Visualization of the alignment between HowNet sememes and representational atoms identified by BiJumpReLU SAEs across layers of Llama3.1-8B. Large light-blue circles denote human sememes and smaller blue circles denote representational atoms selected for their compositional recovery. Solid orange circles indicate the best single atoms that are retained in the composition, whereas dashed orange circles indicate the best single atoms that are not retained. Circle sizes reflect the sizes of the corresponding semantic sets, and the geometric overlap between an atom and its sememe reflects their single-atom precision. Geometric overlap between atom circles is not intended to represent overlap between their associated entity sets.}
  \label{fig:Extended_Fig_5}
\end{figure}

\clearpage
\begin{table*}[p]
\centering
\caption{\textbf{Comprehensive evaluation of sparse autoencoder representations across datasets and models.} In addition to fidelity and stability, the table reports feature utilization and retention.}
\label{tab:sae_evaluation_ci95_full}

\scriptsize
\setlength{\tabcolsep}{2.5pt}
\renewcommand{\arraystretch}{1.08}

\resizebox{\textwidth}{!}{
\begin{tabular}{@{}lllccccccc@{}}
\toprule
\multirow{2}{*}{Dataset} & \multirow{2}{*}{Model} & \multirow{2}{*}{Metric}
& \multicolumn{2}{c}{GemmaScope}
& \multicolumn{5}{c}{Trained SAEs} \\
\cmidrule(lr){4-5}
\cmidrule(lr){6-10}
& & & Small & Large & ReLU & TopK & AbsTopK & JumpReLU & BiJumpReLU \\
\midrule

% =========================================================
% Wikipedia
% =========================================================

\multirow{12}{*}{Wikipedia}
& \multirow{4}{*}{Gemma2-2B}
& Utilization $\uparrow$
& $95.62\,{\scriptstyle(\pm 5.55)}$
& $76.02\,{\scriptstyle(\pm 6.66)}$
& $23.63\,{\scriptstyle(\pm 32.29)}$
& $15.62\,{\scriptstyle(\pm 5.24)}$
& $16.45\,{\scriptstyle(\pm 5.74)}$
& $67.63\,{\scriptstyle(\pm 14.58)}$
& $\mathbf{91.47\,{\scriptstyle(\pm 20.80)}}$ \\
&
& Retention $\alpha_1$ $\uparrow$
& $85.74\,{\scriptstyle(\pm 11.90)}$
& $85.17\,{\scriptstyle(\pm 12.55)}$
& $96.41\,{\scriptstyle(\pm 1.75)}$
& $70.02\,{\scriptstyle(\pm 1.31)}$
& $69.72\,{\scriptstyle(\pm 1.33)}$
& $99.87\,{\scriptstyle(\pm 0.10)}$
& $\mathbf{99.97\,{\scriptstyle(\pm 0.02)}}$ \\
&
& Fidelity $\uparrow$
& $72.17\,{\scriptstyle(\pm 13.21)}$
& $75.52\,{\scriptstyle(\pm 13.12)}$
& $75.46\,{\scriptstyle(\pm 16.17)}$
& $88.68\,{\scriptstyle(\pm 7.05)}$
& $88.11\,{\scriptstyle(\pm 6.93)}$
& $94.91\,{\scriptstyle(\pm 2.38)}$
& $\mathbf{97.25\,{\scriptstyle(\pm 1.44)}}$ \\
&
& Stability $\alpha_2$ $\uparrow$
& $67.87\,{\scriptstyle(\pm 16.69)}$
& $67.43\,{\scriptstyle(\pm 18.43)}$
& $85.58\,{\scriptstyle(\pm 7.40)}$
& $39.34\,{\scriptstyle(\pm 1.01)}$
& $39.07\,{\scriptstyle(\pm 1.02)}$
& $98.41\,{\scriptstyle(\pm 1.57)}$
& $\mathbf{99.78\,{\scriptstyle(\pm 0.13)}}$ \\
\addlinespace[2pt]

&
\multirow{4}{*}{Gemma2-9B}
& Utilization $\uparrow$
& $89.65\,{\scriptstyle(\pm 7.05)}$
& $57.61\,{\scriptstyle(\pm 7.25)}$
& $25.65\,{\scriptstyle(\pm 34.93)}$
& $16.35\,{\scriptstyle(\pm 6.04)}$
& $19.28\,{\scriptstyle(\pm 5.45)}$
& $54.65\,{\scriptstyle(\pm 34.54)}$
& $\mathbf{90.13\,{\scriptstyle(\pm 25.28)}}$ \\
&
& Retention $\alpha_1$ $\uparrow$
& $98.67\,{\scriptstyle(\pm 1.10)}$
& $97.68\,{\scriptstyle(\pm 2.17)}$
& $97.87\,{\scriptstyle(\pm 0.96)}$
& $81.58\,{\scriptstyle(\pm 1.64)}$
& $80.58\,{\scriptstyle(\pm 0.72)}$
& $99.79\,{\scriptstyle(\pm 0.20)}$
& $\mathbf{99.97\,{\scriptstyle(\pm 0.01)}}$ \\
&
& Fidelity $\uparrow$
& $65.56\,{\scriptstyle(\pm 17.21)}$
& $70.78\,{\scriptstyle(\pm 15.96)}$
& $74.25\,{\scriptstyle(\pm 22.06)}$
& $87.42\,{\scriptstyle(\pm 7.56)}$
& $87.65\,{\scriptstyle(\pm 7.05)}$
& $93.38\,{\scriptstyle(\pm 3.24)}$
& $\mathbf{96.75\,{\scriptstyle(\pm 1.86)}}$ \\
&
& Stability $\alpha_2$ $\uparrow$
& $91.93\,{\scriptstyle(\pm 5.17)}$
& $89.00\,{\scriptstyle(\pm 8.03)}$
& $89.46\,{\scriptstyle(\pm 4.57)}$
& $49.25\,{\scriptstyle(\pm 1.81)}$
& $48.12\,{\scriptstyle(\pm 0.70)}$
& $97.92\,{\scriptstyle(\pm 1.43)}$
& $\mathbf{99.81\,{\scriptstyle(\pm 0.08)}}$ \\
\addlinespace[2pt]

&
\multirow{4}{*}{Qwen3.5-9B}
& Utilization $\uparrow$
& -- & --
& $17.05\,{\scriptstyle(\pm 10.09)}$
& $28.03\,{\scriptstyle(\pm 8.79)}$
& $29.13\,{\scriptstyle(\pm 8.52)}$
& $81.69\,{\scriptstyle(\pm 33.50)}$
& $\mathbf{97.48\,{\scriptstyle(\pm 6.48)}}$ \\
&
& Retention $\alpha_1$ $\uparrow$
& -- & --
& $96.07\,{\scriptstyle(\pm 2.17)}$
& $83.46\,{\scriptstyle(\pm 0.98)}$
& $83.14\,{\scriptstyle(\pm 0.86)}$
& $95.13\,{\scriptstyle(\pm 5.98)}$
& $\mathbf{99.95\,{\scriptstyle(\pm 0.01)}}$ \\
&
& Fidelity $\uparrow$
& -- & --
& $77.37\,{\scriptstyle(\pm 7.32)}$
& $87.13\,{\scriptstyle(\pm 6.34)}$
& $86.43\,{\scriptstyle(\pm 6.33)}$
& $94.47\,{\scriptstyle(\pm 4.52)}$
& $\mathbf{95.66\,{\scriptstyle(\pm 2.14)}}$ \\
&
& Stability $\alpha_2$ $\uparrow$
& -- & --
& $82.84\,{\scriptstyle(\pm 6.25)}$
& $51.11\,{\scriptstyle(\pm 1.20)}$
& $50.73\,{\scriptstyle(\pm 1.03)}$
& $82.74\,{\scriptstyle(\pm 14.20)}$
& $\mathbf{99.35\,{\scriptstyle(\pm 0.57)}}$ \\

\midrule

% =========================================================
% Wikidata
% =========================================================

\multirow{12}{*}{Wikidata}
& \multirow{4}{*}{Gemma2-2B}
& Utilization $\uparrow$
& $85.54\,{\scriptstyle(\pm 16.78)}$
& $57.47\,{\scriptstyle(\pm 17.53)}$
& $7.65\,{\scriptstyle(\pm 3.29)}$
& $8.12\,{\scriptstyle(\pm 5.45)}$
& $8.49\,{\scriptstyle(\pm 5.41)}$
& $34.68\,{\scriptstyle(\pm 25.10)}$
& $\mathbf{56.91\,{\scriptstyle(\pm 31.55)}}$ \\
&
& Retention $\alpha_1$ $\uparrow$
& $87.81\,{\scriptstyle(\pm 11.63)}$
& $87.12\,{\scriptstyle(\pm 12.58)}$
& $96.40\,{\scriptstyle(\pm 2.07)}$
& $73.14\,{\scriptstyle(\pm 4.43)}$
& $72.05\,{\scriptstyle(\pm 2.93)}$
& $99.70\,{\scriptstyle(\pm 0.22)}$
& $\mathbf{99.87\,{\scriptstyle(\pm 0.17)}}$ \\
&
& Fidelity $\uparrow$
& $72.44\,{\scriptstyle(\pm 6.51)}$
& $76.50\,{\scriptstyle(\pm 5.75)}$
& $87.42\,{\scriptstyle(\pm 5.58)}$
& $92.15\,{\scriptstyle(\pm 4.33)}$
& $92.03\,{\scriptstyle(\pm 3.86)}$
& $93.78\,{\scriptstyle(\pm 2.97)}$
& $\mathbf{95.21\,{\scriptstyle(\pm 3.96)}}$ \\
&
& Stability $\alpha_2$ $\uparrow$
& $70.15\,{\scriptstyle(\pm 16.62)}$
& $69.21\,{\scriptstyle(\pm 17.35)}$
& $84.77\,{\scriptstyle(\pm 6.95)}$
& $42.06\,{\scriptstyle(\pm 3.97)}$
& $40.95\,{\scriptstyle(\pm 2.38)}$
& $96.97\,{\scriptstyle(\pm 2.21)}$
& $\mathbf{98.32\,{\scriptstyle(\pm 2.00)}}$ \\
\addlinespace[2pt]

&
\multirow{4}{*}{Gemma2-9B}
& Utilization $\uparrow$
& $70.94\,{\scriptstyle(\pm 17.20)}$
& $33.89\,{\scriptstyle(\pm 13.23)}$
& $8.07\,{\scriptstyle(\pm 3.90)}$
& $8.10\,{\scriptstyle(\pm 4.88)}$
& $10.91\,{\scriptstyle(\pm 4.07)}$
& $25.68\,{\scriptstyle(\pm 12.69)}$
& $\mathbf{68.81\,{\scriptstyle(\pm 24.06)}}$ \\
&
& Retention $\alpha_1$ $\uparrow$
& $98.46\,{\scriptstyle(\pm 2.64)}$
& $96.94\,{\scriptstyle(\pm 5.51)}$
& $96.84\,{\scriptstyle(\pm 3.07)}$
& $85.29\,{\scriptstyle(\pm 3.77)}$
& $82.31\,{\scriptstyle(\pm 1.34)}$
& $98.78\,{\scriptstyle(\pm 1.53)}$
& $\mathbf{99.94\,{\scriptstyle(\pm 0.03)}}$ \\
&
& Fidelity $\uparrow$
& $67.81\,{\scriptstyle(\pm 14.09)}$
& $73.76\,{\scriptstyle(\pm 12.01)}$
& $86.72\,{\scriptstyle(\pm 9.09)}$
& $90.57\,{\scriptstyle(\pm 6.68)}$
& $91.19\,{\scriptstyle(\pm 6.35)}$
& $92.52\,{\scriptstyle(\pm 4.49)}$
& $\mathbf{96.85\,{\scriptstyle(\pm 1.39)}}$ \\
&
& Stability $\alpha_2$ $\uparrow$
& $92.47\,{\scriptstyle(\pm 9.33)}$
& $89.53\,{\scriptstyle(\pm 13.16)}$
& $86.18\,{\scriptstyle(\pm 9.03)}$
& $53.69\,{\scriptstyle(\pm 4.78)}$
& $49.90\,{\scriptstyle(\pm 1.41)}$
& $93.70\,{\scriptstyle(\pm 6.48)}$
& $\mathbf{99.11\,{\scriptstyle(\pm 1.02)}}$ \\
\addlinespace[2pt]

&
\multirow{4}{*}{Olmo-Hybrid-7B}
& Utilization $\uparrow$
& -- & --
& $13.72\,{\scriptstyle(\pm 10.60)}$
& $18.00\,{\scriptstyle(\pm 15.33)}$
& $22.52\,{\scriptstyle(\pm 12.38)}$
& $46.94\,{\scriptstyle(\pm 29.97)}$
& $\mathbf{86.61\,{\scriptstyle(\pm 32.96)}}$ \\
&
& Retention $\alpha_1$ $\uparrow$
& -- & --
& $93.34\,{\scriptstyle(\pm 5.63)}$
& $78.03\,{\scriptstyle(\pm 11.11)}$
& $81.86\,{\scriptstyle(\pm 1.13)}$
& $98.04\,{\scriptstyle(\pm 1.84)}$
& $\mathbf{99.95\,{\scriptstyle(\pm 0.03)}}$ \\
&
& Fidelity $\uparrow$
& -- & --
& $80.04\,{\scriptstyle(\pm 12.38)}$
& $86.18\,{\scriptstyle(\pm 9.35)}$
& $86.60\,{\scriptstyle(\pm 8.52)}$
& $90.29\,{\scriptstyle(\pm 6.87)}$
& $\mathbf{96.61\,{\scriptstyle(\pm 1.70)}}$ \\
&
& Stability $\alpha_2$ $\uparrow$
& -- & --
& $79.04\,{\scriptstyle(\pm 12.59)}$
& $46.80\,{\scriptstyle(\pm 8.05)}$
& $49.31\,{\scriptstyle(\pm 1.24)}$
& $90.04\,{\scriptstyle(\pm 7.57)}$
& $\mathbf{98.56\,{\scriptstyle(\pm 1.62)}}$ \\

\midrule

% =========================================================
% LAION-COCO
% =========================================================

\multirow{8}{*}{LAION-COCO}
& \multirow{4}{*}{Stable Diffusion v1.5}
& Utilization $\uparrow$
& -- & --
& $5.79\,{\scriptstyle(\pm 8.96)}$
& $9.67\,{\scriptstyle(\pm 3.74)}$
& $9.59\,{\scriptstyle(\pm 3.84)}$
& $13.96\,{\scriptstyle(\pm 7.99)}$
& $\mathbf{59.46\,{\scriptstyle(\pm 46.56)}}$ \\
&
& Retention $\alpha_1$ $\uparrow$
& -- & --
& $91.85\,{\scriptstyle(\pm 8.65)}$
& $55.81\,{\scriptstyle(\pm 0.73)}$
& $55.41\,{\scriptstyle(\pm 0.62)}$
& $96.29\,{\scriptstyle(\pm 9.17)}$
& $\mathbf{99.89\,{\scriptstyle(\pm 0.26)}}$ \\
&
& Fidelity $\uparrow$
& -- & --
& $63.77\,{\scriptstyle(\pm 8.90)}$
& $81.04\,{\scriptstyle(\pm 18.14)}$
& $80.06\,{\scriptstyle(\pm 16.34)}$
& $75.22\,{\scriptstyle(\pm 22.51)}$
& $\mathbf{85.50\,{\scriptstyle(\pm 21.18)}}$ \\
&
& Stability $\alpha_2$ $\uparrow$
& -- & --
& $72.94\,{\scriptstyle(\pm 14.41)}$
& $29.71\,{\scriptstyle(\pm 0.49)}$
& $29.44\,{\scriptstyle(\pm 0.40)}$
& $82.86\,{\scriptstyle(\pm 28.67)}$
& $\mathbf{98.52\,{\scriptstyle(\pm 4.14)}}$ \\
\addlinespace[2pt]

&
\multirow{4}{*}{FLUX.2-klein-9B}
& Utilization $\uparrow$
& -- & --
& $1.13\,{\scriptstyle(\pm 0.82)}$
& $2.40\,{\scriptstyle(\pm 2.91)}$
& $13.73\,{\scriptstyle(\pm 15.20)}$
& $3.17\,{\scriptstyle(\pm 4.39)}$
& $\mathbf{65.00\,{\scriptstyle(\pm 46.48)}}$ \\
&
& Retention $\alpha_1$ $\uparrow$
& -- & --
& $92.79\,{\scriptstyle(\pm 9.08)}$
& $68.29\,{\scriptstyle(\pm 17.77)}$
& $82.85\,{\scriptstyle(\pm 11.77)}$
& $84.14\,{\scriptstyle(\pm 7.90)}$
& $\mathbf{99.86\,{\scriptstyle(\pm 0.28)}}$ \\
&
& Fidelity $\uparrow$
& -- & --
& $80.61\,{\scriptstyle(\pm 11.97)}$
& $88.33\,{\scriptstyle(\pm 8.43)}$
& $90.05\,{\scriptstyle(\pm 9.11)}$
& $87.43\,{\scriptstyle(\pm 7.18)}$
& $\mathbf{96.03\,{\scriptstyle(\pm 2.37)}}$ \\
&
& Stability $\alpha_2$ $\uparrow$
& -- & --
& $81.83\,{\scriptstyle(\pm 18.80)}$
& $41.56\,{\scriptstyle(\pm 14.95)}$
& $53.56\,{\scriptstyle(\pm 12.21)}$
& $64.12\,{\scriptstyle(\pm 10.20)}$
& $\mathbf{98.47\,{\scriptstyle(\pm 2.48)}}$ \\

\bottomrule
\end{tabular}
}

\vspace{2pt}

\begin{minipage}{\textwidth}
\footnotesize
Values are reported as mean followed by the half-width of the two-sided 95\% \(t\)-confidence interval in parentheses, and all values are scaled by 100. Higher values indicate better performance. \textbf{Bold} indicates the highest mean value in each row. Utilization denotes the fraction of active SAE features and is reported only for the trained SAEs because GemmaScope and the trained SAEs use different SAE widths, which would confound direct comparisons of feature utilization. Retention \(\alpha_1\) is computed from the coherence condition \(\mu_q < 1/(K_q-1)\), whereas stability \(\alpha_2\) is computed from the distinguishability condition \(\mu_q < 1/(2K_q-1)\). For Wikipedia and Wikidata, confidence intervals are computed across six evaluated model layers. For LAION-COCO, confidence intervals are computed across three diffusion timesteps for Stable Diffusion v1.5 and across six layer-step combinations for FLUX.2-klein-9B. GemmaScope Small denotes width 16k; GemmaScope Large denotes width 65k for Gemma2-2B and width 131k for Gemma2-9B. TopK and AbsTopK use \(K=50\). Dashes indicate unavailable or non-comparable evaluations.
\end{minipage}

\end{table*}

\clearpage
\begin{table*}[t]
\centering
\caption{\textbf{Effect of the sparsity parameter \(K\) on the fidelity and stability of TopK SAEs.}}
\label{tab:topk_k_ablation}

\small
\setlength{\tabcolsep}{8pt}
\renewcommand{\arraystretch}{1.08}

\begin{tabular}{@{}lccc@{}}
\toprule
Model & \(K\) & Fidelity $\uparrow$ & Stability $\uparrow$ \\
\midrule

\multirow{2}{*}{Gemma2-2B}
& 20 & $79.61\,{\scriptstyle(\pm 11.86)}$
     & $84.86\,{\scriptstyle(\pm 3.81)}$ \\
& 50 & $88.68\,{\scriptstyle(\pm 7.05)}$
     & $39.34\,{\scriptstyle(\pm 1.01)}$ \\

\addlinespace[2pt]

\multirow{2}{*}{Gemma2-9B}
& 20 & $78.66\,{\scriptstyle(\pm 11.87)}$
     & $93.39\,{\scriptstyle(\pm 1.91)}$ \\
& 50 & $87.42\,{\scriptstyle(\pm 7.56)}$
     & $49.25\,{\scriptstyle(\pm 1.81)}$ \\

\bottomrule
\end{tabular}

\vspace{2pt}

\begin{minipage}{\textwidth}
\footnotesize
All results are obtained using TopK SAEs trained on representations from Wikipedia. \(K\) denotes the number of latent feature activations retained for each input by the TopK activation. Values are reported as mean followed by the half-width of the two-sided 95\% \(t\)-confidence interval in parentheses, and all values are scaled by 100.
\end{minipage}

\end{table*}

\clearpage
\begin{table*}[t]
\centering
\caption{\textbf{Semantic recovery of single atom and atom combinations across models.}}
\label{tab:sememe_alignment_prf_ci95}

\scriptsize
\setlength{\tabcolsep}{3.5pt}
\renewcommand{\arraystretch}{1.10}

\resizebox{\textwidth}{!}{
\begin{tabular}{@{}lcccccccccc@{}}
\toprule
\multirow{2}{*}{Model}
& \multicolumn{3}{c}{Best single atom}
& \multicolumn{3}{c}{Atom combination}
& \multicolumn{3}{c}{Difference}
& \multirow{2}{*}{Mean \# atoms} \\
\cmidrule(lr){2-4}
\cmidrule(lr){5-7}
\cmidrule(lr){8-10}
& Precision
& Recall
& F1
& Precision
& Recall
& F1
& $\Delta$ precision
& $\Delta$ recall
& $\Delta$ F1
& \\
\midrule

Gemma2-2B
& $0.691\,{\scriptstyle(\pm 0.011)}$
& $0.232\,{\scriptstyle(\pm 0.008)}$
& $0.314\,{\scriptstyle(\pm 0.008)}$
& $0.691\,{\scriptstyle(\pm 0.004)}$
& $0.989\,{\scriptstyle(\pm 0.002)}$
& $0.809\,{\scriptstyle(\pm 0.003)}$
& $+0.000\,{\scriptstyle(\pm 0.011)}$
& $+0.757\,{\scriptstyle(\pm 0.008)}$
& $+0.495\,{\scriptstyle(\pm 0.008)}$
& $62.02\,{\scriptstyle(\pm 3.87)}$ \\

Gemma2-9B
& $0.682\,{\scriptstyle(\pm 0.010)}$
& $0.268\,{\scriptstyle(\pm 0.009)}$
& $0.349\,{\scriptstyle(\pm 0.009)}$
& $0.710\,{\scriptstyle(\pm 0.004)}$
& $0.986\,{\scriptstyle(\pm 0.002)}$
& $0.820\,{\scriptstyle(\pm 0.003)}$
& $+0.028\,{\scriptstyle(\pm 0.010)}$
& $+0.718\,{\scriptstyle(\pm 0.009)}$
& $+0.470\,{\scriptstyle(\pm 0.008)}$
& $54.41\,{\scriptstyle(\pm 3.39)}$ \\

Llama3.1-8B
& $0.624\,{\scriptstyle(\pm 0.011)}$
& $0.266\,{\scriptstyle(\pm 0.008)}$
& $0.340\,{\scriptstyle(\pm 0.008)}$
& $0.672\,{\scriptstyle(\pm 0.004)}$
& $0.954\,{\scriptstyle(\pm 0.003)}$
& $0.783\,{\scriptstyle(\pm 0.003)}$
& $+0.048\,{\scriptstyle(\pm 0.011)}$
& $+0.688\,{\scriptstyle(\pm 0.008)}$
& $+0.443\,{\scriptstyle(\pm 0.007)}$
& $57.55\,{\scriptstyle(\pm 3.59)}$ \\

\bottomrule
\end{tabular}
}

\vspace{2pt}

\begin{minipage}{\textwidth}
\footnotesize
Values are reported as mean followed by the half-width of the two-sided 95\% \(t\)-confidence interval in parentheses, computed across evaluated HowNet sememes. Precision, recall and F1 quantify the set match between each sememe and either its best individual representational atom or the selected atom combination. Differences (\(\Delta\)) are computed for each sememe as the atom-combination value minus the corresponding best-single-atom value before calculating the mean and confidence interval. Mean \# atoms denotes the mean number of atoms selected for compositional recovery.
\end{minipage}

\end{table*}

\clearpage
\begin{table*}[t]
\centering
\caption{\textbf{Evaluation of BiJumpReLU SAE representations on HowNet across models.}}
\label{tab:hownet_sae_evaluation_mean}

\scriptsize
\setlength{\tabcolsep}{8pt}
\renewcommand{\arraystretch}{1.10}

\begin{tabular}{@{}lcccc@{}}
\toprule
Model
& Utilization
& Retention
& \textbf{Fidelity}
& \textbf{Stability} \\
\midrule

Gemma2-2B
& $89.60\,{\scriptstyle(\pm 3.94)}$
& $99.98\,{\scriptstyle(\pm 0.01)}$
& $\mathbf{99.62\,{\scriptstyle(\pm 0.10)}}$
& $\mathbf{99.82\,{\scriptstyle(\pm 0.04)}}$ \\

Gemma2-9B
& $93.45\,{\scriptstyle(\pm 2.77)}$
& $99.99\,{\scriptstyle(\pm 0.00)}$
& $\mathbf{99.64\,{\scriptstyle(\pm 0.08)}}$
& $\mathbf{99.91\,{\scriptstyle(\pm 0.02)}}$ \\

Llama3.1-8B
& $86.59\,{\scriptstyle(\pm 10.48)}$
& $99.98\,{\scriptstyle(\pm 0.01)}$
& $\mathbf{99.30\,{\scriptstyle(\pm 0.48)}}$
& $\mathbf{99.89\,{\scriptstyle(\pm 0.05)}}$ \\

\bottomrule
\end{tabular}

\vspace{2pt}

\begin{minipage}{\textwidth}
\footnotesize
Values are reported as the mean followed by the half-width of the two-sided 95\% \(t\)-confidence interval in parentheses, computed across layers. All values are scaled by 100, and \textbf{bold} values denote the two key evaluation criteria for representational atoms: fidelity and stability. Utilization, retention, fidelity and stability are defined as in the preceding evaluation.
\end{minipage}

\end{table*}

\clearpage
\begin{table*}[p]
\centering
\caption{\textbf{Case studies of sememe recovery by atom combinations.}}
\label{tab:sememe_case_study}

\sffamily
\fontsize{5.2}{6.0}\selectfont
\setlength{\tabcolsep}{0.8pt}
\renewcommand{\arraystretch}{1.08}

\begin{tabular}{@{}
    >{\centering\arraybackslash}m{1.00cm}
    >{\centering\arraybackslash}m{1.10cm}
    >{\centering\arraybackslash}m{0.80cm}
    >{\centering\arraybackslash}m{0.90cm}
    >{\centering\arraybackslash}m{1.45cm}
    >{\centering\arraybackslash}m{1.00cm}
    >{\centering\arraybackslash}m{0.40cm}
    >{\centering\arraybackslash}m{0.50cm}
    >{\raggedright\arraybackslash}m{8.36cm}
@{}}

\toprule
Sememe
& Model
& \# atoms
& $|S|/|A|$
& P/R/F1
& Difference
& Size
& Marked
& \multicolumn{1}{c}{Entities} \\
\midrule

% ============================================================
% DISTRICT — Gemma2-2B
% ============================================================

\multirow[c]{2}{=}{DISTRICT}
& \multirow[c]{2}{=}{Gemma2-2B}
& \multirow[c]{2}{=}{49}
& \multirow[c]{2}{=}{1076/1110}
& \multirow[c]{2}{=}{0.968/0.998/0.983}
& $A\setminus S$
& 36
& 24
& \textbf{Afar}; BYD; \textbf{Bhutan}; \textbf{Hongqiao};
\textbf{Huangbeiling}; \textbf{Jiaozhou}; \textbf{Kabukicho};
\textbf{Keshan}; \textbf{Lujiazui}; \textbf{Lukou Street};
\textbf{Male}; \textbf{Mount Songshan}; \textbf{Nankang}; O;
\textbf{Panyu}; \textbf{Pingshan}; \textbf{Potala};
\textbf{Sanba Square}; \textbf{Shengzhou}; \textbf{Tongzhou};
\textbf{Yalta}; analytics; attention; call on a hotline;
\textbf{city of Beijing}; coleorhizae; counterspy;
couple with big age gap; friend; hardtack;
\textbf{newly added district}; protease; \textbf{regional};
schoolhouse; \textbf{small island}; \textbf{urban district}
\\

& & & & &
$S\setminus A$
& 2
& 1
& \textbf{\texttt{""}}; district League committee
\\

\addlinespace[0.6pt]

% ============================================================
% DISTRICT — Gemma2-9B
% ============================================================

\multirow[c]{2}{=}{DISTRICT}
& \multirow[c]{2}{=}{Gemma2-9B}
& \multirow[c]{2}{=}{28}
& \multirow[c]{2}{=}{1076/1050}
& \multirow[c]{2}{=}{1.000/0.976/0.988}
& $A\setminus S$
& 0
& 0
& -- \\

& & & & &
$S\setminus A$
& 26
& 1
& \texttt{""}; Baoan; Beitou; Bund; Honghuagang; Mentougou;
Nangang; Neihu; Potomac; Qingshan; Tongan; Wanhua; Wujin;
Xiangzhou; Xindu; Zengcheng; area; area code; borough; canton;
district League committee; district and county; districts of Beijing;
region; troublespot; whole district
\\

\addlinespace[0.6pt]

% ============================================================
% DISTRICT — Llama3.1-8B
% ============================================================

\multirow[c]{2}{=}{DISTRICT}
& \multirow[c]{2}{=}{Llama3.1-8B}
& \multirow[c]{2}{=}{50}
& \multirow[c]{2}{=}{1076/1114}
& \multirow[c]{2}{=}{0.963/0.997/0.980}
& $A\setminus S$
& 41
& 27
& \textbf{Aoshan}; \textbf{Baboshan}; \textbf{Bank of Chaoyang};
Chinese chequers; \textbf{Columbia}; \textbf{Fengcheng};
\textbf{Fujin}; \textbf{Funing}; \textbf{Hangjinqi}; \textbf{Hindu};
\textbf{Hongqiao}; \textbf{Jiading}; \textbf{Kabukicho};
\textbf{Liangzhou}; \textbf{Linzhi}; \textbf{Maigaoqiao};
Melanogrammus aeglefinus; \textbf{Nankang};
\textbf{National Westminster}; \textbf{Peking};
\textbf{Quanjude}; \textbf{Red Square}; \textbf{Sanba Square};
\textbf{Tongzhou}; \textbf{Xingtai}; Yuwen; and then; be uncertain;
\textbf{city and county}; \textbf{city of Beijing}; fluorocarbon;
footballer; force-fed duck; frontage house; itinerant doctor;
meteorology; object clause; phantasy; \textbf{racecourse};
\textbf{subarea}; utter a threat
\\

& & & & &
$S\setminus A$
& 3
& 1
& \textbf{\texttt{""}}; district; region
\\

\addlinespace[1.0pt]

% ============================================================
% MALAWI
% ============================================================

\multirow[c]{2}{=}{MALAWI}
& \multirow[c]{2}{=}{Gemma2-2B}
& \multirow[c]{2}{=}{2}
& \multirow[c]{2}{=}{8/9}
& \multirow[c]{2}{=}{0.889/1.000/0.941}
& $A\setminus S$
& 1
& 1
& \textbf{Kwacha}
\\

& & & & &
$S\setminus A$
& 0
& 0
& --
\\

\addlinespace[0.6pt]

\multirow[c]{2}{=}{MALAWI}
& \multirow[c]{2}{=}{Gemma2-9B}
& \multirow[c]{2}{=}{2}
& \multirow[c]{2}{=}{8/9}
& \multirow[c]{2}{=}{0.889/1.000/0.941}
& $A\setminus S$
& 1
& 1
& \textbf{Kwacha}
\\

& & & & &
$S\setminus A$
& 0
& 0
& --
\\

\addlinespace[1.0pt]

% ============================================================
% MORNING
% ============================================================

\multirow[c]{2}{=}{MORNING}
& \multirow[c]{2}{=}{Gemma2-2B}
& \multirow[c]{2}{=}{40}
& \multirow[c]{2}{=}{113/133}
& \multirow[c]{2}{=}{0.850/1.000/0.919}
& $A\setminus S$
& 20
& 7
& Heshun; bagel; be up to date;
\textbf{call the roll in the morning}; cryptanalysis;
do not speak; early-wake; evening peak;
\textbf{get out of bed}; \textbf{get up early};
\textbf{just dawning}; lunch; \textbf{matutinal}; night job;
quick-meal stall; \textbf{rosy cloud}; sunset;
try to catch a train;
\textbf{whitish colour of a fish's belly-grey dawn};
work night and day
\\

& & & & &
$S\setminus A$
& 0
& 0
& --
\\

\addlinespace[0.6pt]

\multirow[c]{2}{=}{MORNING}
& \multirow[c]{2}{=}{Gemma2-9B}
& \multirow[c]{2}{=}{34}
& \multirow[c]{2}{=}{113/132}
& \multirow[c]{2}{=}{0.856/1.000/0.922}
& $A\setminus S$
& 19
& 5
& \textbf{brilliance of the rising sun}; elevenses; evening peak;
evening running; evening show; \textbf{get up early}; hypnone;
\textbf{matutinal}; \textbf{on the morning of}; p.m.;
\textbf{period of the day from 9 a.m to 11 a.m}; post meridiem;
quick-lunch; quick-meal stall; signal flare; teatime; time of day;
tourist coach; work night and day
\\

& & & & &
$S\setminus A$
& 0
& 0
& --
\\

\addlinespace[1.0pt]

% ============================================================
% CAPITAL
% ============================================================

\multirow[c]{2}{=}{CAPITAL}
& \multirow[c]{2}{=}{Gemma2-9B}
& \multirow[c]{2}{=}{126}
& \multirow[c]{2}{=}{619/609}
& \multirow[c]{2}{=}{1.000/0.984/0.992}
& $A\setminus S$
& 0
& 0
& --
\\

& & & & &
$S\setminus A$
& 10
& 4
& Ancient Rome; BERIN; \textbf{Bata};
\textbf{Beijing, Shanghai, Guangzhou and Shenzhen};
Chang'an; \textbf{Nine-Gates Provincial Military Commander};
Victoria; \textbf{cockneyfy};
guard the capital and the imperial palace;
move the capital to another place
\\

\addlinespace[1.0pt]

% ============================================================
% BACTERIA
% ============================================================

\multirow[c]{2}{=}{BACTERIA}
& \multirow[c]{2}{=}{Gemma2-2B}
& \multirow[c]{2}{=}{109}
& \multirow[c]{2}{=}{371/419}
& \multirow[c]{2}{=}{0.883/0.997/0.937}
& $A\setminus S$
& 49
& 5
& A/H1N1; \textbf{Bartonella}; CTA; Heilbronn; ISO; PAHs;
\textbf{animalcula}; announcer; archesporia; basidia; bluetongue;
business zone; campaign for president; causal agent; cicadae;
\textbf{clostridia}; dicker; distiller's yeast; echinococcus;
\textbf{enterohemorrhagic}; erysipelas; free port; game theory;
hardtack liquor made with a kind of yeast; herpes; it be alleged;
liquid; lutetium; medroxalol; microbiology; \textbf{moneran};
natatorium; neonicotinoid; pogrom; protophyte; pullulate; snare;
sol; state funeral; sterility; subcommission; subfunction;
sulfadiazine; sulfamethazine; toleration; toxin; tubercle;
waste gas or steam; zyme
\\

& & & & &
$S\setminus A$
& 1
& 1
& \textbf{carrier}
\\

\addlinespace[1.0pt]

% ============================================================
% DAY
% ============================================================

\multirow[c]{2}{=}{DAY}
& \multirow[c]{2}{=}{Gemma2-2B}
& \multirow[c]{2}{=}{166}
& \multirow[c]{2}{=}{761/787}
& \multirow[c]{2}{=}{0.879/0.909/0.894}
& $A\setminus S$
& 95
& 45
& \textbf{April fool}; \textbf{Boxing Day}; \textbf{Chinese New Year};
\textbf{Christmas Day}; \textbf{Christmas Eve}; \textbf{Christmas eve};
\textbf{Eid al-Fitr}; \textbf{Hallowe'en}; \textbf{Halloween};
\textbf{Happy Chinese New Year}; \textbf{Happy Valentine's Day};
Lord's Supper; \textbf{Lunar New Year}; Muharram;
\textbf{September 11th}; Vesper; all afternoon; all evening;
all morning; \textbf{another day}; \textbf{bank holiday};
be up drinking all night; become more prosperous every day;
\textbf{bicennaries}; \textbf{bimillenary}; \textbf{bimillennium};
\textbf{boxing day}; call the roll in the morning;
\textbf{change with each passing day}; day show; \textbf{day-by-day};
early-wake; elevenses; \textbf{every Friday}; \textbf{every Monday};
\textbf{every Saturday}; \textbf{every Sunday};
\textbf{every Thursday}; \textbf{every Tuesday};
\textbf{every Wednesday}; express good wishes; \textbf{film festival};
film week; first month of the lunar year; for years on end;
get up early; glow; \textbf{good day}; group attending a celebration;
\textbf{have a day off}; \textbf{in a couple of days};
\textbf{in a few days}; \textbf{in one day}; in one morning or evening;
\textbf{in these days}; informal dinner;
intercalary month in the lunar calendar; lie awake all night;
lodge for the night; \textbf{market day}; matinee; matutinal;
new calendar; night flight; night school; nightbird;
of the previous night; one morning; overslept;
\textbf{period of the day from 1 a.m. to 3 a.m.};
\textbf{period of the day from 3 a.m. to 5 a.m.};
\textbf{period of the day from 9 a.m to 11 a.m}; pigeonhole;
rays of morning or evening sunlight; same evening;
second month of the lunar year; setting sun; sleep all afternoon;
sleep all morning; solar calendar; \textbf{stay a few days};
stay up the whole night; that very evening; that very noon;
\textbf{today and yesterday}; travel by night; warming sunlight;
wed; weeknight; \textbf{what day is tomorrow};
\textbf{what's the date tomorrow}; \textbf{which day};
whitish colour of a fish's belly-grey dawn;
\textbf{within the next few days}; working lunch
\\

& & & & &
$S\setminus A$
& 69
& 7
& Dec. 12th, 1936; Frost's descent; Grain Full; Grain Rain;
Grain in Ear; \textbf{Islam Corban}; at the end of the week;
birthday gift; breakfast house; breakfast stall; by the due date;
curfew; daily average; daily journal; daily produce;
daily salt intake; date-changing phase; dateless; dateline;
datemark; dawning; day; day of rest for factory workers;
day of retribution; day off for staff as specified by the factory;
dies faustus; \textbf{endorse}; eve; evening class; eventide;
expected date of childbirth; forenone; get up in the night to urinate;
grow with the passing yearss; holiday; journal; logbook;
lunar calendar used by the Zang nationality; lunch meeting;
luncheon meeting; menology; midday meal; millenary; millennium;
misdate; \textbf{monkey jacket}; morning coach; morning train;
nightly; nyctalopia; on the wane; pass the night; per diem pay;
red-eye; rosy clouds; seasonal division point;
\textbf{shell jacket}; sky grows dark; \textbf{solar flare};
statutory holiday; three days from now; tiffin;
travel day and night; \textbf{tux}; \textbf{tuxedo};
twenty-four hours; undated; vigil; white night
\\

\addlinespace[1.0pt]

% ============================================================
% FEEL
% ============================================================

\multirow[c]{2}{=}{FEEL}
& \multirow[c]{2}{=}{Gemma2-2B}
& \multirow[c]{2}{=}{5}
& \multirow[c]{2}{=}{9/13}
& \multirow[c]{2}{=}{0.692/1.000/0.818}
& $A\setminus S$
& 4
& 4
& \textbf{dermal sensation}; \textbf{pictural};
\textbf{somatosense}; \textbf{suddenly feel}
\\

& & & & &
$S\setminus A$
& 0
& 0
& --
\\

\addlinespace[1.0pt]

% ============================================================
% BANGLADESH
% ============================================================

\multirow[c]{2}{=}{BANGLAD}
& \multirow[c]{2}{=}{Llama3.1-8B}
& \multirow[c]{2}{=}{5}
& \multirow[c]{2}{=}{14/16}
& \multirow[c]{2}{=}{0.875/1.000/0.933}
& $A\setminus S$
& 2
& 1
& \textbf{Liberation War}; President Truman
\\

& & & & &
$S\setminus A$
& 0
& 0
& --
\\

\bottomrule
\end{tabular}

\vspace{2pt}

\begin{minipage}{0.98\textwidth}
\fontsize{5.2}{6.0}\selectfont
\sffamily
$S$ denotes the HowNet entity set annotated with the target sememe, and $A$ denotes the entity set recovered by the selected atom combination. P/R/F1 denote precision, recall and F1 for the set match between $A$ and $S$. For $A\setminus S$, bold entities were manually judged to be clearly consistent with the target sememe despite being absent from HowNet, indicating potential missing annotations. For $S\setminus A$, bold entities were manually judged to be clearly inconsistent with the target sememe despite being included in HowNet, indicating potential incorrect annotations. Marked gives the number of such manually identified entities in each difference set. \texttt{""} denotes an empty entity string. Only clear cases were marked during manual inspection.
\end{minipage}

\end{table*}

% Restore normal float-page layout
\makeatletter
\setlength{\@fptop}{\oldfptop}
\setlength{\@fpbot}{\oldfpbot}
\setlength{\@dblfptop}{\olddblfptop}
\setlength{\@dblfpbot}{\olddblfpbot}
\makeatother

\clearpage

% Restart page numbering for Supplementary Information
\pagenumbering{arabic}

\hypersetup{bookmarksdepth=2}

\begin{appendices}

% Do not number supplementary sections/subsections
\setcounter{secnumdepth}{0}

% Start a separate table of contents for the appendix
\startcontents[appendix]
\begin{center}
    \Large\bfseries Supplementary Information
\end{center}
\vspace{1em}
\setcounter{tocdepth}{2}
\begingroup
\makeatletter

% Keep the original subsection TOC style, only add indentation
\let\oldl@subsection\l@subsection
\renewcommand{\l@subsection}[2]{%
    \begingroup
    \setlength{\leftskip}{1.5em}%
    \oldl@subsection{#1}{#2}%
    \endgroup
}

\printcontents[appendix]{}{1}{}

\makeatother
\endgroup
\clearpage

% Beginning of Supplementary Information
\setcounter{table}{0}
\renewcommand{\tablename}{Supplementary Table}

\setcounter{figure}{0}
\renewcommand{\figurename}{Supplementary Fig.}




\section{Proofs}\label{Section:Proofs}

\subsection{Proof of Theorem~\ref{theorem:Explicit_Form}}
\ExplicitForm*
\begin{proof}
    Since \(\langle \cdot, \cdot \rangle_S\) is atomic inner product, for any pair of atoms \(\boldsymbol d_i\) and \(\boldsymbol d_j\) we have
    \begin{equation}
        \boldsymbol d_i^{\top} S \boldsymbol d_j \;=\;\langle \boldsymbol d_i, \boldsymbol d_j \rangle_{S}=\begin{cases}0 & i \ne j, \\[4pt]c^2 & i = j .\end{cases}
    \end{equation}
    Applying this property to the atom set \(D=[\boldsymbol d_1,\boldsymbol d_2,\cdots,\boldsymbol d_{|D|}]\) yields
    \begin{equation}
        c^2 I = D^\top S D .
    \end{equation}
    Since \(\{\boldsymbol d_1,\boldsymbol d_2,\cdots,\boldsymbol d_{|D|}\}\) spans the atomic space \(\mathcal D\), and \(\mathcal D \simeq \mathbb R^H\), it follows that \(\mathrm{rank}(D)=H\). Let \(S^{1/2}\) denote the symmetric positive-definite square root of \(S\). Then
    \begin{equation}
        D^\top S D = (S^{1/2} D)^\top (S^{1/2} D),
    \end{equation}
    which implies \(\mathrm{rank}(I) = \mathrm{rank}(D^\top S D) = \mathrm{rank}(S^{1/2} D) = \mathrm{rank}(D)\). Therefore, \(\mathrm{rank}(D) = |D| \leq H\). Combining this with the earlier condition gives \(|D| = \mathrm{rank}(D) = H\), which shows that \(D\) is invertible. Consequently,
    \begin{equation}
        S = c^2 (D D^\top)^{-1}.
    \end{equation}
\end{proof}

\subsection{Proof of Corollary~\ref{corollary:NAIP}}
\NAIP*
\begin{proof}
    Since \(\langle \boldsymbol{d}_i, \boldsymbol{d}_j \rangle_S = 0\) for \(i \neq j\), and \(\|\boldsymbol{d}_i\|_S = \|\boldsymbol{d}_j\|_S = c > 0\), it follows that \(D^\top S D = c^2 I\). Therefore, \(\tilde{S} = \frac{1}{c^2} S\) satisfies \(D^\top \tilde{S} D = I\), which implies that the atoms are orthonormal. Since $D$ is invertible, we also have \(\tilde{S} = D^{-\top} D^{-1} = (D D^\top)^{-1}\), and thus \(\langle \cdot, \cdot \rangle_{\tilde{S}}\) is a symmetric positive-definite inner product.
\end{proof}

\subsection{Proof of Theorem~\ref{theorem:Coherence–RIP_Upper_Bound}}
\RIPUpperBound*
\begin{proof}
    Let \(\mathrm{supp}(\boldsymbol \delta) = \mathcal{S} \subseteq [|D|]\) and \(|\mathcal{S}| \leq K\). Then, we have
    \begin{equation}
        \|\tilde{D} \boldsymbol{\delta}\|_2^2 = \boldsymbol{\delta}^\top (\tilde{D}^\top \tilde{D}) \boldsymbol{\delta} = \sum_{i \in \mathcal{S}} \delta_i^2 + 2 \sum_{\substack{i < j \\ i, j \in \mathcal{S}}} \delta_i \delta_j \langle \tilde{\boldsymbol{d}_i}, \tilde{\boldsymbol{d}_j} \rangle.
    \end{equation}
    By the fact that \(|\langle \tilde{\boldsymbol{d}_i}, \tilde{\boldsymbol{d}_j} \rangle| \leq \mu\) and applying the triangle inequality, we obtain:
    \begin{align}
        \|\tilde{D} \boldsymbol{\delta}\|_2^2 & \geq \sum_{i \in \mathcal{S}} \delta_i^2 - 2 \mu \sum_{i < j} |\delta_i \delta_j|, \\
        \|\tilde{D} \boldsymbol{\delta}\|_2^2 & \leq \sum_{i \in \mathcal{S}} \delta_i^2 + 2 \mu \sum_{i < j} |\delta_i \delta_j|.
    \end{align}
    Next, we observe that:
    \begin{equation}
        \left( \sum_{i \in \mathcal{S}} |\delta_i| \right)^2 = \sum_{i \in \mathcal{S}} \delta_i^2 + 2 \sum_{i < j} |\delta_i \delta_j| \leq |\mathcal{S}| \sum_{i \in \mathcal{S}} \delta_i^2 \leq K \sum_{i \in \mathcal{S}} \delta_i^2.
    \end{equation}
    Thus, we have \(2 \sum_{i < j} |\delta_i \delta_j| \leq (K - 1) \sum_{i \in \mathcal{S}} \delta_i^2\). Substituting this back, we conclude the proof.
\end{proof}

\subsection{Proof of Theorem~\ref{theorem:Uniqueness_of_Sparse_Representation_and_Exact_Recovery}}
\UniquenessRecovery*
\begin{proof}
    We first prove that under the condition $\mu < \tfrac{1}{2K-1}$, the $K$-sparse representation is unique.

    Suppose there exist two distinct \(K\)-sparse coefficient vectors \(\boldsymbol \delta, \boldsymbol \delta'\) such that \(\tilde D \boldsymbol \delta = \tilde D \boldsymbol \delta'\). Let \(\boldsymbol h = \boldsymbol \delta - \boldsymbol \delta' \neq \boldsymbol 0\). Then \(\tilde D \boldsymbol h = 0\) and \(\|\boldsymbol h\|_0 \leq 2K\). By Theorem~\ref{theorem:Coherence–RIP_Upper_Bound} (applied with \(K\) replaced by \(2K\)), we have
    \begin{equation}
        \big(1 - (2K-1)\mu \big)\,\|\boldsymbol h\|_2^2\ \leq\ \|\tilde D \boldsymbol h\|_2^2\ =\ 0.
    \end{equation}
    If $\mu < \tfrac{1}{2K-1}$, then the prefactor on the left is strictly positive, which forces $\|\boldsymbol h\|_2=0$. This contradicts $\boldsymbol h \neq \boldsymbol 0$. Hence uniqueness holds.

    Next, we prove that under the same condition $\mu < \tfrac{1}{2K-1}$, the sparse vector \(\boldsymbol \delta\) is also the unique solution of the convex optimization problem
    \begin{equation}
        \min_{\boldsymbol x \in \mathbb{R}^{|D|}} \|\boldsymbol x\|_1\quad \text{s.t.} \quad\tilde D \boldsymbol x = \tilde{\boldsymbol m}.
    \end{equation}
    The overall strategy is as follows: (i) show that the null space property of order $K$ (NSP$_K$) holds under the assumption $\mu < \frac{1}{2K-1}$; (ii) recall the equivalence NSP$_K$ $\iff$ exact and unique recovery of any $K$-sparse solution via noiseless $\ell_1$-minimization.

    Formally, the null space property of order $K$ (NSP$_K$) is defined as
    \begin{equation}
        \forall \boldsymbol h \in \ker(\tilde D) \setminus \{\boldsymbol 0\},\ \forall \mathcal S \subseteq [n],\ |\mathcal S| \le K:\quad\,\|\boldsymbol h_\mathcal S\|_1 < \|\boldsymbol h_{\mathcal S^c}\|_1\, ,
    \end{equation}
    where $\ker(\tilde D) = \{ \boldsymbol h : \tilde D \boldsymbol h = \boldsymbol 0 \}$, $\mathcal S \subseteq [n]$ is an index set with $[n]=\{1,\dots,n\}$, $\boldsymbol h_\mathcal S$ denotes the restriction of $\boldsymbol h$ to the coordinates in $\mathcal S$ (with other entries set to zero), and \(\mathcal S^c=[n]\setminus \mathcal S\) is a complementary set.

    \paragraph{Step (i): Proof of NSP\texorpdfstring{\(_K\)}.} Let $G = \tilde D^\top \tilde D$ denote the Gram matrix. Since each column of $\tilde D$ is normalized, we have $G_{jj} = 1$ and $|G_{ij}| \leq \mu$ for $i \neq j$. Take any $\boldsymbol h \in \ker(\tilde D)\setminus\{\boldsymbol 0\}$ and any index set $\mathcal S$ with $|\mathcal S| = K$. 
    
    Since $\tilde D \boldsymbol h = 0$, we have $G \boldsymbol h = 0$. For any $j$,
    \begin{equation}
        0 = (G \boldsymbol h)_j = \sum_i G_{ji} h_i= G_{jj} h_j + \sum_{i \neq j} G_{ji} h_i\quad \Rightarrow \quad h_j = -\sum_{i \neq j} G_{ji} h_i.
    \end{equation}
    
    Taking absolute values and using $|G_{ji}| \leq \mu$, we obtain
    \begin{equation}
        |h_j| \leq \mu \sum_{i \neq j} |h_i|.
    \end{equation}
    
    Summing over $j \in \mathcal S$ gives
    \begin{equation}
        \sum_{j \in \mathcal S} |h_j|\ \leq \ \mu \sum_{j \in \mathcal S} \sum_{i \neq j} |h_i|.
    \end{equation}
    
    The inner summation can be decomposed into contributions from $i \in \mathcal S \setminus \{j\}$ and $i \in \mathcal S^c$:
    \begin{equation}
        \sum_{j \in \mathcal S}\sum_{i \neq j}|h_i|= \underbrace{\sum_{j \in \mathcal S}\sum_{\substack{i \in \mathcal S \\ i \neq j}} |h_i|}_{\text{each }i \in \mathcal S \text{ counted }K-1\text{ times}}+ \underbrace{\sum_{j \in \mathcal S}\sum_{i \in \mathcal S^c} |h_i|}_{\text{each }i \in \mathcal S^c \text{ counted }K\text{ times}}.
    \end{equation}
    
    Hence,
    \begin{equation}
        \|\boldsymbol h_\mathcal S\|_1 \ \leq\ \mu \big( (K-1)\|\boldsymbol h_\mathcal S\|_1 + K \|\boldsymbol h_{\mathcal S^c}\|_1 \big).
    \end{equation}
    
    Rearranging,
    \begin{equation}
        \big(1 - (K-1)\mu \big)\|\boldsymbol h_\mathcal S\|_1 \ \leq\ K\mu\,\|\boldsymbol h_{\mathcal S^c}\|_1.
    \end{equation}
    
    Dividing through by the positive factor $1 - (K-1)\mu$, define
    \begin{equation}
        \alpha := \frac{K\mu}{1 - (K-1)\mu}.
    \end{equation}
    
    When $\mu < \tfrac{1}{2K-1}$, we have $\alpha < 1$. Since $\boldsymbol h \neq 0$ and $\tilde D \boldsymbol h = 0$, it is impossible for $\|\boldsymbol h_{\mathcal S^c}\|_1 = 0$ (otherwise both terms would vanish, forcing $\boldsymbol h=0$, a contradiction). Therefore,
    \begin{equation}
        \|\boldsymbol h_\mathcal S\|_1 \ \leq\ \alpha \|\boldsymbol h_{\mathcal S^c}\|_1 \ < \ \|\boldsymbol h_{\mathcal S^c}\|_1.
    \end{equation}
    
    Take any $\mathcal S_0$ with $|\mathcal S_0| = k \leq K$, and extend it to a superset $\mathcal S \supseteq \mathcal S_0$ such that $|\mathcal S| = K$. Then,
    \begin{equation}
        \|\boldsymbol h_{\mathcal S_0}\|_1 \ \leq\ \|\boldsymbol h_{\mathcal S}\|_1, \qquad \|\boldsymbol h_{\mathcal S_0^c}\|_1 \ \geq\ \|\boldsymbol h_{\mathcal S^c}\|_1.
    \end{equation}
    
    If we know that $\|\boldsymbol h_\mathcal S\|_1 < \|\boldsymbol h_{\mathcal S^c}\|_1$ holds for all $\mathcal S$ of size $K$, then it follows that
    \begin{equation}
        \|\boldsymbol h_{\mathcal S_0}\|_1 \ \leq\ \|\boldsymbol h_{\mathcal S}\|_1 \ <\ \|\boldsymbol h_{\mathcal S^c}\|_1 \ \leq\ \|\boldsymbol h_{\mathcal S_0^c}\|_1.
    \end{equation}
    
    Thus, the inequality also holds for any $\mathcal S_0$ with $|\mathcal S_0| \leq K$, which establishes NSP$_K$.

    \paragraph{Step (ii): Equivalence between NSP\texorpdfstring{$_K$}\ and \texorpdfstring{$\ell_1$} recovery.}

    \paragraph{NSP\texorpdfstring{\(_K\)} \texorpdfstring{\(\Rightarrow\)}\ unique \texorpdfstring{\(\ell_1\)}\ recovery:} Suppose $\hat{\boldsymbol x}$ is another feasible solution such that $\tilde D \hat{\boldsymbol x} = \tilde D \boldsymbol \delta$. Let $\boldsymbol h = \hat{\boldsymbol x} - \boldsymbol \delta \in \ker(\tilde D) \setminus \{\boldsymbol 0\}$, and let $\mathcal S = \mathrm{supp}(\boldsymbol \delta)$ with $|\mathcal S|\le K$. Then
    \begin{equation}
        \|\hat{\boldsymbol x}\|_1 = \|\boldsymbol \delta + \boldsymbol h\|_1= \|\boldsymbol \delta_\mathcal S + \boldsymbol h_\mathcal S\|_1 + \|\boldsymbol h_{\mathcal S^c}\|_1\ \geq\ \|\boldsymbol \delta_\mathcal S\|_1 - \|\boldsymbol h_\mathcal S\|_1 + \|\boldsymbol h_{\mathcal S^c}\|_1\ >\ \|\boldsymbol \delta_\mathcal S\|_1= \|\boldsymbol \delta\|_1,
    \end{equation}
    where the strict inequality follows from NSP$_K$. Hence, $\boldsymbol \delta$ is the unique minimizer of the $\ell_1$ problem.

    \paragraph{Unique \texorpdfstring{$\ell_1$}\ recovery \texorpdfstring{$\Rightarrow$}\ NSP\texorpdfstring{$_K$}\ :} We argue by contradiction. If NSP$_K$ does not hold, then there exists $\boldsymbol h \in \ker(\tilde D)\setminus\{\boldsymbol 0\}$ and some $\mathcal S$ with $|\mathcal S|\le K$ such that $\|\boldsymbol h_\mathcal S\|_1 \geq \|\boldsymbol h_{\mathcal S^c}\|_1$. Take any nonzero $K$-sparse $\boldsymbol \delta$ with $\mathrm{supp}(\boldsymbol \delta) = \mathcal S$, and choose $\delta_j = \alpha_j \operatorname{sgn}(h_j)$ with $\alpha_j \geq |h_j|$ coordinate-wise. Consider $\hat{\boldsymbol x} = \boldsymbol \delta - \boldsymbol h$. Since $\tilde D \boldsymbol h = \boldsymbol 0$, both $\boldsymbol \delta$ and $\hat{\boldsymbol x}$ are feasible, and
    \begin{equation}
        \|\hat{\boldsymbol x}\|_1 = \|\boldsymbol \delta_\mathcal S - \boldsymbol h_\mathcal S\|_1 + \|\boldsymbol h_{\mathcal S^c}\|_1= \|\boldsymbol \delta_\mathcal S\|_1 - \|\boldsymbol h_\mathcal S\|_1 + \|\boldsymbol h_{\mathcal S^c}\|_1\ \leq \ \|\boldsymbol \delta_\mathcal S\|_1= \|\boldsymbol \delta\|_1.
    \end{equation}
    
    Thus, $\boldsymbol \delta$ is not the unique minimizer of the $\ell_1$ problem (and may even fail to be a minimizer). This contradicts the uniqueness assumption. Therefore, NSP$_K$ must hold.
\end{proof}

\subsection{Proof of Corollary~\ref{corollary:Monorepresentationality}}
\Monorepresentationality*
\begin{proof}
    Take arbitrary \(\boldsymbol{x}, \boldsymbol{y} \in \Sigma_K\) such that \(\tilde{D}\boldsymbol{x} = \tilde{D}\boldsymbol{y}\). Let \(\tilde{\boldsymbol{m}} := \tilde{D}\boldsymbol{x}\), then clearly \(\tilde{\boldsymbol{m}} = \tilde{D}\boldsymbol{y}\), with \(\|\boldsymbol{x}\|_0 \le K\) and \(\|\boldsymbol{y}\|_0 \le K\).
    
    By Theorem~\ref{theorem:Uniqueness_of_Sparse_Representation_and_Exact_Recovery}, under the condition \(\mu < \tfrac{1}{2K-1}\), the \(K\)-sparse representation \(\boldsymbol{\delta}\) determined by the equation \(\tilde{\boldsymbol{m}} = \tilde{D}\boldsymbol{\delta}\) is unique: among all coefficient vectors satisfying \(\|\boldsymbol{\delta}\|_0 \le K\), there exists only one that generates \(\tilde{\boldsymbol{m}}\).
    
    Since both \(\boldsymbol{x}\) and \(\boldsymbol{y}\) are feasible solutions satisfying \(\|\boldsymbol{\delta}\|_0 \le K\) and \(\tilde{D}\boldsymbol{\delta} = \tilde{\boldsymbol{m}}\), uniqueness forces \(\boldsymbol{x} = \boldsymbol{y}\). Hence, \(\Phi\) is injective on \(\Sigma_K\).
\end{proof}

\subsection{Proof of Theorem~\ref{theorem:Identifiability_of_SAEs}}
\IdentifiabilitySAEs*
\begin{proof}
    Let \(\boldsymbol G := \boldsymbol D^{\top} \widetilde{\boldsymbol S} \boldsymbol D\). By the definition of the normalized atomic inner product, \(G_{kk}=1\), and for every \(k\neq j\), \(|G_{kj}| = \left| \langle \boldsymbol d_k,\boldsymbol d_j \rangle_{\widetilde{\boldsymbol S}} \right| \le \epsilon\). Since \(\boldsymbol m_i = \boldsymbol D\boldsymbol\delta_i\), we have
    \begin{equation}
        \boldsymbol W_{\mathrm{enc}}\boldsymbol m_i = \boldsymbol D^{\top} \widetilde{\boldsymbol S} \boldsymbol D\boldsymbol\delta_i = \boldsymbol G\boldsymbol\delta_i.
    \end{equation}
    
    Consider the \(k\)-th coordinate.
    
    \paragraph{For \texorpdfstring{\(k\notin\mathcal S_i\)}\ ,}
    \begin{equation}
        (\boldsymbol G\boldsymbol\delta_i)_k = \sum_{j\in\mathcal S_i^{+}} \delta_{ij}G_{kj} + \sum_{j\in\mathcal S_i^{-}} \delta_{ij}G_{kj}.
    \end{equation}
    By the coherence bound,
    \begin{equation}
        \left| (\boldsymbol G\boldsymbol\delta_i)_k \right| \le \epsilon\left( |\mathcal S_i^{+}|\delta_{\max}^{+} + |\mathcal S_i^{-}|\delta_{\max}^{-} \right) = B_0.
    \end{equation}
    Hence
    \begin{equation}
        -B_0 \le (\boldsymbol G\boldsymbol\delta_i)_k \le B_0.
    \end{equation}
    Because \(\tau_-\le -B_0\) and \(\tau_+\ge B_0\), it follows that
    \begin{equation}
        \tau_- \le (\boldsymbol G\boldsymbol\delta_i)_k \le \tau_+,
    \end{equation}
    and this coordinate is therefore set to zero by the bidirectional threshold activation.

    \paragraph{For \texorpdfstring{\(k\in\mathcal S_i^{+}\)}\ ,}
    \begin{equation}
        (\boldsymbol G\boldsymbol\delta_i)_k = \delta_{ik} + \sum_{j\in\mathcal S_i\setminus\{k\}} \delta_{ij}G_{kj}.
    \end{equation}
    Define
    \begin{equation}
        e_{ik}^{+} := \sum_{j\in\mathcal S_i\setminus\{k\}} \delta_{ij}G_{kj}.
    \end{equation}
    Again using the coherence bound,
    \begin{equation}
        |e_{ik}^{+}| \le \epsilon\left[ (|\mathcal S_i^{+}|-1)\delta_{\max}^{+} + |\mathcal S_i^{-}|\delta_{\max}^{-} \right] = B_{+}.
    \end{equation}
    Therefore,
    \begin{equation}
        (\boldsymbol G\boldsymbol\delta_i)_k \ge \delta_{\min}^{+}-B_{+}.
    \end{equation}
    If \(\tau_{+} < \delta_{\min}^{+}-B_{+}\), then
    \begin{equation}
        (\boldsymbol G\boldsymbol\delta_i)_k > \tau_{+}.
    \end{equation}
    Thus, the positive active coordinate is retained and its sign remains positive.

    \paragraph{For \texorpdfstring{\(k\in\mathcal S_i^{-}\)}\ ,}
    \begin{equation}
        (\boldsymbol G\boldsymbol\delta_i)_k = \delta_{ik} + \sum_{j\in\mathcal S_i\setminus\{k\}} \delta_{ij}G_{kj} = \delta_{ik} + e_{ik}^{-}.
    \end{equation}
    The coherence bound gives
    \begin{equation}
        |e_{ik}^{-}| \le \epsilon\left[ |\mathcal S_i^{+}|\delta_{\max}^{+} + (|\mathcal S_i^{-}|-1)\delta_{\max}^{-} \right] = B_{-}.
    \end{equation}
    Since \(\delta_{ik} \le -\delta_{\min}^{-}\), we obtain
    \begin{equation}
        (\boldsymbol G\boldsymbol\delta_i)_k \le -\delta_{\min}^{-}+B_{-}.
    \end{equation}
    If \(-\delta_{\min}^{-}+B_{-} < \tau_-\), then
    \begin{equation}
        (\boldsymbol G\boldsymbol\delta_i)_k < \tau_-.
    \end{equation}
    Hence, the negative active coordinate is retained and its sign remains negative.

    Combining the three cases, all coordinates outside the support lie within \([\tau_-,\tau_+]\) and are set to zero, whereas positive and negative support coordinates lie above \(\tau_+\) and below \(\tau_-\), respectively. Therefore,
    \begin{equation}
        \operatorname{supp} \left( \sigma_{\tau_-,\tau_+} (\boldsymbol G\boldsymbol\delta_i) \right) = \mathcal S_i,
    \end{equation}
    and the signs of the retained activations agree with those of the true coefficients.

    Finally, for the bidirectional thresholds to exist, non-zero gaps must separate the inactive coordinates from both the positive and negative active coordinates. For the positive threshold, this requires
    \begin{equation}
        B_0 < \delta_{\min}^{+}-B_{+},
    \end{equation}
    or equivalently,
    \begin{equation}
        \delta_{\min}^{+} > B_0+B_{+} = \epsilon\left[ (2|\mathcal S_i^{+}|-1)\delta_{\max}^{+} + 2|\mathcal S_i^{-}|\delta_{\max}^{-} \right].
    \end{equation}
    For the negative threshold, we require
    \begin{equation}
        -\delta_{\min}^{-}+B_{-} < -B_0,
    \end{equation}
    which is equivalent to
    \begin{equation}
        \delta_{\min}^{-} > B_0+B_{-} = \epsilon\left[ 2|\mathcal S_i^{+}|\delta_{\max}^{+} + (2|\mathcal S_i^{-}|-1)\delta_{\max}^{-} \right].
    \end{equation}
    Here, \(B_0\) bounds the maximum interference on inactive coordinates, whereas \(B_{+}\) and \(B_{-}\) bound the maximum cross-interference on positive and negative active coordinates, respectively. Therefore, the conditions above ensure that the minimum active magnitudes remain sufficiently separated from the interference induced by the remaining atoms. Thus, the conclusion holds.
\end{proof}

\section{Empirical validation of the atomic inner product}\label{Section:Empirical validation of the atomic inner product}

To further assess whether the AIP provides a more appropriate characterization of neural representational geometry than the Euclidean inner product, we compare the distributions of pairwise angles under the two metrics. Specifically, we randomly sample 128 entities from Wikidata \cite{vrandevcic2014wikidata} and extract their internal representations across multiple layers of several language models. For each entity, the activation at its final token position is used as the corresponding representation \cite{meng2022locating}, and pairwise angles between entity representations are computed independently at each layer.

\begin{figure}[htbp]
    \centering
    \includegraphics[width=\linewidth]{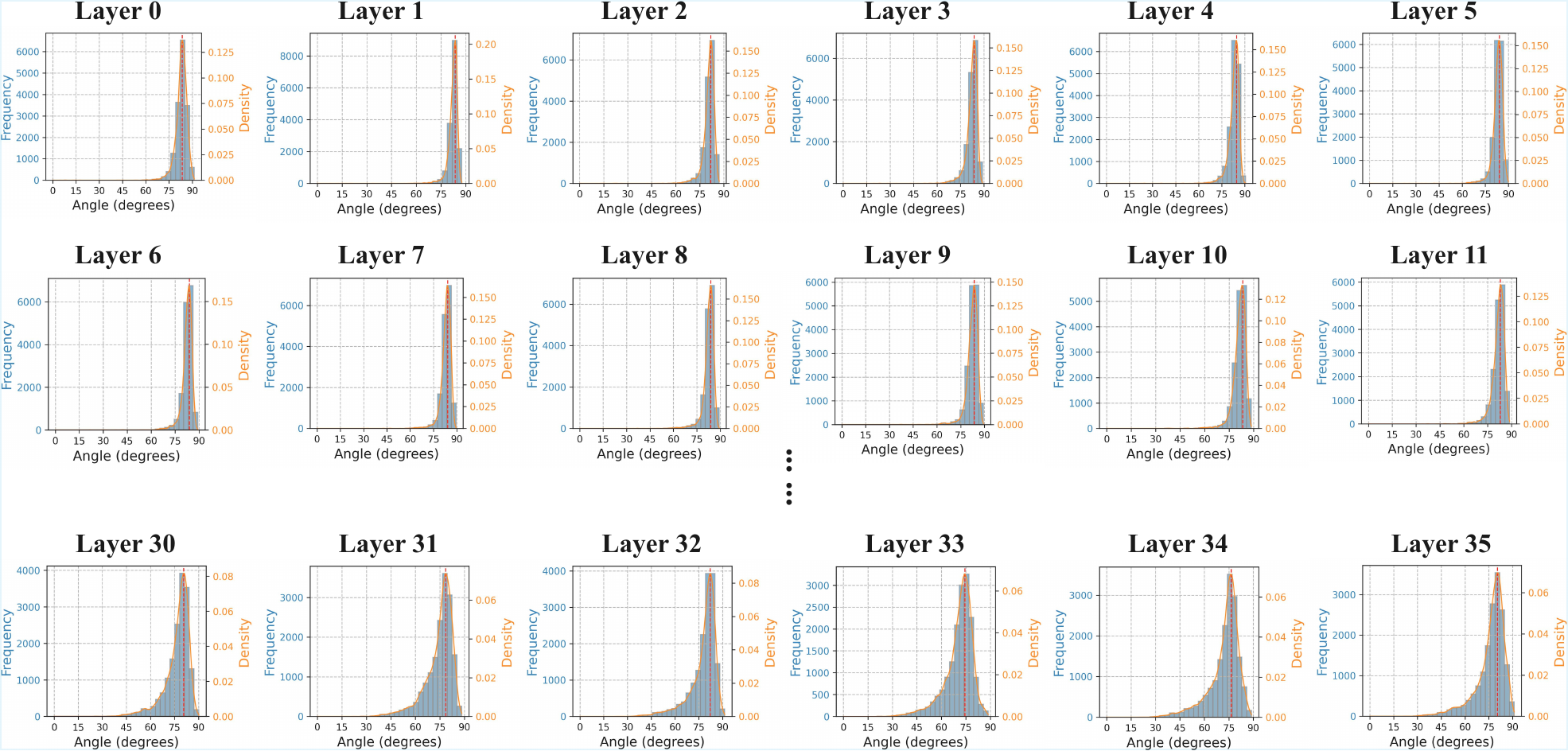}
    \caption{\textbf{Representation shift in GPT2-Large under the Euclidean inner product.} Angular distributions between representations of 128 randomly sampled Wikidata entities across model layers, extracted at the final token position of each entity. Across layers, the distributions are systematically shifted away from \(90^\circ\), indicating a persistent global directional bias in the Euclidean representation geometry.}
    \label{fig:Supp_Representation_Shift_GPT2_Large}
\end{figure}

\begin{figure}[htbp]
    \centering
    \includegraphics[width=\linewidth]{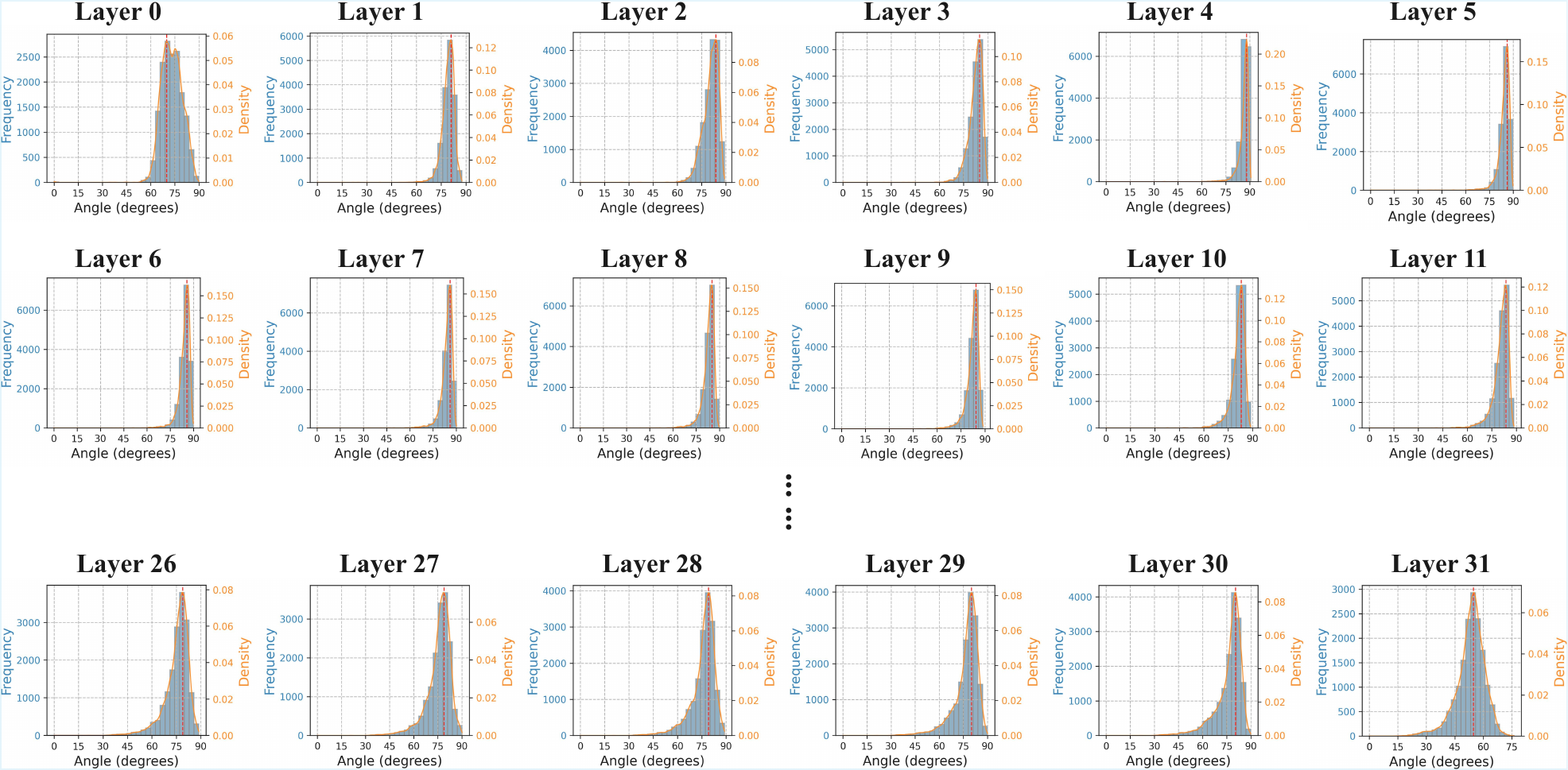}
    \caption{\textbf{Representation shift in Pythia-6.9B under the Euclidean inner product.} Angular distributions between representations of 128 randomly sampled Wikidata entities across model layers, extracted at the final token position of each entity. Across layers, the distributions are systematically shifted away from \(90^\circ\), indicating a persistent global directional bias in the Euclidean representation geometry.}
    \label{fig:Supp_Representation_Shift_Pythia-6.9B}
\end{figure}

\begin{figure}[htbp]
    \centering
    \includegraphics[width=\linewidth]{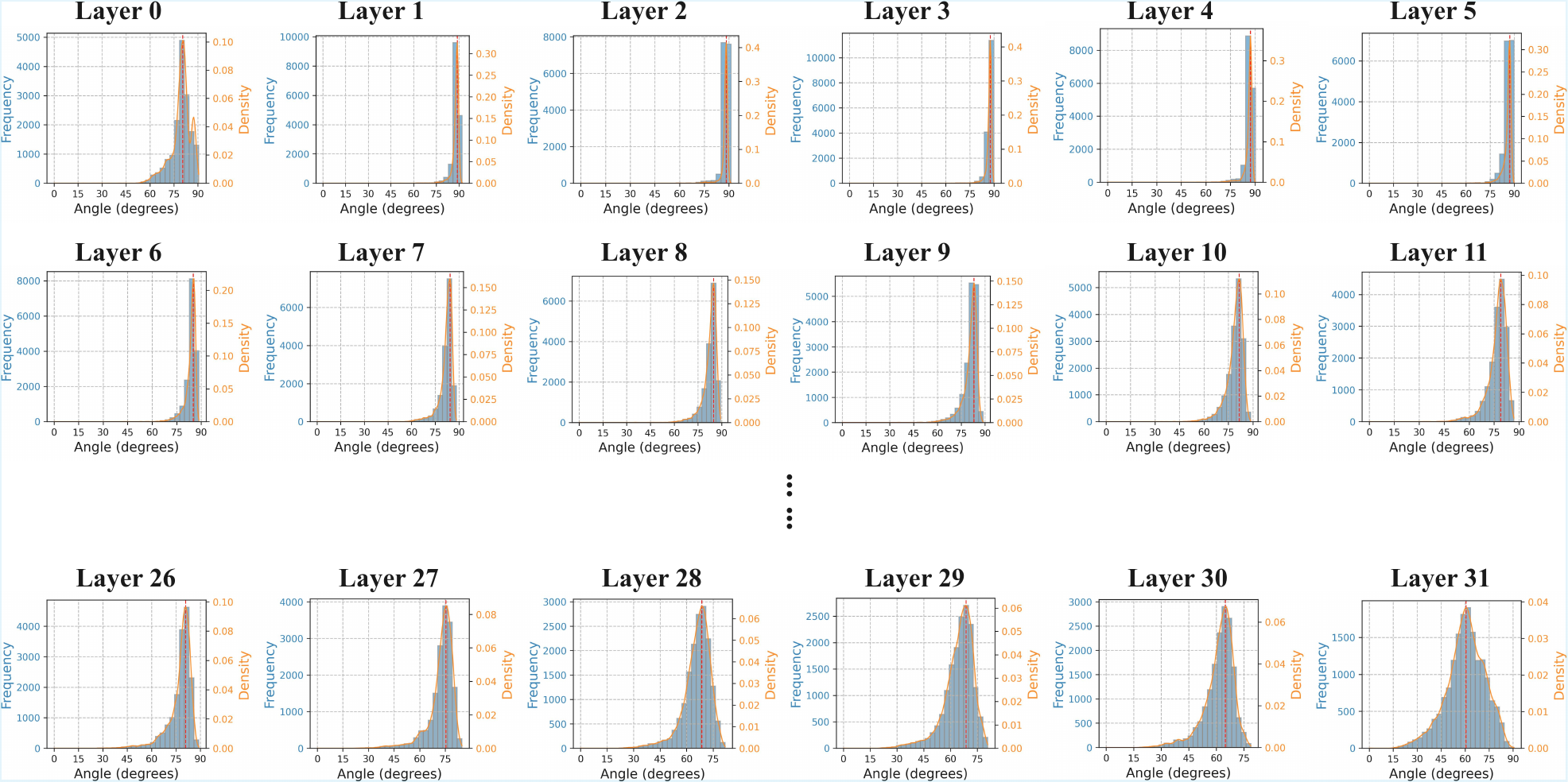}
    \caption{\textbf{Representation shift in Llama3.1-8B under the Euclidean inner product.} Angular distributions between representations of 128 randomly sampled Wikidata entities across model layers, extracted at the final token position of each entity. Across layers, the distributions are systematically shifted away from \(90^\circ\), indicating a persistent global directional bias in the Euclidean representation geometry.}
    \label{fig:Supp_Representation_Shift_Llama3.1-8B}
\end{figure}

\begin{figure}[htbp]
    \centering
    \includegraphics[width=\linewidth]{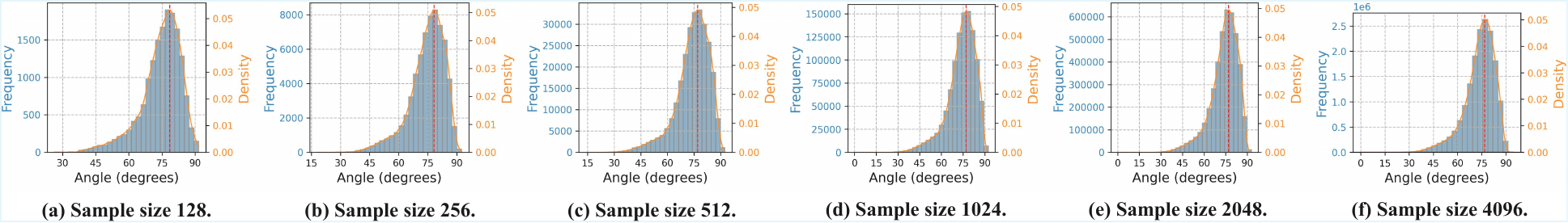}
    \caption{\textbf{Ablation of sample size for representation shift in GPT2-Large.}
    Angular distributions under the Euclidean inner product estimated using different numbers of randomly sampled Wikidata entities. The distribution centroid remains largely unchanged as the sample size increases, indicating that the observed representation shift is not attributable to finite-sample estimation.}
    \label{fig:Supp_Representation_Shift_Ablation_GPT2_Large}
\end{figure}

We first consider the standard Euclidean inner product. For two representations \(\boldsymbol u,\boldsymbol v\in\mathbb R^H\), the Euclidean cosine similarity is defined as
\begin{equation}
    \cos(\boldsymbol u,\boldsymbol v) = \frac{\boldsymbol u^\top\boldsymbol v} {\|\boldsymbol u\|_2\|\boldsymbol v\|_2},
\end{equation}
which we convert to angles for analysis. Across models and layers, the resulting angular distributions consistently deviate from \(90^\circ\) (Supplementary Fig.~\ref{fig:Supp_Representation_Shift_GPT2_Large}-\ref{fig:Supp_Representation_Shift_Llama3.1-8B}). We refer to this phenomenon as representation shift. This shift indicates that representations measured in Euclidean geometry are systematically influenced by a shared dominant direction, such that even semantically unrelated representations can exhibit elevated cosine similarity. Varying the number of samples used to estimate the angular distribution leaves its centroid largely unchanged (Supplementary Fig.~\ref{fig:Supp_Representation_Shift_Ablation_GPT2_Large}), indicating that this effect is not a finite-sampling artifact but instead reflects a persistent global directional bias under the Euclidean metric.

We next recompute the geometry of the same representations using the AIP. For each model and layer, we independently collect \(100\text{K}\) internal representations \(\boldsymbol k\) from Wikipedia and estimate the representation-induced metric from their second-order statistics as \(\widetilde{\boldsymbol S} = \left( \mathbb E[\boldsymbol k\boldsymbol k^\top] \right)^{-1}\). The normalized cosine similarity induced by this metric is then given by
\begin{equation}
    \cos_{\widetilde S}(\boldsymbol u,\boldsymbol v) = \frac{ \boldsymbol u^\top\widetilde{\boldsymbol S}\boldsymbol v }{ \sqrt{\boldsymbol u^\top\widetilde{\boldsymbol S}\boldsymbol u} \sqrt{\boldsymbol v^\top\widetilde{\boldsymbol S}\boldsymbol v} }.
\end{equation}
As above, we convert these similarities to angles and examine their empirical distributions.

\begin{figure}[htbp]
    \centering
    \includegraphics[width=\linewidth]{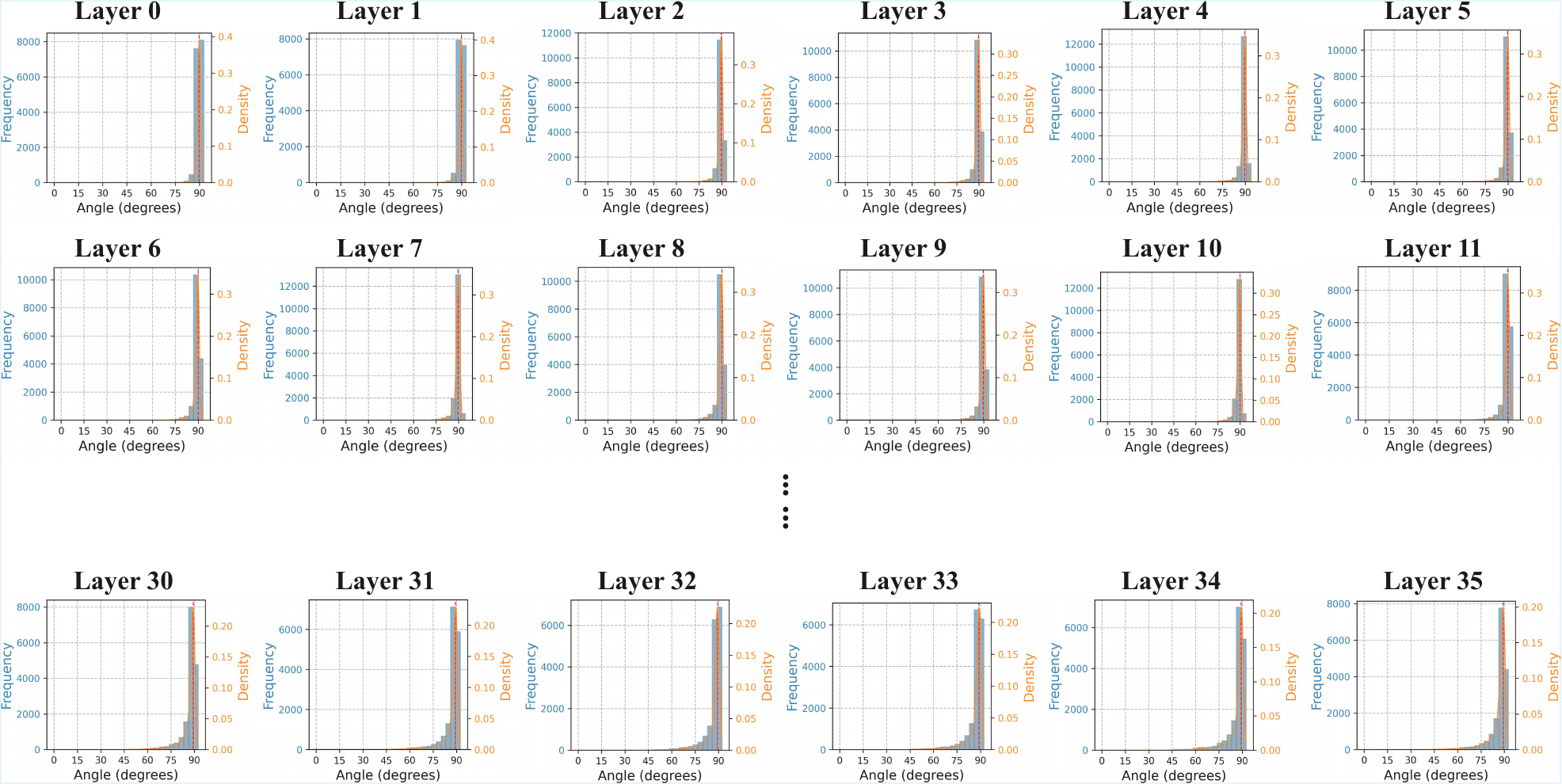}
    \caption{\textbf{Representational geometry of GPT2-Large under the atomic inner product.} Angular distributions between representations of 128 randomly sampled Wikidata entities across model layers, computed using the atomic inner product. Across layers, the distributions are centered close to \(90^\circ\), indicating that the AIP substantially reduces the global directional bias observed under the Euclidean inner product.}
    \label{fig:Supp_AIP_Geometry_GPT2_Large}
\end{figure}

\begin{figure}[htbp]
    \centering
    \includegraphics[width=\linewidth]{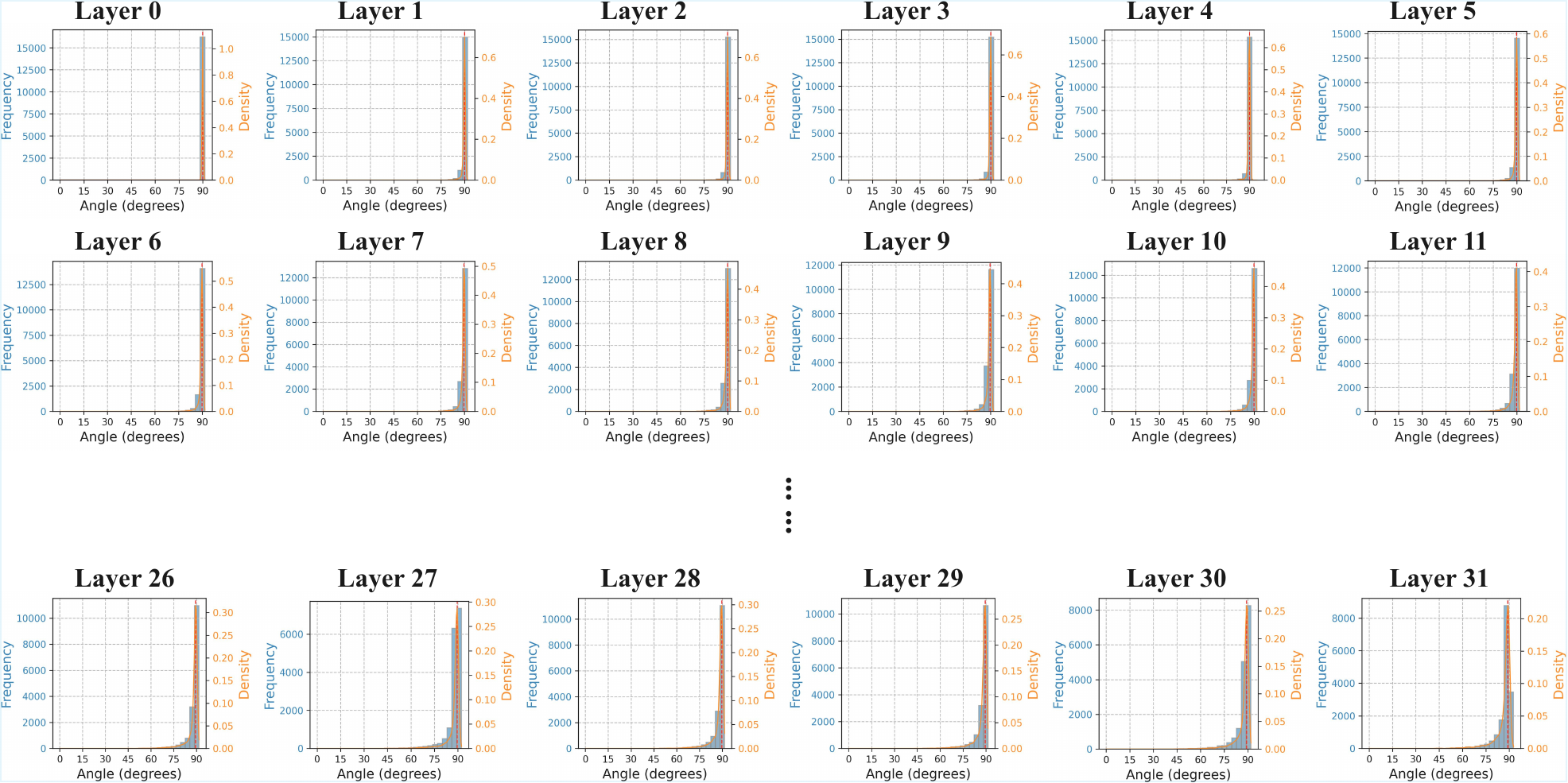}
    \caption{\textbf{Representational geometry of Pythia-6.9B under the atomic inner product.} Angular distributions between representations of 128 randomly sampled Wikidata entities across model layers, computed using the atomic inner product. Across layers, the distributions are centered close to \(90^\circ\), indicating that the AIP substantially reduces the global directional bias observed under the Euclidean inner product.}
    \label{fig:Supp_AIP_Geometry_Pythia-6.9B}
\end{figure}

\begin{figure}[htbp]
    \centering
    \includegraphics[width=\linewidth]{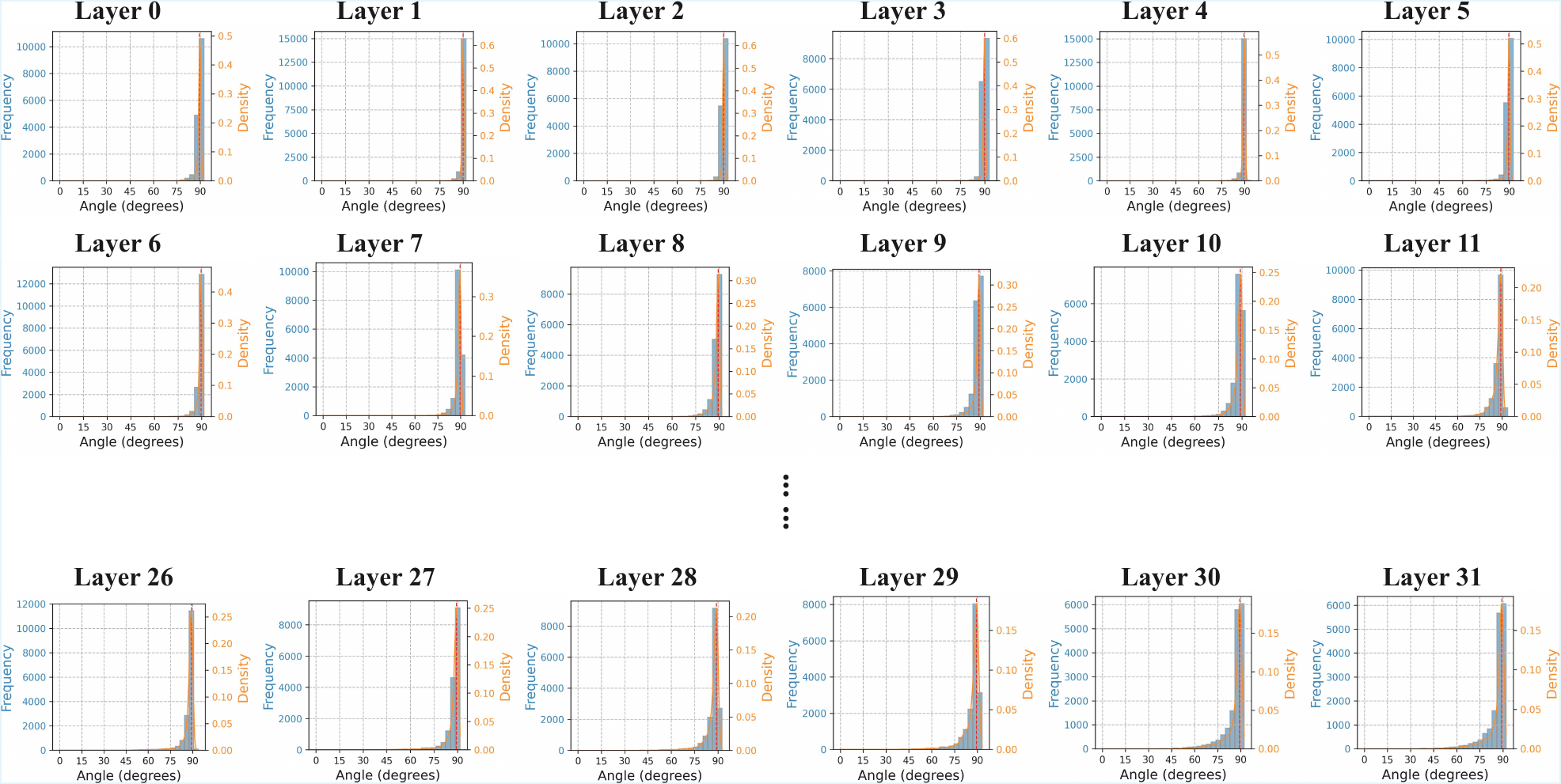}
    \caption{\textbf{Representational geometry of Llama3.1-8B under the atomic inner product.} Angular distributions between representations of 128 randomly sampled Wikidata entities across model layers, computed using the atomic inner product. Across layers, the distributions are centered close to \(90^\circ\), indicating that the AIP substantially reduces the global directional bias observed under the Euclidean inner product.}
    \label{fig:Supp_AIP_Geometry_Llama3.1-8B}
\end{figure}

Compared with the Euclidean inner product, the AIP substantially reduces the systematic angular shift, with the distributions across models and layers becoming centered close to \(90^\circ\) (Supplementary Fig.~\ref{fig:Supp_AIP_Geometry_GPT2_Large}-\ref{fig:Supp_AIP_Geometry_Llama3.1-8B}). At the same time, the high-similarity tail remains dominated by semantically related representation pairs, indicating that the AIP reduces the global directional bias while preserving local geometric structure associated with specific relationships between samples. This pattern is consistent across model families and network layers.

These observations are particularly relevant to the distinguishability criterion used in our framework. Stability depends on the coherence between candidate representational units, which is determined by their relative geometry. Under the Euclidean metric affected by a shared directional bias, units without strong structural correspondence may nevertheless exhibit elevated similarity, thereby obscuring the coherence arising from the underlying representational structure. By contrast, the AIP corrects this global bias and captures the underlying geometry relevant to representational distinguishability. These results therefore provide empirical support for using the AIP, rather than the standard Euclidean inner product, in our theoretical analysis and stability evaluation.

\section{Monosemanticity evaluation}\label{Section:Monosemanticity evaluation}

This section provides further details on the implementation of the monosemanticity evaluation described in Methods, including the criteria used to determine semantic coherence, the LLM-as-a-judge prompts and manual verification procedure for language models, and the manual evaluation criteria for diffusion models. All evaluations follow the unified monosemanticity criterion defined in the main text: for the activation instances of a representational unit, we identify the largest subset \(S_{\max}\) that can be consistently explained by a common semantic interpretation, and compute its monosemanticity score as \(|S_{\max}|/|S|\).

\bmhead{Criteria for semantic coherence}

For a given representational unit, we identify the largest subset of instances that can be explained by a shared concept, attribute or semantic pattern with a clear and specific meaning. The shared interpretation must be sufficiently semantically specific and cannot rely solely on an overly broad superordinate category. For example, the fact that multiple instances are all ``entities", ``words" or ``visual patterns" is not, by itself, sufficient to constitute a valid shared interpretation. Likewise, weak topical relatedness, indirect association or incidental co-occurrence is insufficient to group instances into the same subset.

Instances may differ substantially in their surface form, provided that they share the same underlying semantic concept, attribute or function in the relevant context. For a candidate interpretation, only instances that are directly and consistently accounted for by that interpretation are included in \(S_{\max}\); an instance is not included merely because it bears a weak relation to the other instances. If multiple plausible semantic interpretations exist, we select the interpretation that consistently accounts for the largest number of instances, and define the corresponding set of instances as \(S_{\max}\).

\bmhead{LLM-as-a-judge evaluation for language models}

For language models, we evaluate Wikipedia and Wikidata representations separately. For Wikipedia, each evaluation instance corresponds to a target token that activates the representational unit. To avoid basing the evaluation solely on the lexical form of the target token, we provide the evaluation model with the local context surrounding the token and explicitly highlight the target token. The evaluation model then determines whether different instances share a coherent semantic pattern based on the meaning or semantic function of the target token in its specific context. For Wikidata, each instance corresponds to the name of an entity that activates the representational unit, and the evaluation model identifies the largest coherent subset based on the entities' identities, types, attributes or semantic roles.

For each representational unit, the evaluation model is provided with all instances included in the evaluation and is asked to determine the largest subset that can be consistently explained by a common semantic concept, attribute or pattern.

For Wikipedia representations, we use the following prompt:

\begin{promptbox}
Below are \texttt{\{num\_examples\}} activation examples from the same SAE feature.

\medskip
Focus on the highlighted token [\ldots] in each context.

\medskip
Determine the largest number of examples that can reasonably be grouped into ONE coherent semantic category. Use a reasonably specific category based on the highlighted token's meaning or semantic function.

\medskip
\texttt{\{examples\_text\}}

\medskip
Reply ONLY as M/\texttt{\{num\_examples\}}, where M is the size of the largest coherent semantic group. Do not explain.
\end{promptbox}

For Wikidata representations, we use the following prompt:

\begin{promptbox}
Below are \texttt{\{num\_examples\}} activation examples from the same SAE feature.

\medskip
Each example corresponds to one Wikidata entity name.

\medskip
Determine the largest number of examples that can reasonably be grouped into ONE coherent semantic category. Use a reasonably specific category based on the entities' identities, types, or semantic roles.

\medskip
\texttt{\{examples\_text\}}

\medskip
Reply ONLY as M/\texttt{\{num\_examples\}}, where M is the size of the largest coherent semantic group. Do not explain.
\end{promptbox}

\bmhead{Manual verification of LLM-judge outputs}

The LLM judge returns the size of the largest semantically coherent group of instances as \(M/|S|\), where \(M\) corresponds to \(|S_{\max}|\). This output format directly provides the subset cardinality required to compute the monosemanticity score, while minimizing the influence of additional generated content on the evaluation.

To mitigate potential inconsistencies in automated judgements of semantic boundaries, overly broad categories or weak semantic associations, we further manually verify the scores produced by the LLM judge. Manual verification follows the same criteria as the automated evaluation. Specifically, the evaluator examines the same set of activation instances and determines the largest number of instances that can be consistently explained by a shared concept, attribute or semantic pattern with reasonable semantic specificity. Groups based only on overly broad categories, indirect semantic associations or weak topical relatedness are not considered semantically coherent.

When the value of \(M\) returned by the LLM judge is consistent with these criteria, the original score is retained. When the automated judgement and manual assessment disagree, \(M\) is revised according to the size of the largest semantically coherent group identified during manual verification. The verified value of \(M\) is then used as \(|S_{\max}|\) to compute the final monosemanticity score for the representational unit.

\bmhead{Manual evaluation for diffusion models}

For Stable Diffusion v1.5 and FLUX.2-klein-9B, we apply the same largest-coherent-subset criterion as for language models. Because the evaluation instances correspond to local visual regions in LAION-COCO images, we perform the evaluation manually. For each representational unit, the evaluator jointly examines all local image instances included in the evaluation and identifies the largest subset that can be explained by a shared visual semantic concept, attribute or pattern. Valid shared interpretations may correspond to object categories, object parts, semantically meaningful visual attributes, or other local structures that retain a consistent semantic meaning across different images. Representative examples of this evaluation procedure for Stable Diffusion v1.5 and FLUX.2-klein-9B are shown in Supplementary Figs.~\ref{fig:SD15_monosemanticity_cases} and \ref{fig:FLUX_monosemanticity_cases}, respectively.

Instances that exhibit only low-level visual similarity without a shared semantic interpretation are not considered monosemantic matches. For example, if multiple local regions share similar colours, simple textures or other low-level statistical properties but cannot be assigned a consistent semantic interpretation, they are not grouped into the same subset on this basis alone. Conversely, a visual attribute can be treated as a shared pattern if it has a stable and interpretable semantic meaning across instances. Once \(S_{\max}\) is identified, the monosemanticity score is computed using the same definition as for language models.

% 补充示例图

\begin{figure}
    \centering
    \includegraphics[width=0.98\textwidth]{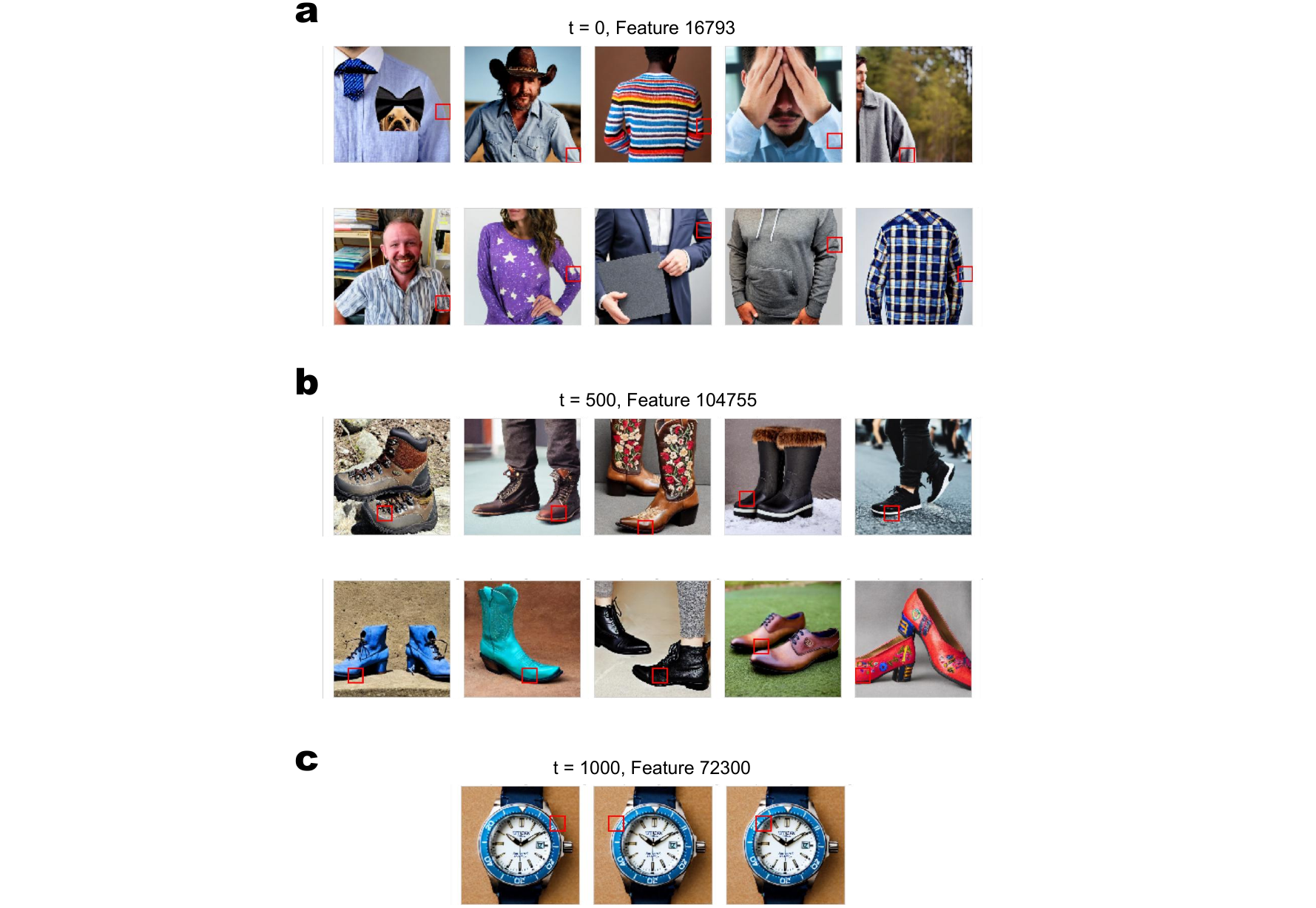}
    \caption{\textbf{Representative monosemanticity evaluation examples for BiJumpReLU features in Stable Diffusion v1.5.} \textbf{a--c}, Activation instances for representative BiJumpReLU features extracted from the conditional branch at diffusion timesteps \(t=0\) (\textbf{a}; Feature 16793), \(t=500\) (\textbf{b}; Feature 104755) and \(t=1000\) (\textbf{c}; Feature 72300). Red boxes indicate the local image regions corresponding to feature activations used in the monosemanticity evaluation. The selected features exhibit coherent visual-semantic responses across their activated instances, corresponding predominantly to arm in \textbf{a}, footwear in \textbf{b}, and wristwatch regions in \textbf{c}. Up to ten activation instances are shown for each feature; when fewer than ten instances are available, all available instances are included.}
    \label{fig:SD15_monosemanticity_cases}
\end{figure}

\begin{figure}
    \centering
    \includegraphics[width=0.98\textwidth]{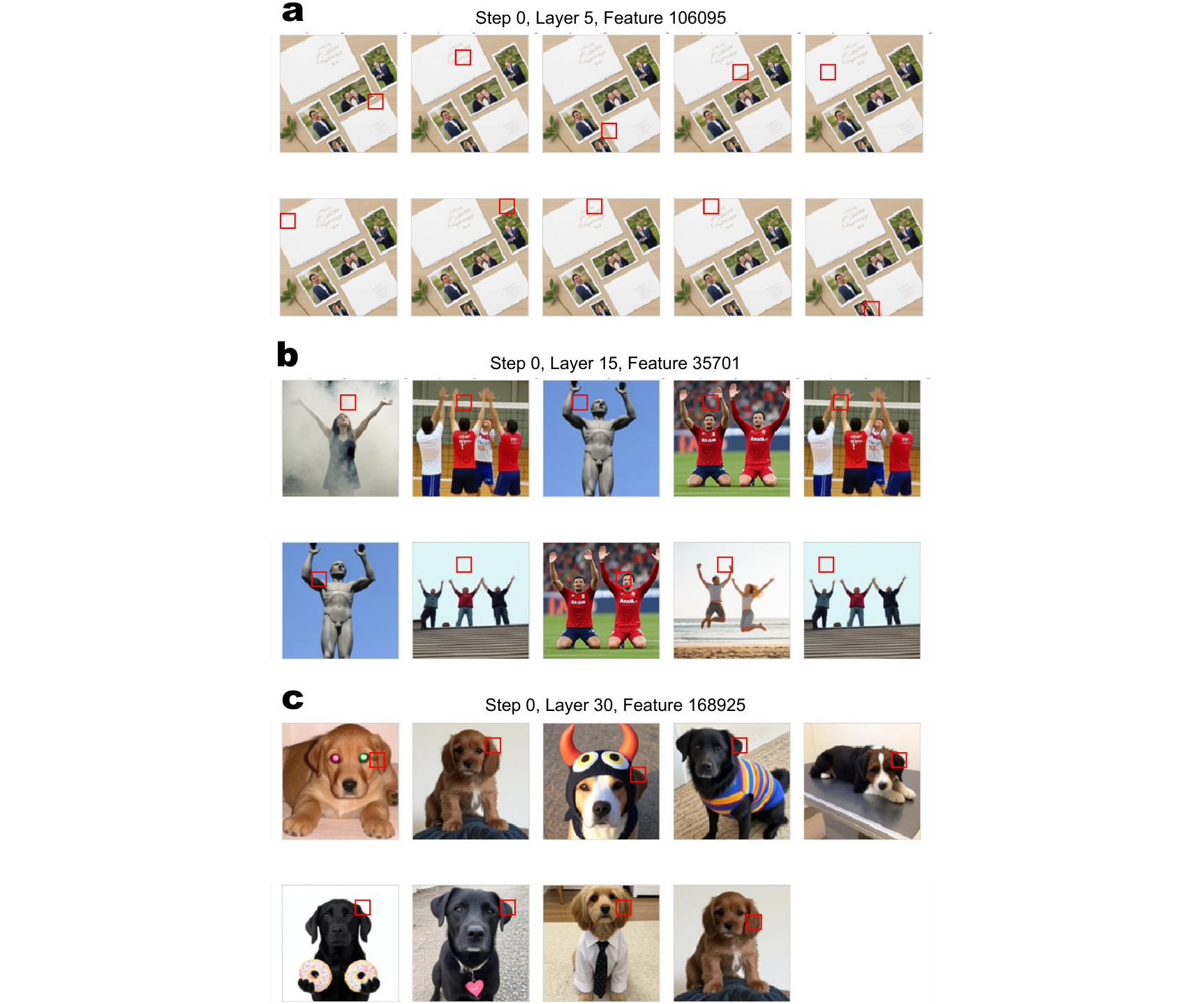}
    \caption{\textbf{Representative monosemanticity evaluation examples for BiJumpReLU features in FLUX.2-klein-9B.} \textbf{a--c}, Activation instances for representative BiJumpReLU features extracted at step 0 from Layer 5 (\textbf{a}; Feature 106095), Layer 15 (\textbf{b}; Feature 35701) and Layer 30 (\textbf{c}; Feature 168925). Red boxes indicate the local image regions corresponding to feature activations used in the monosemanticity evaluation. The selected features exhibit coherent visual-semantic responses across their activated instances, corresponding predominantly to multiple patches within the same semantically coherent region of a single image in \textbf{a}, raised arms or hands in \textbf{b}, and dog ear regions in \textbf{c}. Up to ten activation instances are shown for each feature; when fewer than ten instances are available, all available instances are included.}
    \label{fig:FLUX_monosemanticity_cases}
\end{figure}

\bmhead{Sampling and aggregation details}

Monosemanticity evaluation is performed independently for each model, data source and representation extraction site. For each setting, we randomly sample 10 candidate representational units learned by the corresponding SAE for evaluation. For each unit, we use at most 10 activation instances. If more than 10 activated instances are available, we randomly sample 10 from the full set of activated instances; if fewer than 10 are available, we use all available instances. Consequently, \(|S|\) may be smaller than 10 for some units. We then compute the monosemanticity score for each representational unit separately and average the scores of the sampled units within the same SAE setting to obtain the overall monosemanticity score for that setting.

\section{Semantic alignment evaluation}\label{Section:Semantic alignment evaluation}

\subsection{Single-atom semantic alignment evaluation}\label{Section:Single-atom semantic alignment evaluation}

To further detail the single-atom semantic alignment evaluation described in the main text, we represent both HowNet sememes and candidate representational units in the model as entity sets and quantify their correspondence based on set overlap. As described in the main text, we retain only HowNet sememes and candidate representational units that are associated with at least two unique entities.

For a given sememe \(s\), let \(E_s\) denote its corresponding entity set. For a candidate representational unit \(d_i^{(l)}\) at layer \(l\), let \(E_i^{(l)}\) denote the set of HowNet entities that activate that unit. We define the precision, recall and F1 score between the two sets as
\begin{equation}
    P\!\left(d_i^{(l)},s\right)=\frac{\left|E_i^{(l)} \cap E_s\right|}{\left|E_i^{(l)}\right|},
\end{equation}
\begin{equation}
    R\!\left(d_i^{(l)},s\right)=\frac{\left|E_i^{(l)} \cap E_s\right|}{\left|E_s\right|},
\end{equation}
and
\begin{equation}
    F_1\!\left(d_i^{(l)},s\right)=\frac{2P\!\left(d_i^{(l)},s\right)R\! \left(d_i^{(l)},s\right)}{P\!\left(d_i^{(l)},s\right)+R\!\left(d_i^{(l)},s\right)}.
\end{equation}
When the two entity sets do not overlap, the F1 score is defined as 0. Precision measures the proportion of entities activating the representational unit that belong to the target sememe, and therefore reflects the semantic purity of the unit with respect to that sememe. Recall measures the extent to which the unit covers the entities associated with the target sememe, whereas F1 jointly accounts for both quantities.

For each sememe \(s\), we search across all valid candidate representational units from all model layers and identify the unit with the highest F1 score:
\begin{equation}
    d_s^{*}=\arg\max_{l,i}F_1\!\left(d_i^{(l)},s\right).
\end{equation}
The corresponding \(F_1\!\left(d_s^{*},s\right)\) is used as the single-atom semantic alignment score for sememe \(s\). For each sememe, we additionally record the model layer containing the best-matching unit, together with its precision, recall and F1 score. We also retain the intersection between the two entity sets, the entities covered only by the candidate unit, and the entities belonging only to the target sememe, enabling further analysis of the semantic purity and coverage of the single-atom match. If multiple candidate units attain the same maximum F1 score, the first unit encountered in the candidate ordering is retained. This choice does not affect the resulting single-atom semantic alignment score.

\subsection{Cross-layer compositional semantic alignment evaluation}\label{Section:Cross-layer compositional semantic alignment evaluation}

To assess whether combinations of representational atoms can more completely recover the entity structure associated with HowNet sememes, we further perform a cross-layer compositional semantic alignment analysis. For each model, we aggregate candidate representational units that satisfy the filtering criteria described in the main text across all model layers, thereby forming a unified pool of candidate atoms.

For representational units exhibiting both positive and negative responses, we treat the entity sets associated with positive and negative activations as separate candidates, such that each candidate atom corresponds to a single, well-defined activation direction. For a target sememe \(s\), let \(E_s\) denote its entity set, and let \(E_i\) denote the entity set associated with candidate atom \(d_i\).

\bmhead{Candidate prefiltering}

Because exhaustively enumerating all possible combinations of cross-layer candidate atoms is computationally infeasible, we first prefilter the candidate atoms independently for each target sememe. Specifically, we use the same definition of precision as in the single-atom analysis:
\begin{equation}
    P(d_i,s)=\frac{|E_i \cap E_s|}{|E_i|}.
\end{equation}
We retain only candidate atoms satisfying \(P(d_i,s)\geq\tau_P\) and \(|E_i\cap E_s|>0\), where \(\tau_P\) denotes the precision threshold used for candidate filtering. The retained candidates are then ranked, in order, by precision, F1 score and the size of their entity-set intersection with the target sememe, and at most the top \(K_{\mathrm{cand}}\) candidates are kept. We set \(K_{\mathrm{cand}}=4000\) for all experiments. This prefiltering step removes candidates that lack a basic entity-level correspondence with the target sememe, thereby reducing the search space for the subsequent compositional search.

\bmhead{Greedy compositional search}

On the filtered candidate pool, we construct multi-atom combinations using a greedy set-union strategy. Let \(A_t\) denote the set of atoms selected after iteration \(t\). The entity set associated with the combination is defined as the union of the entity sets of the selected candidate atoms:
\begin{equation}
    E(A_t)=\bigcup_{d_i\in A_t} E_i.
\end{equation}
The precision, recall and F1 score between combination \(A_t\) and target sememe \(s\) are defined as
\begin{equation}
    P(A_t,s)=\frac{|E(A_t)\cap E_s|}{|E(A_t)|},
\end{equation}
\begin{equation}
    R(A_t,s)=\frac{|E(A_t)\cap E_s|}{|E_s|},
\end{equation}
and
\begin{equation}
    F_1(A_t,s)=\frac{2P(A_t,s)R(A_t,s)}{P(A_t,s)+R(A_t,s)}.
\end{equation}

The search starts from the empty combination, \(A_0=\varnothing\). At each iteration, we evaluate the F1 score obtained by adding each unselected candidate atom to the current combination and select the candidate that yields the largest F1 improvement:
\begin{equation}
    d_{t+1}^{*}=\arg\max_{d_i\notin A_t}\left[F_1\!\left(A_t\cup\{d_i\},s\right)-F_1(A_t,s)\right].
\end{equation}
We denote the corresponding maximum improvement as
\begin{equation}
    \Delta_{t+1}=F_1\!\left(A_t\cup\{d_{t+1}^{*}\},s\right)-F_1(A_t,s).
\end{equation}
The selected atom is added to the combination only when \(\Delta_{t+1}>0\):
\begin{equation}
    A_{t+1}=A_t\cup\{d_{t+1}^{*}\}.
\end{equation}
The search terminates when no remaining candidate atom can further improve the F1 score. We additionally impose a maximum combination size \(N_{\max}\), and terminate the search when \(|A_t|=N_{\max}\). Consequently, the final number of atoms selected for different sememes is not fixed in advance, but is determined adaptively by the F1 improvements subject to the maximum combination-size constraint.

\bmhead{Selection of search parameters}

The candidate-filtering threshold \(\tau_P\) and the maximum combination size \(N_{\max}\) jointly affect the candidate search space and the resulting compositional semantic alignment. We therefore systematically examine their effects over a predefined parameter grid.

Specifically, we first evaluate combinations of
\[
\tau_P\in\{0.1,\ 0.2,\ 0.3,\ 0.4,\ 0.5,\ 0.6,\ 0.7,\ 0.8,\ 0.9\}
\]
and
\[
N_{\max}\in\{20,\ 40,\ 60,\ 80,\ 100,\ 120,\ 140,\ 160,\ 180\}.
\]
For each parameter setting, overall compositional alignment performance is measured by the mean final F1 score across all valid HowNet sememes.

\begin{figure}
    \centering
    \includegraphics[width=0.98\textwidth]{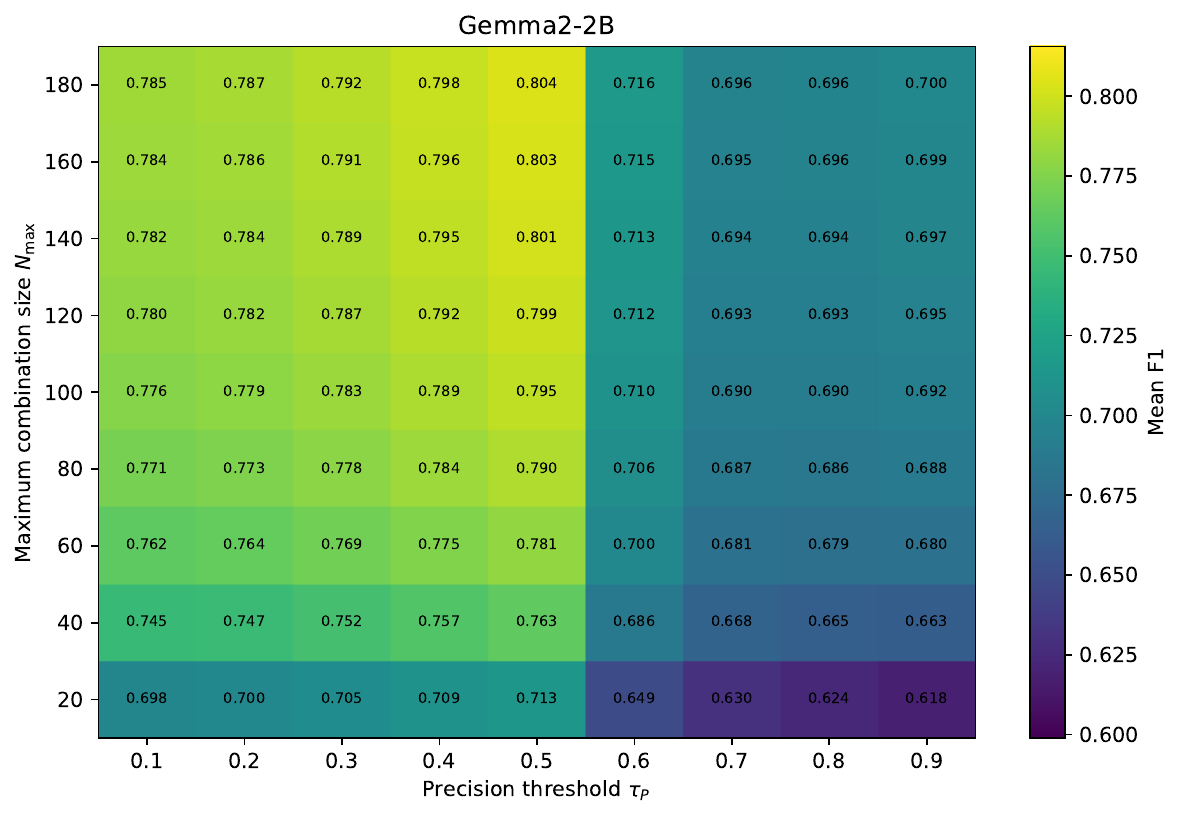}
    \caption{\textbf{Grid search for compositional semantic alignment hyperparameters in Gemma2-2B.} Mean final F1 score across valid HowNet sememes under different combinations of the candidate-filtering precision threshold \(\tau_P\) and maximum combination size \(N_{\max}\). Each cell reports the mean final F1 score obtained after compositional search under the corresponding parameter setting, with colour indicating the same quantity. Within the evaluated grid, \(\tau_P=0.5\) consistently yields the highest overall alignment performance, while increasing \(N_{\max}\) progressively improves the final F1 score.}
    \label{fig:Gemma2_2B_grid_search_mean_f1_heatmap}
\end{figure}

\begin{figure}
    \centering
    \includegraphics[width=0.98\textwidth]{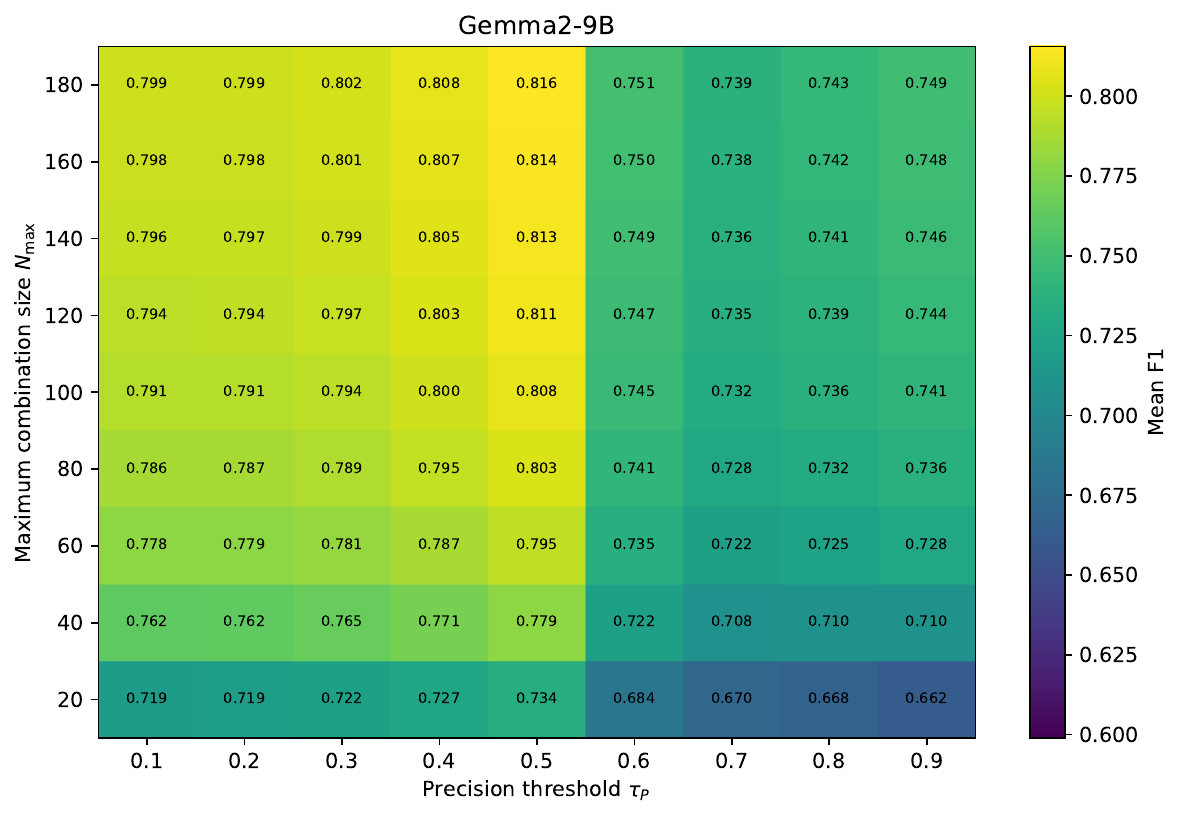}
    \caption{\textbf{Grid search for compositional semantic alignment hyperparameters in Gemma2-9B.} Mean final F1 score across valid HowNet sememes under different combinations of the candidate-filtering precision threshold \(\tau_P\) and maximum combination size \(N_{\max}\). Each cell reports the mean final F1 score obtained after greedy compositional search under the corresponding parameter setting, with colour indicating the same quantity. Within the evaluated grid, \(\tau_P=0.5\) consistently yields the highest overall alignment performance, while increasing \(N_{\max}\) progressively improves the final F1 score.}
    \label{fig:Gemma2_9B_grid_search_mean_f1_heatmap}
\end{figure}

\begin{figure}
    \centering
    \includegraphics[width=0.98\textwidth]{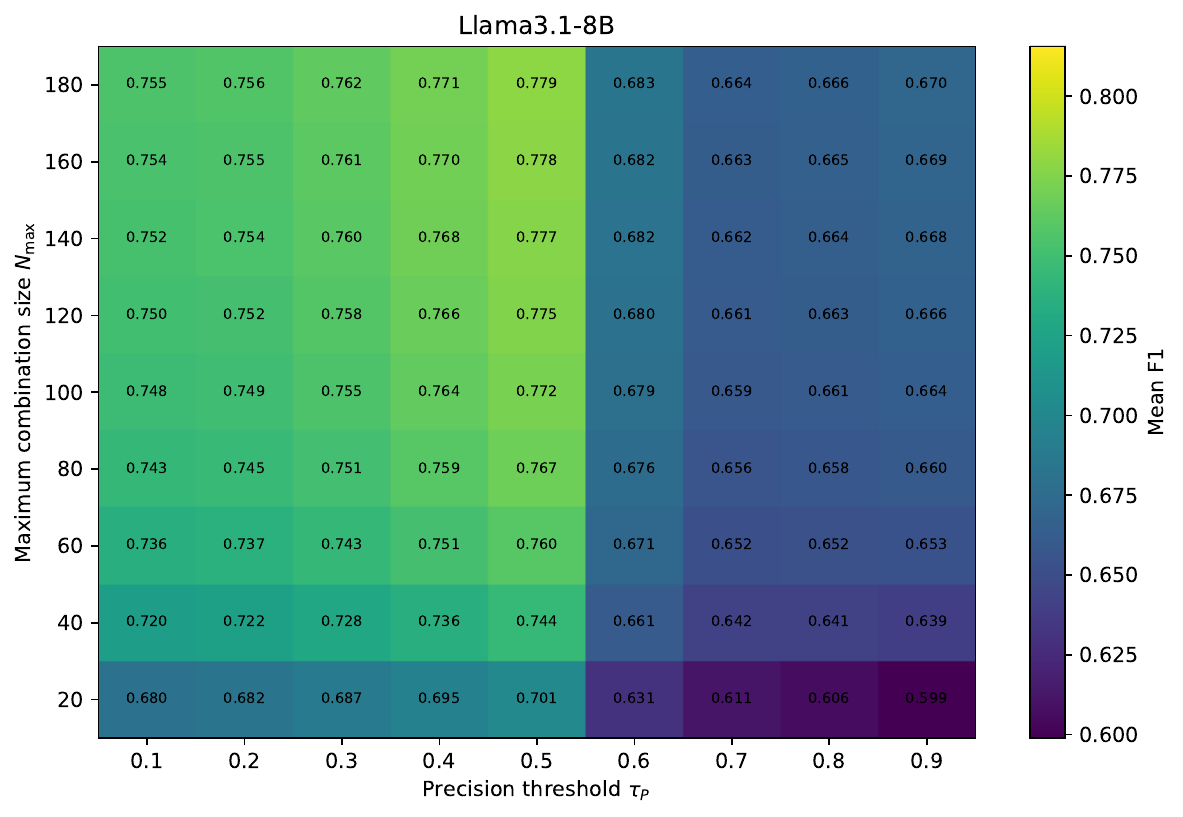}
    \caption{\textbf{Grid search for compositional semantic alignment hyperparameters in Llama3.1-8B.} Mean final F1 score across valid HowNet sememes under different combinations of the candidate-filtering precision threshold \(\tau_P\) and maximum combination size \(N_{\max}\). Each cell reports the mean final F1 score obtained after greedy compositional search under the corresponding parameter setting, with colour indicating the same quantity. Within the evaluated grid, \(\tau_P=0.5\) consistently yields the highest overall alignment performance, while increasing \(N_{\max}\) progressively improves the final F1 score.}
    \label{fig:Llama3_1_8B_grid_search_mean_f1_heatmap}
\end{figure}

Supplementary Fig.~\ref{fig:Gemma2_2B_grid_search_mean_f1_heatmap}--\ref{fig:Llama3_1_8B_grid_search_mean_f1_heatmap} shows the results under different parameter settings. The initial grid search indicates that \(\tau_P=0.5\) yields the best overall semantic alignment performance, while increasing \(N_{\max}\) continues to improve compositional alignment. Based on this result, we further fix \(\tau_P=0.5\) and additionally evaluate
\[
N_{\max}\in\{200,\ 400,\ 600,\ 800\}.
\]
As shown in Supplementary Fig.~\ref{fig:three_model_extended_nmax_tau_0.5_mean_f1}, the overall alignment performance continues to improve as \(N_{\max}\) increases across all three models, while the gains become progressively smaller at larger combination sizes and the performance approaches a plateau by \(N_{\max}=800\). Accordingly, unless otherwise specified, all subsequent cross-layer compositional semantic alignment analyses use \(\tau_P=0.5\) and \(N_{\max}=800\).

For each sememe, we record the final set of selected atoms and their source layers, the F1 improvement resulting from each successive atom addition, and the precision, recall and F1 score of the final combination. We also retain the entities covered jointly by the combination and the target sememe, the entities covered only by the combination, and the sememe entities that remain uncovered by the combination. These quantities are used for further comparisons between single-atom matches and multi-atom combinations in terms of semantic purity and coverage.

To improve the computational efficiency of large-scale cross-layer search, we represent the atom--entity and sememe--entity correspondences as sparse binary matrices and compute set overlaps and candidate statistics using batched sparse-matrix operations. This implementation is used solely for computational efficiency and does not alter the set definitions or search criteria described above.

\begin{figure}
    \centering
    \includegraphics[width=0.98\textwidth]{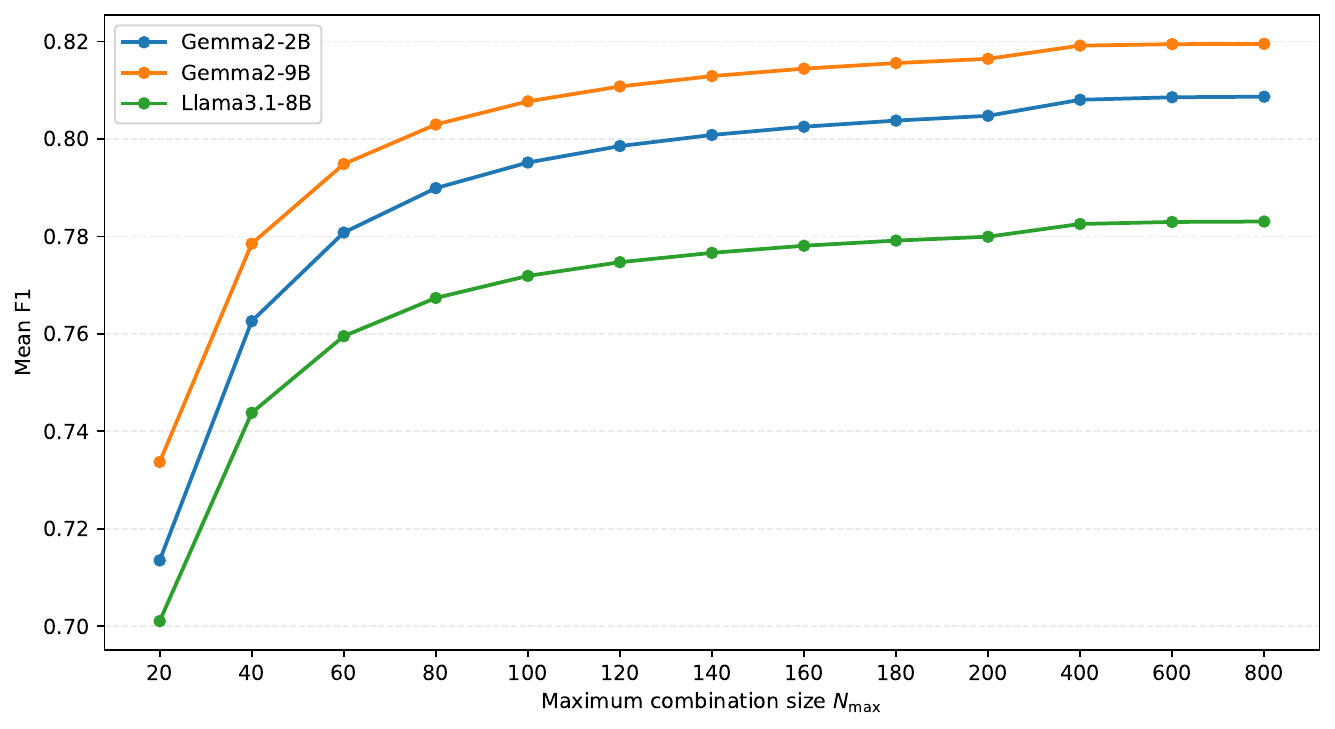}
    \caption{\textbf{Effect of the maximum combination size on cross-layer compositional semantic alignment.} Mean final F1 score across valid HowNet sememes as a function of the maximum combination size \(N_{\max}\) for Gemma2-2B, Gemma2-9B and Llama3.1-8B, with the candidate-filtering precision threshold fixed at \(\tau_P=0.5\). For all three models, increasing \(N_{\max}\) progressively improves compositional semantic alignment, with the gains becoming substantially smaller at larger combination sizes and the performance approaching a plateau by \(N_{\max}=800\). These results support the use of \(N_{\max}=800\) in the subsequent cross-layer compositional semantic alignment analyses.}
    \label{fig:three_model_extended_nmax_tau_0.5_mean_f1}
\end{figure}

\stopcontents[appendix]

\end{appendices}

\end{document}